\documentclass[sn-mathphys]{sn-jnl}

\jyear{2021}

\usepackage{subcaption}
\usepackage{multirow}

\theoremstyle{thmstyleone}\newtheorem{theorem}{Theorem}

\theoremstyle{thmstyletwo}

\theoremstyle{thmstylethree}\newtheorem{definition}{Definition}

\usepackage{amsthm}
\usepackage{amsmath}
\usepackage{xspace}
\usepackage{amssymb}
\usepackage{color}

\usepackage{bbm}

\def\Int{\mathbb{N}}
\def\NonNegInt{\mathbb{N}_{0}}
\def\Real{\mathbb{R}}

\newcommand{\dbtopo}{DBSR-46K}
\newcommand{\dbtopocomplex}{DBSR-cplx46K}
\newcommand{\mnistcomplex}{MNIST-cplx70k}
\newcommand{\mnistcomplexaug}{MNIST-cplx70k+AUG}

\newcommand{\determin}{Deterministic}
\newcommand{\veercnn}{VeerCNN}
\newcommand{\gcae}{GCAE}
\newcommand{\resnetoned}{ResNet1D}
\newcommand{\ddsl}{DDSL}
\newcommand{\ddsllenet}{\ddsl+LeNet5}
\newcommand{\nuftmlp}{NUFTspec}

\newcommand{\kdeltaenc}{KDelta}

\newcommand{\geomset}{\mathcal{G}}
\newcommand{\pgonset}{\mathcal{P}}
\newcommand{\mtpgonset}{\mathcal{Q}}

\newcommand{\mtpgonidx}{k}
\newcommand{\pgonidx}{i}
\newcommand{\holeidx}{j}

\newcommand{\geom}{g}
\newcommand{\pgon}{p}
\newcommand{\mtpgon}{q}

\newcommand{\border}{\mathbf{B}}
\newcommand{\numborderpt}[1]{N_{b_{#1}}}
\newcommand{\holeset}{h}
\newcommand{\hole}{\mathbf{H}}
\newcommand{\numholept}[1]{N_{\holeset_{#1}}}

\newcommand{\pgonenc}{Enc}
\newcommand{\pgonembdim}{d}

\newcommand{\onemat}{\mathbf{1}}

\newcommand{\geomsetall}{\mathcal{G}^*}
\newcommand{\params}{\theta}

\newcommand{\ptcoord}{\mathbf{x}}
\newcommand{\loopmat}{\mathbf{L}}
\newcommand{\loopdelta}{s}

\newcommand{\pgonemb}{\mathbf{p}}

\newcommand{\simplexmesh}{\mathcal{S}}
\newcommand{\simplex}{\mathbf{S}}
\newcommand{\simplexidx}{n}
\newcommand{\simplexdim}{j}

\newcommand{\vtxmat}{\mathbf{V}}
\newcommand{\edgmat}{\mathbf{E}}
\newcommand{\desmat}{\mathbf{D}}

\newcommand{\numpgonpt}[1]{N_{g_{#1}}}
\newcommand{\desdim}{d_{d}}

\newcommand{\edge}{\varepsilon}

\newcommand{\fftfreq}{\mathbf{w}}
\newcommand{\fftfreqidx}{k}
\newcommand{\fftfreqnum}{N_{w}}

\newcommand{\fftfreqnumX}{N_{wx}}
\newcommand{\fftfreqnumY}{N_{wy}}

\newcommand{\fftfreqitem}[1]{w_{#1}}
\newcommand{\maxscale}{w_{max}}
\newcommand{\minscale}{w_{min}}
\newcommand{\scaleratio}{g}
\newcommand{\certainscale}{u}
\newcommand{\nscale}{U}
\newcommand{\fftfreqvec}{W}
\newcommand{\fftfreqmat}{\mathcal{W}}

\newcommand{\linearfreqmtd}{fft}
\newcommand{\geometricfreqmtd}{gmf} 

\newcommand{\pcf}{f}
\newcommand{\desfunc}{\rho}

\newcommand{\fftpcf}{F}
\newcommand{\simplexsign}{\mu}
\newcommand{\simplexcont}{C}
\newcommand{\simplexcontratio}{\gamma}

\newcommand{\expbase}{e}

\newcommand{\jacobmat}{J}

\newcommand{\nuftspecmlpfunc}{MLP_{\fftpcf}}
\newcommand{\nuftspecmlplayernum}{K_{\fftpcf}}
\newcommand{\nuftspecnorm}{\Psi}

\newcommand{\nuftnumhiddenlayer}{h}
\newcommand{\nuftnumhiddendim}{o}

\newcommand{\ptemb}{\mathbf{l}}
\newcommand{\ptembmat}{\mathbf{L}}
\newcommand{\ptidx}{m}
\newcommand{\kdelta}{t}

\newcommand{\resnetonedlayer}{\mbox{ResNet1D}_{cp}}
\newcommand{\resnetonedlayeridx}{k}
\newcommand{\resnetonedlayernum}{\mathcal{K}}

\newcommand{\cnnonedlayer}{\mbox{CNN1D}}
\newcommand{\batchnormonedlayer}{\mbox{BN1D}}
\newcommand{\maxpoolonedlayer}{\mbox{MP1D}}
\newcommand{\relu}{\mbox{ReLU}}
\newcommand{\globalmaxpoolonedlayer}{\mbox{GMP1D}}
\newcommand{\dropout}{\mbox{DP}}

\newcommand{\shapecla}{y}
\newcommand{\shapeclamlpfunc}{MLP_{shp}}

\newcommand{\sparelmlpfunc}{MLP_{rel}}
\newcommand{\softmax}{softmax}

\newcommand{\relat}{r}
\newcommand{\relset}{R}
\newcommand{\relnum}{N_{\relat}}
\newcommand{\relidx}{i}

\newcommand{\pcavar}{\sum_{PCA}}
\newcommand{\pcadim}{K_{PCA}}

\newcommand{\lr}{lr}

\newcommand{\ent}{e}
\newcommand{\subject}{sub}
\newcommand{\object}{obj}

\newcommand{\arearatio}{R_{A}}

\newcommand{\tripleset}{\mathcal{T}}
\newcommand{\entset}{\mathcal{E}}

\newcommand{\reviseone}[1]{\textcolor{black}{#1}} 
\newcommand{\rvtwo}[1]{\textcolor{black}{#1}}  

\begin{document}

\title[\reviseone{Polygon Representation Learning}]{\reviseone{
Towards General-Purpose Representation Learning of Polygonal Geometries }}

\author*[1,2,3,4]{\fnm{Gengchen} \sur{Mai}}\email{gengchen.mai25@uga.edu}

\author[5]{\fnm{Chiyu} \sur{Jiang}}\email{maxjiang93@gmail.com}

\author[6]{\fnm{Weiwei} \sur{Sun}}\email{weiweis@cs.ubc.ca}

\author[7,3,4]{\fnm{Rui} \sur{Zhu}}\email{rui.zhu@bristol.ac.uk}

\author[8]{\fnm{Yao} \sur{Xuan}}\email{yxuan@math.ucsb.edu}

\author[3,4]{\fnm{Ling} \sur{Cai}}\email{ling.cai@geog.ucsb.edu}

\author[3,4,9]{\fnm{Krzysztof} \sur{Janowicz}}\email{jano@geog.ucsb.edu}

\author[2,10]{\fnm{Stefano} \sur{Ermon}}\email{ermon@cs.stanford.edu}

\author[11]{\fnm{Ni} \sur{Lao}$^{11+}$} \email{nlao@google.com} 

\affil*[1]{\orgdiv{Spatially Explicit Artificial Intelligence Lab, Department of Geography}, \orgname{University of Georgia}, \orgaddress{\city{Athens}, \postcode{30602}, \state{Georgia}, \country{USA}}}

\affil*[2]{\orgdiv{Department of Computer Science}, \orgname{Stanford University}, \orgaddress{\city{Stanford}, \postcode{94305}, \state{California}, \country{USA}}}

\affil*[3]{\orgdiv{STKO Lab}, \orgname{University of California Santa Barbara}, \orgaddress{\city{Santa Barbara}, \postcode{93106}, \state{California}, \country{USA}}}

\affil*[4]{\orgdiv{Center for Spatial Studies}, \orgname{University of California Santa Barbara}, \orgaddress{\city{Santa Barbara}, \postcode{93106}, \state{California}, \country{USA}}}

\affil[5]{\orgdiv{Department of Mechanical Engineering}, \orgname{University of California Berkeley}, \orgaddress{\city{Berkeley}, \postcode{94720}, \state{California}, \country{USA}}}

\affil[6]{\orgdiv{Department of Computer Science}, \orgname{University of British Columbia}, \orgaddress{\city{Vancouver}, \postcode{V6T 1Z4}, \state{British Columbia}, \country{Canada}}}

\affil[7]{\orgdiv{School of Geographical Sciences}, \orgname{University of Bristol}, \orgaddress{\city{Bristol}, \postcode{BS8 1TH},  \country{United Kingdom}}}

\affil[8]{\orgdiv{Department of Mathematics}, \orgname{University of California Santa Barbara}, \orgaddress{\city{Santa Barbara}, \postcode{93106}, \state{California}, \country{USA}}}

\affil[9]{\orgdiv{Department of Geography and Regional Research}, \orgname{ University of Vienna},
\orgaddress{\city{Vienna}, \postcode{1040}, \country{Austria}} }

\affil[10]{\orgname{Chan Zuckerberg Biohub}, \orgaddress{\city{San Francisco}, \postcode{94158}, \state{California}, \country{USA}} }

\affil[11]{\orgname{Google}, \orgaddress{\city{Mountain View}, \postcode{94043}, \state{California}, \country{USA}} \\
{$^+$ Work done while working at mosaix.ai}}

\abstract{

\reviseone{Neural network representation learning for} 
spatial data (e.g., points, polylines, polygons, and networks) 
is a common need for geographic artificial intelligence (GeoAI) problems. 
In recent years, \reviseone{\rvtwo{many advancements} have been made} in representation learning for points, polylines, and networks, whereas \reviseone{little progress has been made for} polygons, especially complex polygonal geometries.
In this work, we focus on developing a general-purpose polygon encoding model, which \reviseone{can encode} a polygonal geometry (with or without holes, single or multipolygons) into \reviseone{an} embedding space.
\reviseone{The result embeddings can be leveraged directly (or finetuned) \rvtwo{for} downstream tasks}
such as shape classification, spatial relation prediction, building pattern classification, cartographic building generalization, and so on. 
\rvtwo{To achieve model generalizability guarantees, we identify a few desirable properties that the encoder should} satisfy:
loop \reviseone{origin} invariance, trivial vertex invariance, part permutation invariance, and topology awareness. 
\reviseone{We explore two different designs for the encoder: one derives all representations in the spatial domain and can naturally capture local structures of polygons; the other leverages spectral domain representations and can easily capture global structures of polygons. }
\reviseone{For the spatial domain approach we} propose \resnetoned, a 1D CNN-based polygon encoder,  which  uses circular padding to achieve loop origin invariance \reviseone{on simple polygons.}
\reviseone{For the spectral domain approach we develop \nuftmlp~based on} Non-Uniform Fourier Transformation (NUFT), which naturally satisfies all the desired properties. 
We conduct experiments on two different tasks: 1) polygon shape classification based on the commonly used MNIST dataset; 2) polygon-based spatial relation prediction based on two new \rvtwo{datasets (\dbtopo~and \dbtopocomplex)} constructed from OpenStreetMap and DBpedia.   
Our results show that \nuftmlp~and \resnetoned~outperform multiple existing baselines with significant margins. While \resnetoned~suffers from model performance degradation after  shape-invariance geometry modifications, \nuftmlp~is very robust to these modifications due to the nature of the NUFT representation.
\nuftmlp~is able to jointly consider all parts of a multipolygon and their spatial relations during prediction while \resnetoned~can recognize the shape details which are sometimes important for classification.
\reviseone{This result points to a promising research direction of combining spatial and spectral representations.}

 }


\maketitle
\vspace{-0.5cm}
\section{Introduction}   \label{sec:intro}

Deep neural networks have shown great success for numerous tasks from computer vision, natural language processing, to audio analysis, \rvtwo{in} which the underlining data is usually in a regular structure such as grids (e.g., images) or sequences (e.g., sentences, audios) \citep{bronstein2017geometric,mai2021review}. These successes can be largely attributed to the fact that such \reviseone{regular data structures are natively supported by} the neural networks \citep{bronstein2017geometric}. 
For example, convolutional neural network\rvtwo{s} (CNN) \rvtwo{are} naturally suitable for image and video analysis. Recurrent neural network\rvtwo{s} (RNN) \rvtwo{are} suitable for data with sequence structures such as sentences and time series. 
However, it is hard to apply similar models on data with \reviseone{more complex} structures. 
Recent years have witnessed \rvtwo{increasing} interests in geometric deep learning \citep{bronstein2017geometric,monti2017geometric}, which focuses on developing deep models for non-Euclidean geometric data such as graphs \citep{defferrard2016convolutional,kipf2016semi,hamilton2017inductive,schlichtkrull2018modeling,cai2019transgcn,mai2020se}, points \citep{qi2017pointnet,li2018pointcnn,mac2019presence,mai2020multiscale}, and manifolds \citep{masci2015geodesic,monti2017geometric} that have rather irregular structure\rvtwo{s}. In fact, deep learning models on irregularly structured data or non-Euclidean geometric data have various applications in different domains such as computational social science (e.g., social network \citep{lazer2009life,fan2019graph}), chemistry (e.g., organic molecules~\citep{gilmer2017neural}), bioinformatics (e.g., gene regulatory network~\citep{davidson2002genomic}), and geoscience (e.g., traffic network~\citep{li2019diffusion,cai2020traffic}, air quality sensor network \citep{lin2018exploiting}, weather sensor networks~\citep{apple2020kriging}, and species occurrences~\citep{mac2019presence,mai2020multiscale}). 
\reviseone{This trend indicates that}
representing various types of spatial data \reviseone{in an embedding} space \reviseone{for downstream neural network models} is an important task for geographic artificial intelligence (GeoAI) research \citep{mai2021review}. 

\begin{figure*}[ht!]
	\centering \tiny
	\vspace*{-0.2cm}
	\begin{subfigure}[b]{0.49\textwidth}  
		\centering 
		\includegraphics[width=\textwidth]{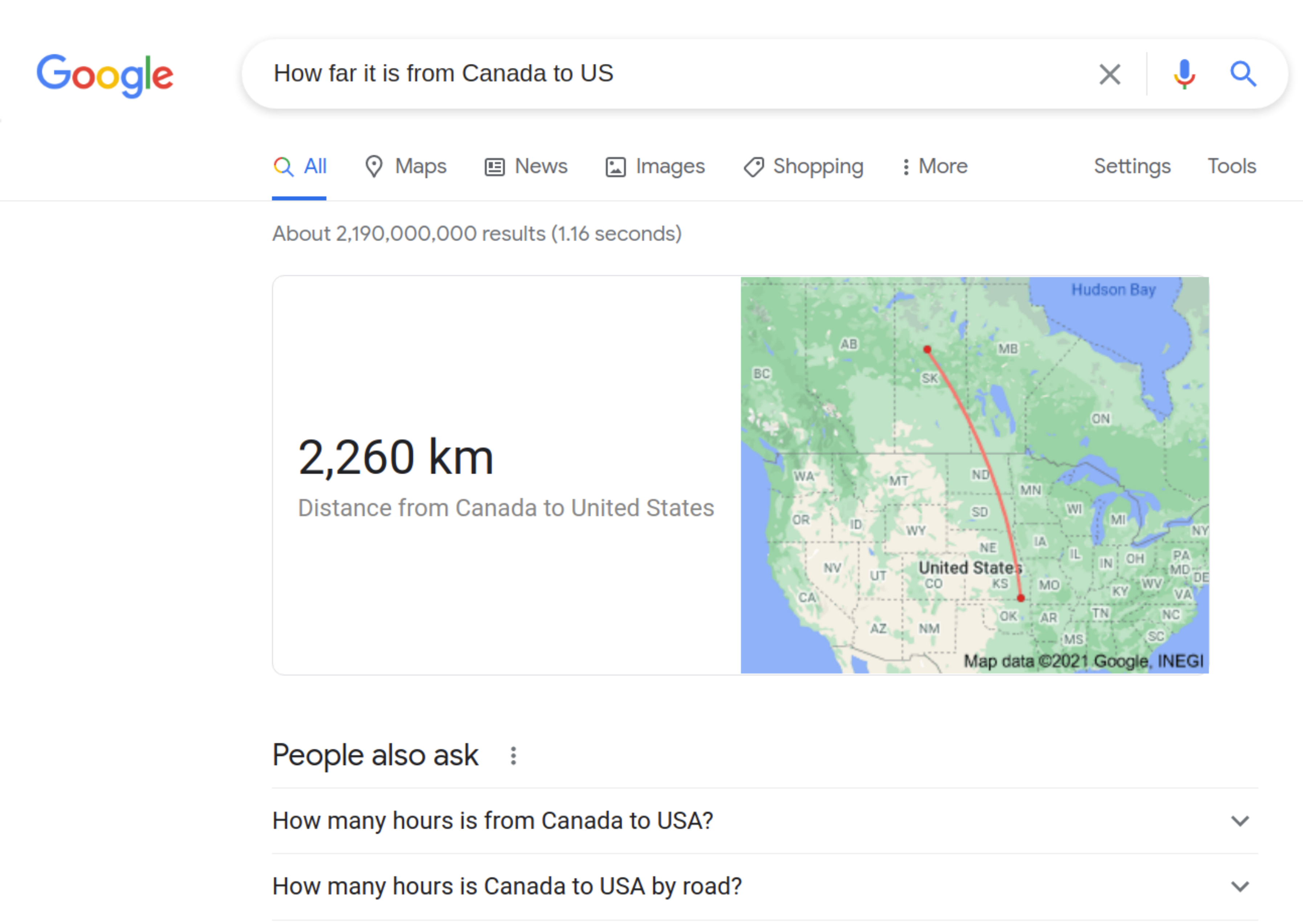}\vspace*{-0.2cm}
		\caption[]{{Geographic Question Answering
		}}    
		\label{fig:geoqa}
	\end{subfigure}
	\hfill
	\begin{subfigure}[b]{0.49\textwidth}  
		\centering 
		\includegraphics[width=\textwidth]{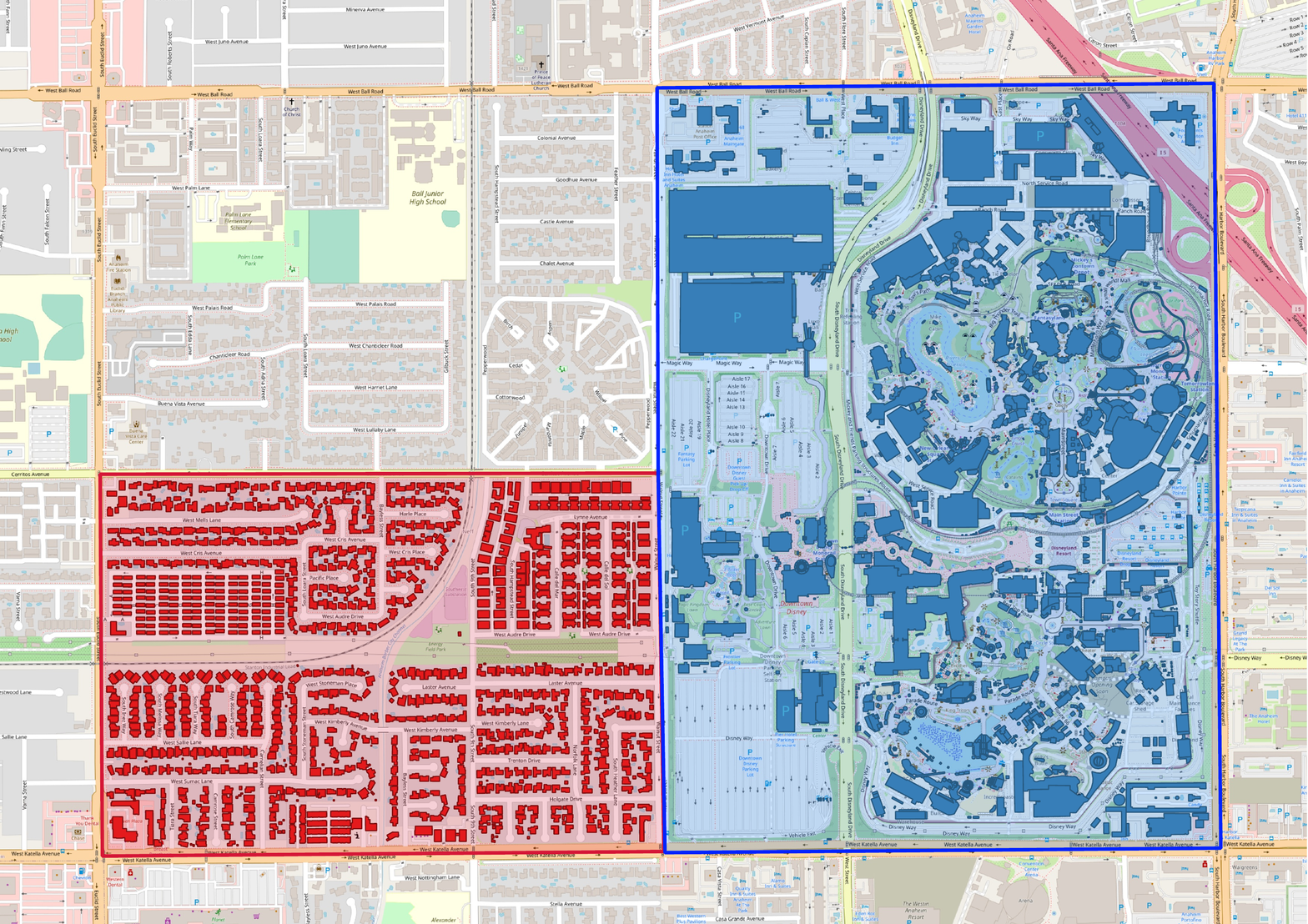}\vspace*{-0.2cm}
		\caption[]{{Building Pattern Classification
		}}    
		\label{fig:build_pattern}
	\end{subfigure}
	\caption{Two geospatial tasks involving polygon data:
	(a) GeoQA: many geographic questions can only be answered correctly based on polygon representations of geographic entities. Otherwise, it will yield an incorrect answer (2260 km) such as `\textit{How far it is from Canada to US}'.
	(b) BPC: the shape, scale, and arrangement of building footprints in a neighborhood are indicative for
	\reviseone{the neighborhood types.}
\reviseone{The blue neighborhood (Disneyland Park in Los Angles) shows irregular patterns whereas the red neighborhood (residential area) shows regular patterns \citep{yan2019graph}.}
} 
	\vspace{-0.3cm}
	\label{fig:example_task}
\end{figure*}

Recently, we have seen many research advancements in developing representation learning models for points \citep{mac2019presence, mai2020multiscale,wu2021inductive,mai2021review}, polylines \citep{xu2018encoding,zhang2019sr,rao2020lstm}, and network \citep{li2017diffusion,cai2020traffic}. 
Compared with other non-Euclidean geometric data, 
few efforts have been taken to develop deep models on polygons, especially complex polygonal geometries (e.g., polygons with holes, multipolygons), despite the fact that they are widely utilized in multiple applications, especially (geo)spatial applications such as shape coding and classification~\citep{veer2018deep,yan2021graph}, building pattern classification (BPC)~\reviseone{\citep{he2018recognition,yan2019graph,bei2019spatial}}, building grouping~\citep{he2018recognition,yan2020graph}, cartographic building generalization~\citep{feng2019learning}, geographic question answering (GeoQA)~\citep{zelle1996learning,punjani2018template,scheider2021geo,mai2019relaxing,mai2020se,mai2021geographic}, and so on. 
Figure \ref{fig:example_task} \rvtwo{demonstrates the importance} of polygon data for two geospatial tasks - GeoQA and BPC. Without proper polygon representations of Canada and the US (Figure \ref{fig:geoqa}), Question `\textit{How far it is from Canada to US}' cannot be answered correctly even by state-of-the-art QA system\footnote{The answer to this brain teaser question should be 0 because Canada and the US are adjacent to each other. However, since Google utilizes geometric central points as the spatial representations for geographic entities, Google QA returns 2260 km as the answer as the distance between them.}. As for the BPC task (Figure \ref{fig:build_pattern}), the shape and arrangement of building polygons in a neighborhood are indicative for \reviseone{its types, e.g., regular or irregular building groups~\citep{yan2019graph}.}

\reviseone{A representation learning model on polygons is desired. 
In many previous GeoAI study, due to the lack of ways to directly encode polygons into the embedding space, researchers have to \rvtwo{rely on} feature engineering to convert polygon shapes into a set of predefined shape descriptors before feeding them into the neural networks. For 
building pattern classification, given a set of building polygons, Yan et al. \citep{yan2019graph}, He et al. \citep{he2018recognition}, and Bei et al. \citep{bei2019spatial} converted the polygon set into a graph in which each node represents a building polygon and edges represent the spatial adjacent relations among buildings. They compute a feature vector for each building polygon/node based on a set of predefined shape descriptors. These vectors are used as initial node embeddings for the following graph neural network for building pattern recognition. These feature engineering approaches have several disadvantages: 
1) these shape descriptors can not fully capture the shape information polygons have which yield information loss; 
2) lots of domain knowledge is needed to design these descriptors; 
3) this practice lacks generalizability – it is hard to used the developed shape descriptors in other polygon tasks. 
In contrast, developing a general-purpose polygon encoder has several advantages: 
1) it allows us to develop end-to-end neural architectures directly taking polygons as inputs which increases the model expressivity; 
2) it eliminates the need of domain knowledge when handling polygon data; 
3) this model is task agnostic and can benefit a wide range of GeoAI tasks. 
}

\rvtwo{For the object instance segmentation task, existing deep models decode} a simple polygon\footnote{A simple polygon is a polygon that does not intersect itself and has no holes.} \rvtwo{based on the object mask image} ~\citep{sun2014free,castrejon2017annotating,acuna2018efficient}. 
\reviseone{These approaches can be seen as a reverse process of polygon encoding and they cannot decode complex polygonal geometries.
}
In contrast, \textbf{\reviseone{we propose to develop general-purpose polygon representation learning models, which directly encode}
a polygonal geometry (with or without holes, single or multipolygons) in an embedding space}. 
\reviseone{
Furthermore, we identify a few desirable properties for \rvtwo{polygon encoders to guarantee} their model generalizability: \reviseone{\textit{loop origin invariance}}, \textit{trivial vertex invariance}, \textit{part permutation invariance}, and \textit{topology awareness}, which will be discussed in detail in Section \ref{sec:prob_stat}. }
The resulting polygon embeddings can be subsequently utilized in multiple downstream tasks such as  shape classification~\citep{bai2009integrating,wang2014bag,veer2018deep,yan2021graph}, spatial relation prediction between geographic entities~\citep{regalia2019computing}, GeoQA~\citep{punjani2018template}, and so on.

\begin{figure*}[t!]
	\centering \tiny
	\vspace*{-0.2cm}
	\begin{subfigure}[b]{1.0\textwidth}  
		\centering 
		\includegraphics[width=\textwidth]{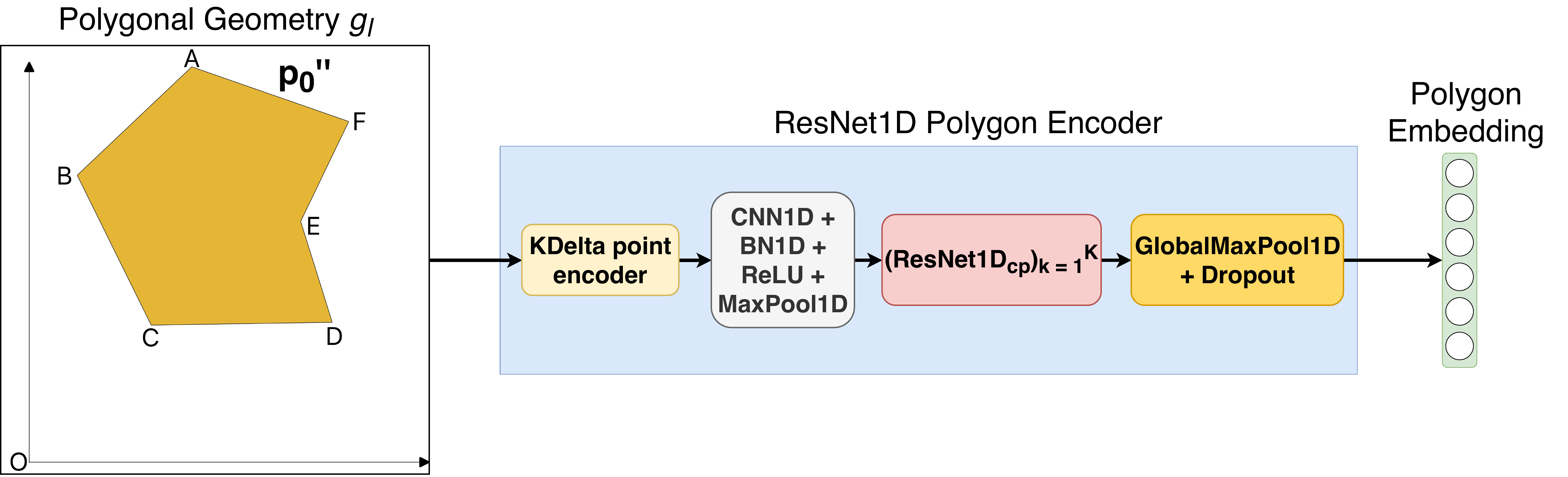}\vspace*{-0.2cm}
		\caption[]{\reviseone{\resnetoned~}}    
		\label{fig:restnet1d_model}
	\end{subfigure}
	\hfill
	\begin{subfigure}[b]{1.0\textwidth}  
		\centering 
		\includegraphics[width=\textwidth]{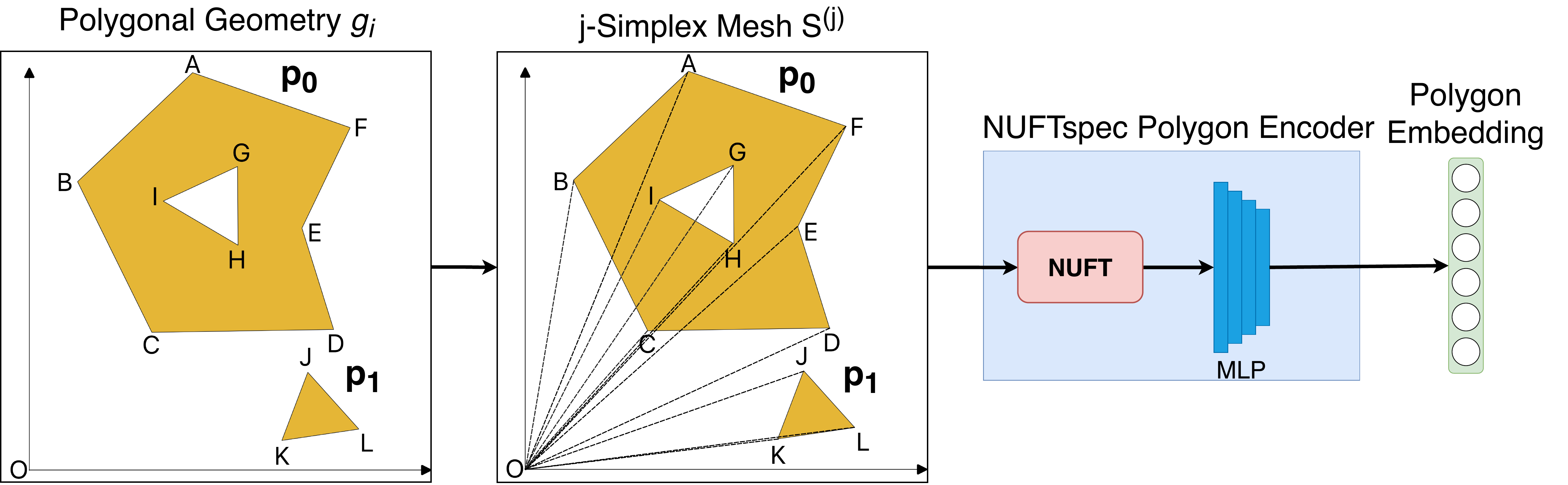}\vspace*{-0.2cm}
		\caption[]{\reviseone{\nuftmlp }}    
		\label{fig:nuftmlp_model}
	\end{subfigure}
	\caption{
	Illustrations of the proposed 
polygon encoders.
	\reviseone{
	(a) \resnetoned: given a simple polygon (i.e., without hole) $\geom_{l}$,  its exterior coordinate sequence can be encoded as a 1D point embedding sequence $\ptembmat$ by using \kdeltaenc~point encoder. $\ptembmat$ is fed into an initial 1D CNN layerand 1D max pooling layer, followed by $\resnetonedlayernum$ 1D ResNet layers with circular padding. Eventually, a global max pooling  layer produces the final polygon embedding.
	(b) \nuftmlp: given a polygonal geometry $\geom_{i}$ (i.e., a single polygon with/without holes or multipolygons), first it is converted into a j-simple mesh $\simplexmesh^{(\simplexdim)}$ (j = 2) based on \textit{auxiliary node method} (See Section \ref{sec:nuft_convert_simplex}). After NUFT, the Fourier features are fed into a multi-layer perceptron (MLP) to obtain the final polygon embedding.
	}
	} 
	\vspace{-0.3cm}
	\label{fig:pgon_enc_model}
\end{figure*} 
\reviseone{
Existing polygon encoding approaches can be classified into two groups: spatial domain polygon encoders and spectral domain polygon encoders. Spatial domain polygon encoders such as VeerCNN \citep{veer2018deep}, GCAE \citep{yan2021graph}, directly learn polygon embeddings from polygon features in the spatial domain (e.g., vertex coordinate features). 
In contrast, spectral domain polygon encoders such as DDSL \citep{jiang2019ddsl,jiang2019convolutional} first convert a polygonal geometry into spectral domain features by using Fourier transformations and learn polygon embeddings based on these spectral features. Both practices have unique advantages and disadvantages.} 
\reviseone{To study the pros and cons of these two approaches, 
we propose two polygon encoders: \textit{\resnetoned}~ and \textit{\nuftmlp}.
\resnetoned~ directly takes the polygon features in the spatial domain, i.e., the polygon vertex coordinate sequences, and \reviseone{uses a} 1D convolutional neural network (CNN) based architecture to \reviseone{produce} polygon embeddings.
In contrast, \textit{\nuftmlp} 
first transforms a polygonal geometry into the spectral domain by the Non-Uniform Fourier Transformation (NUFT) and then learns polygon embeddings from these spectral features using feed forward layers.
Figure \ref{fig:restnet1d_model} and \ref{fig:nuftmlp_model} illustrate the general model architectures of \resnetoned~and \nuftmlp~polygon encoder respectively.}

We compare the effectiveness of \reviseone{\resnetoned~and \nuftmlp~}with various deterministic or deep learning-based baselines on two types of tasks -- 1) polygon shape classification and 2) polygon-based spatial relation prediction. 
\reviseone{For the first task, we}
show that \reviseone{both \resnetoned~and \nuftmlp~}are able to outperform multiple baselines with statistically significant margins \reviseone{whereas \resnetoned~are better at capturing local features of the polygons, and \nuftmlp~better at capturing global features of the polygons}. \reviseone{Our analysis shows that}
because of NUFT, \nuftmlp~is robust to multiple shape-invariant geometry modifications such as loop origin randomization, vertex upsampling, and part permutation \reviseone{whereas}
\resnetoned~suffers from significant performance degredations. 
\reviseone{For the spatial relation prediction task, }
\nuftmlp~outperforms \resnetoned, as well as multiple determinstic and deep learning-based baselines on both \reviseone{\dbtopo~and \dbtopocomplex~} datasets because it can learn robust polygon embeddings from the spectral domain derived from NUFT. 
In addition, experiments on both tasks show that 
compared with other NUFT-based methods such as \ddsl\citep{jiang2019ddsl}, \nuftmlp~is more flexible in the choice of NUFT frequency maps. Designing appropriate NUFT frequency maps for \nuftmlp~can help learn more robust and effective polygon representations which is the key to \reviseone{its better performance}. 
Our contribution can be summarized as follows:
\begin{enumerate}
    \item  We formally define the problem of representation learning on polygonal geometries (including simple polygons, polygons with holes, and multipolygons), and identify four desirable polygon encoding properties \reviseone{to test their model generalizability}.

    \item \reviseone{We propose two polygon encoders, \resnetoned~and \nuftmlp, which learn polygon embeddings from spatial and spectral feature domains respectively. }
\item \reviseone{We compare the performance of the proposed polygon encoders as well as multiple baseline models on two representative tasks -- shape classification and spatial relation prediction, and introduce three new datasets -- \mnistcomplex, \dbtopo, and \dbtopocomplex.}

    \item \reviseone{We provide a detailed analysis of the invariance/awareness properties \reviseone{on these two models and discuss the pros and cons of polygon representation learning in the spatial or spectral domain. Our analysis points to interesting future research directions.}
    }

\end{enumerate}

This paper is organized as follows: We discuss the motivation of polygon encoding in Section \ref{sec:necesity}. Then, in Section \ref{sec:prob_stat}, we define the problem of representation learning on polygons and discuss four expected polygon encoding properties. Related work are discussed in Section \ref{sec:related}. We present \reviseone{\resnetoned~and \nuftmlp~}polygon encoder and compare their properties in Section \ref{sec:method}. Experiments on shape classification and spatial relation prediction \rvtwo{tasks} are presented in Section \ref{sec:exp_shape_clas} and \ref{sec:exp_spa_rel}, respectively.
Finally, we conclude this work in Section \ref{sec:conclusion}.

\begin{figure*}[b!]
	\centering \tiny
	\vspace*{-0.2cm}
	\begin{subfigure}[b]{0.245\textwidth}  
		\centering 
		\includegraphics[width=\textwidth]{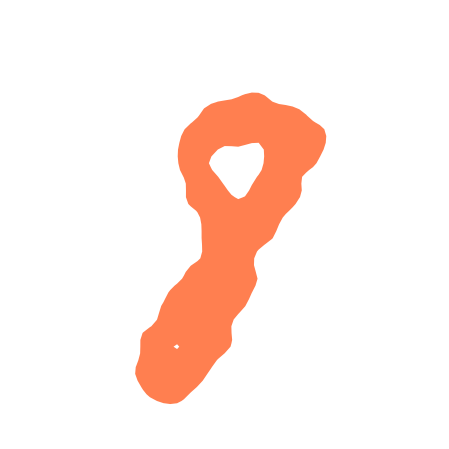}\vspace*{-0.2cm}
		\caption[]{{Polygon $\geom_{A}$
		}}    
		\label{fig:pgon_pgon_A}
	\end{subfigure}
	\hfill
	\begin{subfigure}[b]{0.245\textwidth}  
		\centering 
		\includegraphics[width=\textwidth]{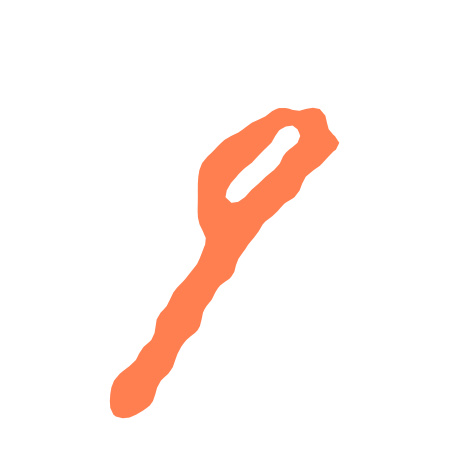}\vspace*{-0.2cm}
		\caption[]{{Polygon $\geom_{B}$
		}}    
		\label{fig:pgon_pgon_B}
	\end{subfigure}
	\hfill
	\begin{subfigure}[b]{0.245\textwidth}  
		\centering 
		\includegraphics[width=\textwidth]{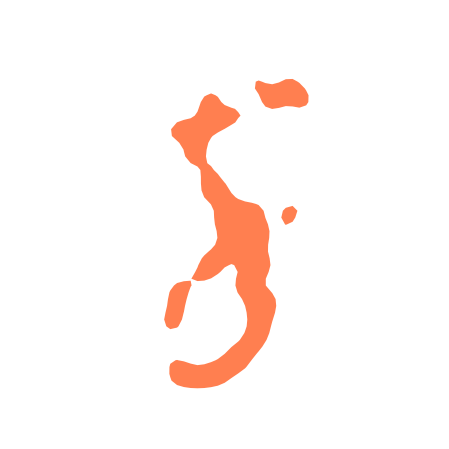}\vspace*{-0.2cm}
		\caption[]{{Polygon $\geom_{C}$
		}}    
		\label{fig:pgon_pgon_C}
	\end{subfigure}
	\hfill
	\begin{subfigure}[b]{0.245\textwidth}  
		\centering 
		\includegraphics[width=\textwidth]{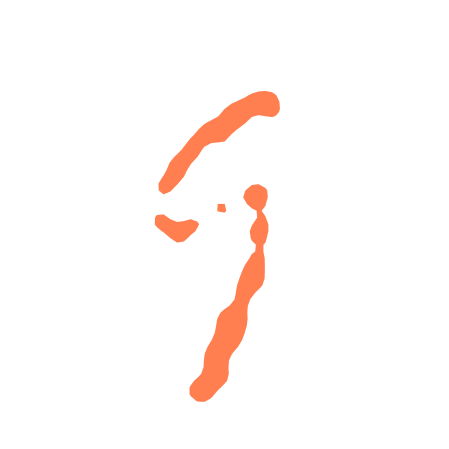}\vspace*{-0.2cm}
		\caption[]{{Polygon $\geom_{D}$
		}}    
		\label{fig:pgon_pgon_D}
	\end{subfigure}
	\caption{An illustration of the challenges of polygon encoding for shape classification task
	\reviseone{with examples from the \mnistcomplex~dataset.}
Please refer to Section \ref{sec:mnist_data} for  details of the data set construction.
(a) and (b) show that it is important to encode polygon holes since only encoding the polygon exteriors are not sufficient for shape classification.
	(c) and (d) shows that it is also critical to encode all parts of a multipolygon at once in order to capture its global shape, since none of its subpolygons is sufficient for shape classification.
	} 
	\label{fig:shape_cla_necesity_illu}
\end{figure*}

\section{Motivations}    \label{sec:necesity}

First of all, we discuss the challenges of representing polygons, especial complex polygonal geometries into \reviseone{an embedding space.} 
Given a polygonal geometry, there are two pieces of important information we are especially interested in: its shape and spatial relations with other geometries. 
\reviseone{These two pieces of information}
correspond to two polygon-based tasks -- shape classification and polygon-based spatial relation prediction.
In the following, we motivate polygon encoding from these two aspects by using real-world examples.

\subsection{Polygon Encoding for Shape Classification}   
\label{sec:shape_cla_necesity}

\reviseone{\textit{Shape classification}~\citep{kurnianggoro2018survey} (\reviseone{a.k.a.} shape-based object recognition) aims at predicting the category label for a given shape represented by a polygon or multipolygon.
}
Figure \ref{fig:shape_cla_necesity_illu} shows shape examples from the MNIST dataset~\citep{lecun1998gradient} illustrating the challenges of polygon encoding for shape classification, especially for complex polygonal geometries\reviseone{, which we summarize as the following: }
\begin{enumerate}
\item \reviseone{\textbf{Automatic representation learning for polygonal geometries}}:
Traditional shape classification models are based on handcrafted shape descriptors
    \reviseone{which capture different geometric properties based on vector geometries. Examples of shape descriptors}
are center of gravity, circularity ratio, radial distance, fractality, and so on~\citep{kurnianggoro2018survey,yan2019graph}. 
    For more advanced shape descriptors such as Bag of Contour Fragments (BCF)~\citep{wang2014bag}, the first step is also to \textit{vectorize} a given image into vector geometries -- polygons, so-called ``shape contours''. Then a feature extraction pipeline \reviseone{ can be applied to produce shape representations. }
\reviseone{P}olygon encoding can be seen as an alternative \reviseone{to} these traditional shape encoding models by replacing the feature extraction pipeline with an end-to-end deep learning model.
\item \textbf{Encoding the topology (such as holes) in polygons}: 
    Most existing work on polygon encoding~\citep{veer2018deep,yan2021graph} focus on encoding simple polygons, i.e., single part polygons without holes. This is insufficient to capture the 
    \reviseone{overall shapes for polygons with holes. }
For example, Polygon $\geom_{A}$ and $\geom_{B}$ shown in Figure \ref{fig:pgon_pgon_A} and \ref{fig:pgon_pgon_B} indicate two handwritten digits ``8'' and ``9'' from our \mnistcomplex~dataset. If we ignore their holes but only encode the exteriors, 
    \reviseone{Polygon $\geom_{A}$ might be misinterpreted as ``9'' or ``7'' and Polygon $\geom_{B}$ might be recognized as ``1'' or ``7''.}
\item \textbf{Jointly encoding all sub-parts of a multpolygon}: 
    Polygon $\geom_{C}$ and $\geom_{D}$ shown in Figure \ref{fig:pgon_pgon_C} and \ref{fig:pgon_pgon_D} indicate another two handwritten digits ``8'' and ``9'' from our \mnistcomplex~dataset. They are represented as two multipolygons. A model can make the correct shape classification only if it jointly considers all sub-parts of a multipolygon, \reviseone{whereas} none of \reviseone{the subpolygons of $\geom_{C}$ and $\geom_{D}$ \rvtwo{is}}
    sufficient for shape classification. 
\end{enumerate}

\begin{figure*}[t!]
	\centering \tiny
	\vspace*{-0.2cm}
	\begin{subfigure}[b]{0.33\textwidth}  
		\centering 
		\includegraphics[width=\textwidth]{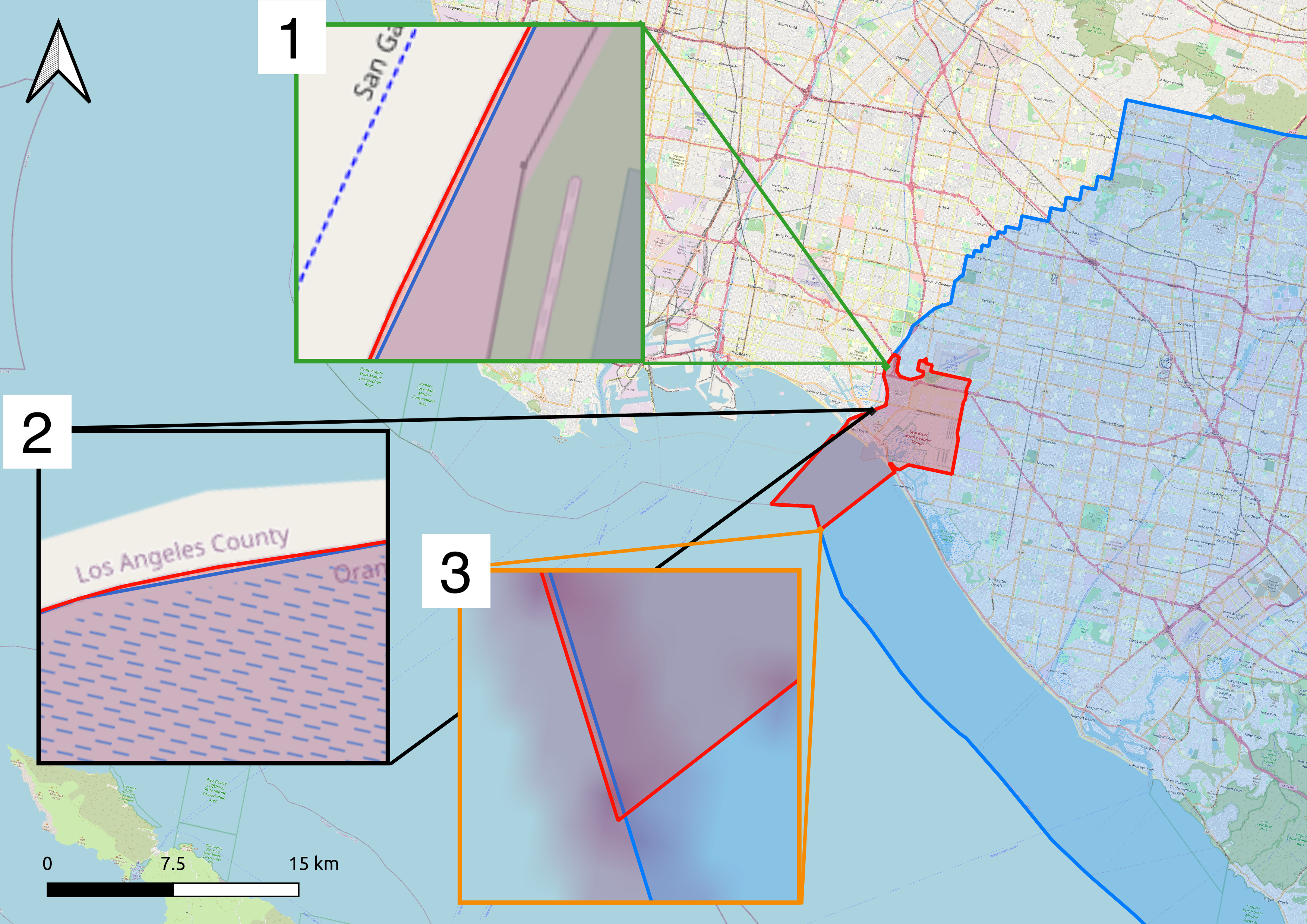}\vspace*{-0.2cm}
		\caption[]{{Sliver Polygon
		}}    
		\label{fig:sliver_pogon}
	\end{subfigure}
	\hspace{-0.1cm}
\begin{subfigure}[b]{0.33\textwidth}  
		\centering 
		\includegraphics[width=\textwidth]{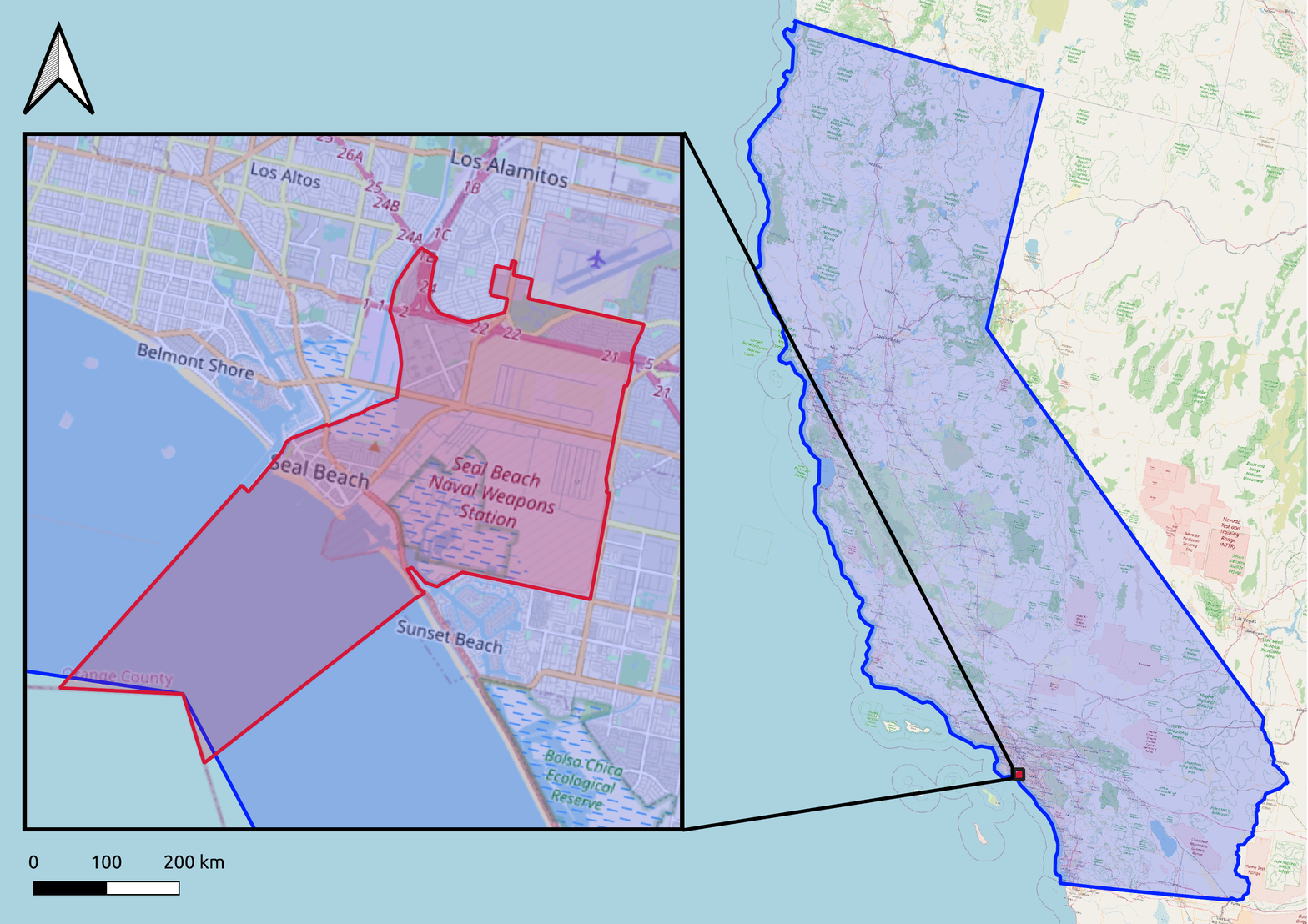}\vspace*{-0.2cm}
		\caption[]{\reviseone{Extreme Scale
		}}    
		\label{fig:scale}
	\end{subfigure}
	\hspace{-0.1cm}
\begin{subfigure}[b]{0.33\textwidth}  
		\centering 
		\includegraphics[width=\textwidth]{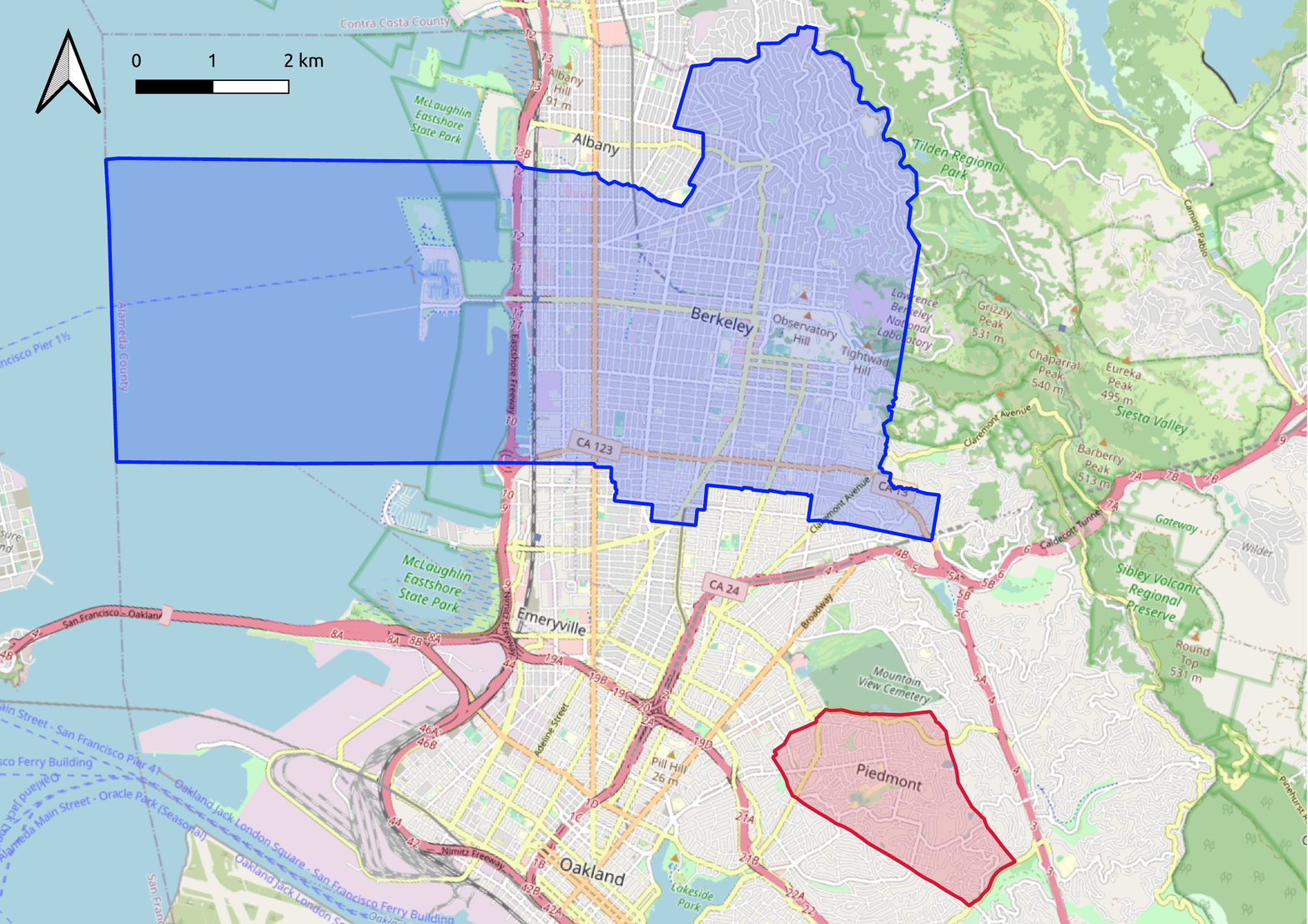}\vspace*{-0.2cm}
		\caption[]{\reviseone{Semantic Vagueness
		}}    
		\label{fig:direction}
	\end{subfigure}
	\caption{Three real-world examples \reviseone{illustrating  different challenges in predicting spatial relations between two polygons:
	\textbf{(a) Sliver Polygon Problem:}} \textit{dbr:Seal\_Beach,\_California} (the red polygon) should be tangential proper part (TPP) of \textit{dbr:Orange\_County,\_California} (the blue polygon). However, because of those sliver polygons shown in \reviseone{the zoom-in window 1, 2, and 3,} a deterministic spatial reasoner such as those used by PostGIS and GeoSPARQL-enabled triple stores will return \textit{partially overlapping} as the result.
\textbf{(b) Extreme Scale Problem:} \textit{dbr:Seal\_Beach,\_California} (the red polygon) 
	\reviseone{is extremely small }
compared with \textit{dbr:California} (the blue polygon). On one hand, a deterministic reasoner will pay too much attention to the geometry detail and predict wrong relations because of the sliver polygons \rvtwo{as shown in Figure (a). }On the other hand, if we convert these polygons into two images (e.g., two $128\times128$ images), that share the same bounding box, 
the red polygon becomes too small and cannot cover even one pixel which will also affect the result.
\textbf{(c) \reviseone{Semantic Vagueness Problem:}} \textit{dbr:Berkeley,\_California} (the blue polygon) \reviseone{is }in the north of \textit{dbr:Piedmont,\_California} (the red polygon) according to DBpedia and Wikipedia. 
	However, \reviseone{based on their polygon representation, their cardinal direction is vague which can be ``north'' or ``northwest''.}
} 
	\label{fig:example_rel}
\end{figure*}
 
\subsection{Polygon Encoding for Spatial Relation Prediction}    \label{sec:rel_necesity}

\reviseone{Next, we discuss the challenge of polygon encoding for predicting the correct spatial relations between two polygons, such as topological relations and relative cardinal directions.
}
At the first glance, this task \reviseone{may seem trivial, and a} GIScience expert may question the necessity of a polygon encoder
\reviseone{for spatial relation prediction given} that we have a set of well-defined deterministic spatial operators for spatial relation computation based on region connection calculus (RCC8)~\citep{randell1992spatial} or the dimensionally extended 9-intersection model (DE-9IM)~\citep{egenhofer1991point}. 
A computer vision researcher might also question \reviseone{its necessity} by suggesting an alternative approach that first rasterizes two polygons under consideration into two images that share the same bounding box so that a \reviseone{traditional computer vision model such as CNN} can be \reviseone{applied }
for relation prediction.
\reviseone{However, we will demonstrate}
 three different problems related to the spatial relation prediction task - \textit{sliver polygon problem}, \textit{extreme scale problem}, and \textit{semantic vagueness problem}, which are illustrated by 
\reviseone{Figure \ref{fig:example_rel} with real-world examples from OpenStreetMap.  In the following, we discuss how these problems \rvtwo{pose} challenges to
the above mentioned alternative approaches: }
\begin{enumerate}
    \item \textbf{Sliver polygon problem}: 
As shown in \reviseone{zoom-in window 1, 2, and 3 in Figure \ref{fig:sliver_pogon}, }
    those three tiny polygons yielded from the geometry difference between the red and blue polygon are called sliver polygons\footnote{In GIScience, sliver polygon is a technical term referring to the small unwanted polygons resulting from polygon intersection or difference.}. Here, \textit{dbr:Seal\_Beach,\_California} (the red polygon) should be tangential proper part (TPP) of \textit{dbr:Orange\_County,\_California} (the blue polygon). However, because of map digitization error, the boundary of the red polygon \reviseone{slightly stretches out of }the boundary of the blue polygon. A deterministic spatial operator will return ``intersect'' instead of ``part of'' as their relation. Sliver polygons are very common in map data, hard to prevent, and require a lot of efforts to correct, while deterministic spatial operators are very sensitive to them. \reviseone{Please refer to a detailed analysis on the sliver polygon problem on our \dbtopo~and \dbtopocomplex~dataset in Section \ref{sec:sliver_pgon_analysis}.}

    \item \reviseone{\textbf{Scale problem}:}
\rvtwo{Two} polygons might \reviseone{have very different sizes and thus need to use different map scales for visualization} (Figure \ref{fig:scale}). \textit{dbr:Seal\_Beach,\_California} (the red polygon) \reviseone{is extremely small }compared with \textit{dbr:California} (the blue polygon). When using deterministic spatial operators, sliver polygons will lead to wrong answers while as for the rasterization method, the red polygon become too small to occupy even one pixel of the image.
    
    \item \reviseone{\textbf{Semantic vagueness problem}:}
Some spatial relations such as cardinal direction relations are conceptually vague. While this vagueness can be well handled by neural networks through \reviseone{end-to-end learning from the labels, }
    it is hard to design a deterministic method to predict them. As shown in Figure \ref{fig:direction}, \textit{dbr:Berkeley,\_California} (the blue polygon) sits in the north of \textit{dbr:Piedmont,\_California} instead of \reviseone{northwest} according to DBpedia. 
    However, based on their polygon representations, both ``north'' and \reviseone{``northwest''} seems to be true. 
    In other words, their cardinal direction is vague. \end{enumerate}

\reviseone{
Because of those challenges, designing a general-purpose neural network-based representation learning model for polygonal geometries
} is necessary and can benefit multiple downstream applications.

 \section{Problem Statement}   \label{sec:prob_stat}

Based on the Open Geospatial Consortium (OGC) standard, we first give the definition of polygons and multipolygons. 

Let $\geomset = \{ \geom_{\pgonidx} \}$ be a set of polygonal geometries in a 2D Euclidean space $\Real^2$ where $\geomset$ is a union of a polygon set $\pgonset = \{ \pgon_{\pgonidx} \}$ and a multipolygon set $\mtpgonset = \{ \mtpgon_{\pgonidx} \}$, a.k.a $\geomset = \pgonset \cup \mtpgonset$ and $\pgonset \cap \mtpgonset = \emptyset$. 
We have $ \geom_{\pgonidx} \in \pgonset \vee \geom_{\pgonidx} \in \mtpgonset$.

\vspace{-0.4cm}
\begin{definition}[Polygon] \label{def:pgon}
Each {\em polygon} $\pgon_{\pgonidx}$ can be represented as a tuple $(\border_{\pgonidx}, \holeset_{\pgonidx} = \{\hole_{\pgonidx\holeidx}\})$ where $\border_{\pgonidx} \in \Real^{\numborderpt{\pgonidx} \times 2}$ indicates a point coordinate matrix for the exterior of $\pgon_\pgonidx$ defined in a \textit{counterclockwise} direction. $\holeset_\pgonidx = \{\hole_{\pgonidx\holeidx}\}$ is a set of holes for $\pgon_\pgonidx$ where each hole $\hole_{\pgonidx\holeidx} \in \Real^{\numholept{\pgonidx\holeidx} \times 2 }$ is a point coordinate matrix for one interior linear ring of $\pgon_{\pgonidx}$ defined in a \textit{clockwise} direction. $\numborderpt{\pgonidx}$ indicates the number of unique points in $\pgon_{\pgonidx}$'s exterior. The first and last point of $\border_{\pgonidx}$ are not the same and $\border_{\pgonidx}$ does not intersect with itself. Similar logic applies to each hole $\hole_{\pgonidx\holeidx}$ and $\numholept{\pgonidx\holeidx}$ is the number of unique points in the $\holeidx$th hole of $\pgon_{\pgonidx}$.  
\end{definition}
\vspace{-0.4cm}

\vspace{-0.4cm}
\begin{definition}[Multipolygon] \label{def:multipgon}
A {\em multipolygon} $\mtpgon_{\mtpgonidx} \in \mtpgonset$ is a set of polygons $\mtpgon_{\mtpgonidx} = \{ \pgon_{\mtpgonidx\pgonidx} \}$ which represents one entity (e.g., Japan, United States, and Santa Barbara County).
\end{definition}
\vspace{-0.4cm}

\vspace{-0.4cm}
\begin{definition}[Polygonal Geometry] \label{def:pgon_geometry}
A {\em polygonal geometry} $\geom_{\pgonidx} \in \pgonset \cup \mtpgonset$ can be either a polygon or a multipolygon. $\{\edge_{\pgonidx\simplexidx}\}_{\simplexidx=1}^{\numpgonpt{\pgonidx}}$ is defined as the set of all boundary segments/edges of the exterior(s) and interiors/holes of $\geom_{\pgonidx}$ or all its sub-polygons. 
$\numpgonpt{\pgonidx}$ is the total number of edges in $\geom_{\pgonidx}$ which is equal to the total number of vertices of $\geom_{\pgonidx}$.
\end{definition}
\vspace{-0.4cm}

\vspace{-0.4cm}
\begin{definition}[Simple Polygon and Complex Polygonal Geometry] \label{def:simple_complex_pgon}
If a polygonal geometry $\geom_{\pgonidx}$ is a single polygon without any holes, i.e., $\geom_{\pgonidx} = (\border_{\pgonidx}, \holeset_{\pgonidx} = \emptyset)$, we call it a {\em simple polygon}. Otherwise, we call it a complex polygonal geometry which might be a multipolygon or a polygon with holes.
\end{definition}
\vspace{-0.4cm}

\vspace{-0.4cm}
\begin{definition}[Distributed representation of polygonal geometries] \label{def:pgon_enc}
{\em Distributed representation of polygonal geometries in the 2D Euclidean space $\Real^2$} can be defined as a function $\pgonenc_{\geomset,\params}(\geom_{\pgonidx}): \geomsetall \to \Real^{\pgonembdim}$ which is parameterized by $\params$ and maps any polygonal geometry $\geom_{\pgonidx} \in \geomsetall$ in $\Real^2$ to a vector representation of $\pgonembdim$ dimension\footnote{We use $\pgonenc(\geom_{\pgonidx})$ to represent $\pgonenc_{\geomset,\params}(\geom_{\pgonidx})$ in the following}. Here $\geomsetall$ indicates the set of all possible polygonal geometries in $\Real^2$ and $\geomset \subseteq \geomsetall$.
\end{definition}
\vspace{-0.4cm}

\begin{figure*}[ht!]
	\centering \tiny
	\vspace*{-0.2cm}
	\begin{subfigure}[b]{0.4\textwidth}  
		\centering 
		\includegraphics[width=\textwidth]{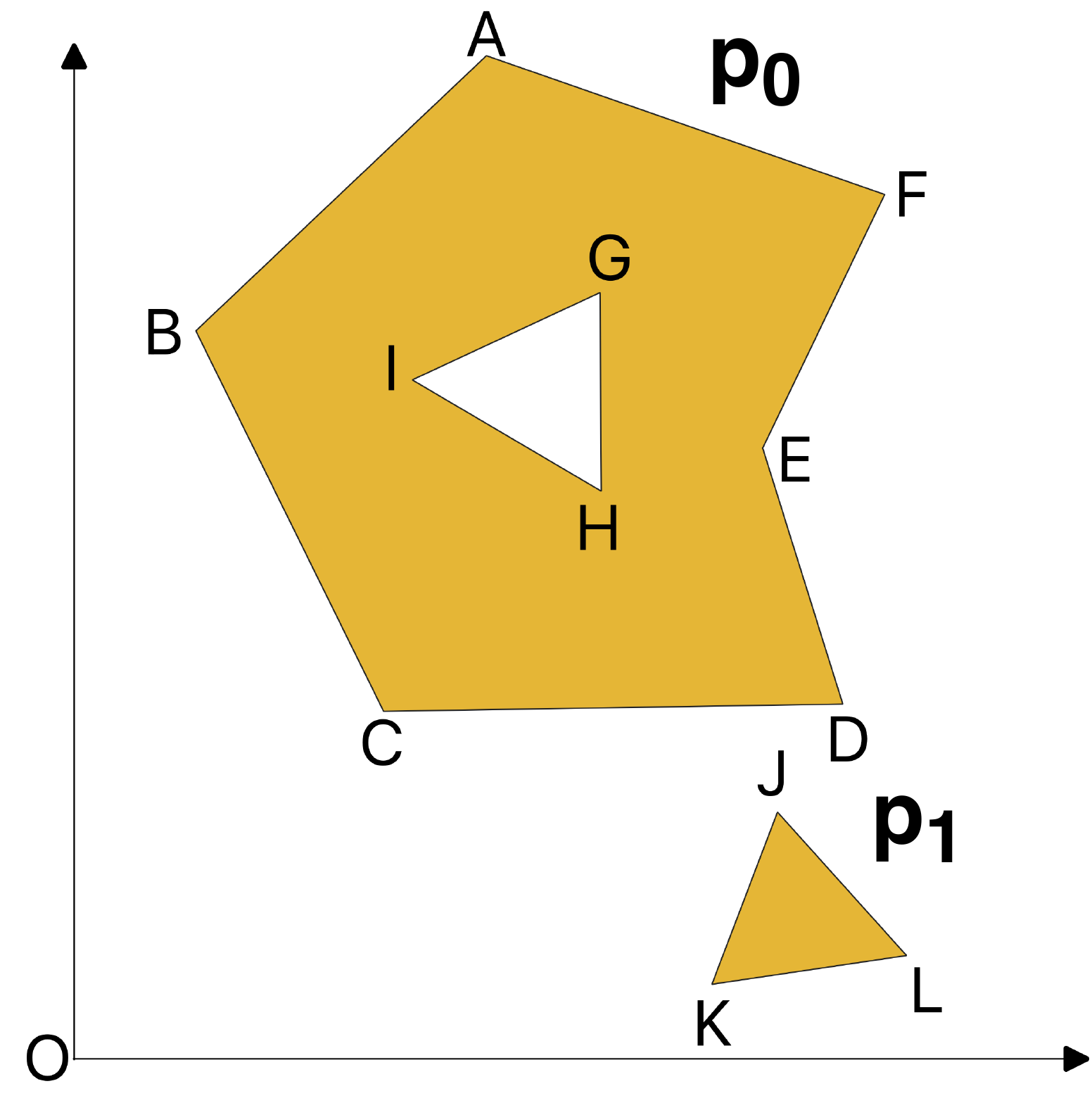}\vspace*{-0.2cm}
		\caption[]{\reviseone{$\mtpgon = \{\pgon_0, \pgon_1\}$
		}}    
		\label{fig:pgon}
	\end{subfigure}
	\hfill
	\begin{subfigure}[b]{0.4\textwidth}  
		\centering 
		\includegraphics[width=\textwidth]{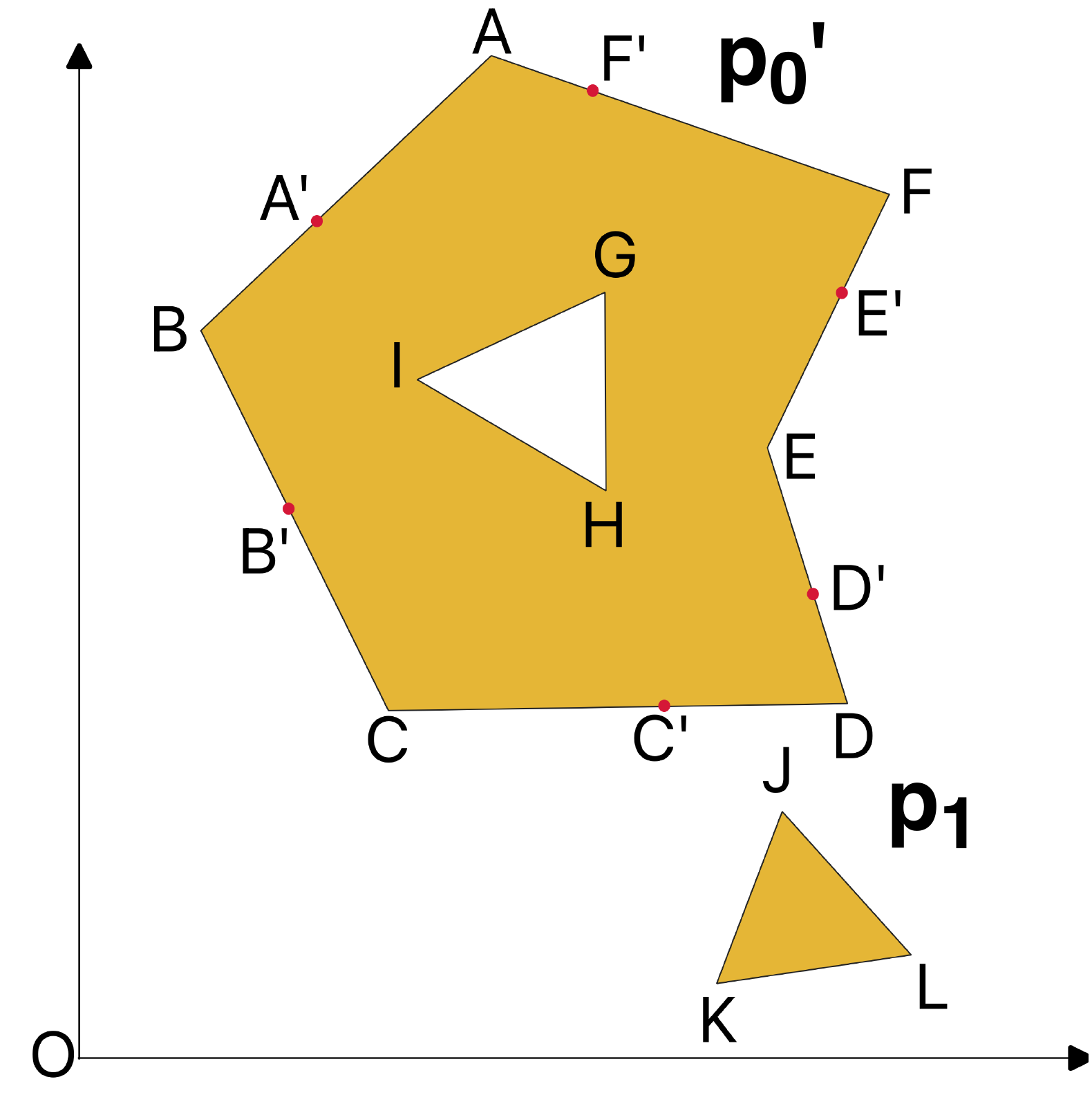}\vspace*{-0.2cm}
		\caption[]{\reviseone{$\mtpgon^{\prime} = \{\pgon_0^{\prime}, \pgon_1\}$
		}}    
		\label{fig:pgon_shape}
	\end{subfigure}
	\hfill
	\begin{subfigure}[b]{0.4\textwidth}  
		\centering 
		\includegraphics[width=\textwidth]{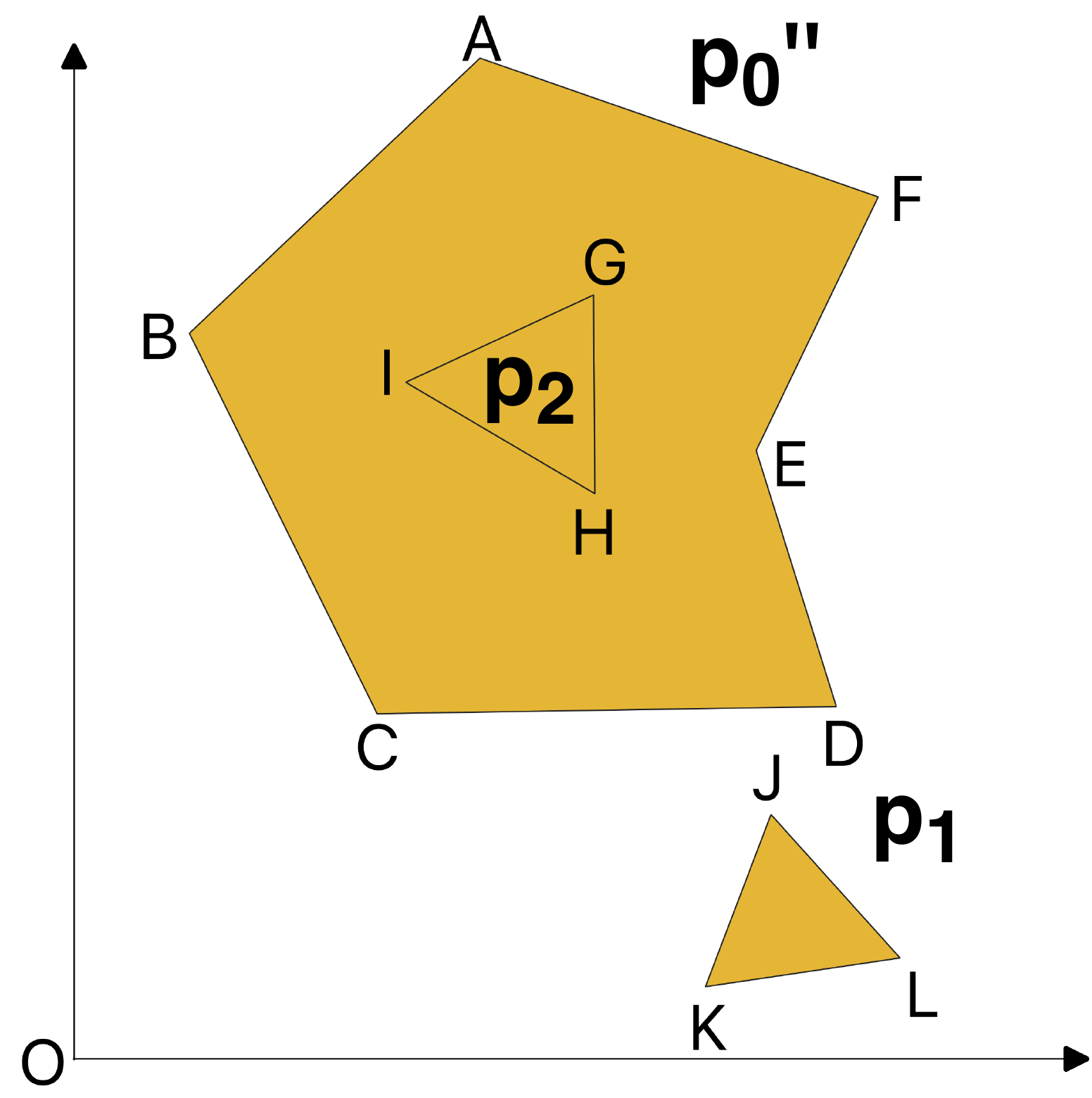}\vspace*{-0.2cm}
		\caption[]{\reviseone{$\mtpgon^{\prime\prime} = \{\pgon_0^{\prime\prime}, \pgon_1, \pgon_2\}$
		}}    
		\label{fig:pgon1}
	\end{subfigure}
	\hfill
	\begin{subfigure}[b]{0.4\textwidth}  
		\centering 
		\includegraphics[width=\textwidth]{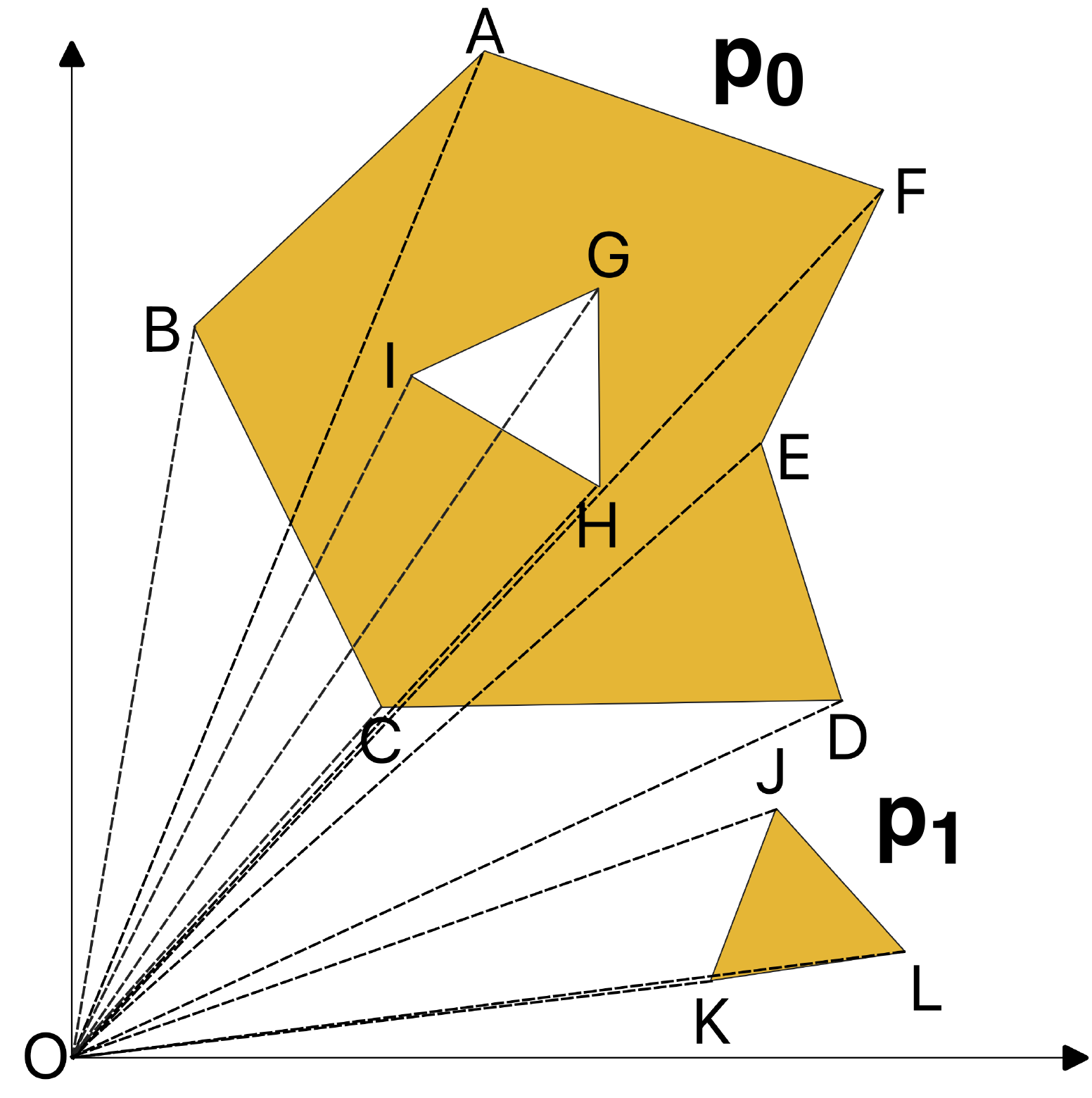}\vspace*{-0.2cm}
		\caption[]{\reviseone{2-simplex mesh $\simplexmesh^{(\simplexdim)}$
		}}    
		\label{fig:pgon_simplex}
	\end{subfigure}
	\caption{\rvtwo{Illustrations of the four polygon encoding properties and the auxiliary node method with a multipolygon.} 
	\textbf{(a)} A multipolygon $\mtpgon = \{\pgon_0, \pgon_1\}$ with two parts. $\pgon_0 = (\border_0, \holeset_0 = \{\hole_{00}\})$ has one hole and $\pgon_1 = (\border_1, \holeset_1 = \emptyset)$ has no hole. 
	\textbf{(b)} A multipolygon $\mtpgon^{\prime} = \{\pgon_0^{\prime}, \pgon_1\}$ where $\pgon_0^{\prime}$ has the same shape as $\pgon_0$ but adding additional 6 trivial vertices (red dots) -  $A^\prime, B^\prime, C^\prime, D^\prime, E^\prime, F^\prime$ - to its exterior.
	\textbf{(c)} A multipolygon  $\mtpgon^{\prime\prime} = \{\pgon_0^{\prime\prime}, \pgon_1, \pgon_2\}$ with three parts.  $\pgon_0^{\prime\prime} = (\border_0, \emptyset)$ is a simple polygon made up from the exterior of $\pgon_0$. $\pgon_2$ is a simple polygon made up from the boundary of $\pgon_0$'s interior $\hole_{00}$.
	\textbf{(d)} \rvtwo{The auxiliary node method} converts multipolygon $\mtpgon$ into a 2-simplex mesh $\simplexmesh^{(\simplexdim)}$ by adding the origin point $\ptcoord_O$ as another vertex (See Section \ref{sec:nuft_convert_simplex}).
	} 
	\label{fig:pgon_illu}
\end{figure*}

Figure \ref{fig:pgon} illustrates a multipolygon $\mtpgon = \{\pgon_0, \pgon_1\}$ where $\pgon_0 = (\border_0, \holeset_0 = \{\hole_{00}\})$ has one hole and $\pgon_1 = (\border_1, \holeset_1 = \emptyset)$ has no hole. \reviseone{Our objective is to develop general-purpose polygon encoders, so \textbf{model generalizability} is the key consideration. To test the generalizibility of a polygon encoder $\pgonenc(\geom_{\pgonidx})$, we propose four properties: }

\begin{enumerate}
    \item \reviseone{\textbf{ Loop origin invariance (Loop)}:} The encoding result of a polygon $\pgon_{\pgonidx}$ should be invariant when starting with different vertices to loop around its exterior/interior. Let consider $\pgon_0^{\prime\prime} = (\border_0, \emptyset)$ as a simple polygon made up from the exterior of $\pgon_0$. $\border_0$ can be written as $\border_0 = [\ptcoord_{A}^T;\ptcoord_{B}^T;\ptcoord_{C}^T;\ptcoord_{D}^T;\ptcoord_{E}^T;\ptcoord_{F}^T] \in \Real^{6 \times 2}$. 
Let $\border_0^{(\loopdelta)} = \loopmat_{\loopdelta}\border_0$ as another representation of $\pgon_0^{\prime\prime}$'s exterior where $\loopmat_{\loopdelta} \in \Real^{6 \times 6}$ is a loop matrix \reviseone{that} shifts the order of $\border_0$ by $\loopdelta$. For example, 
$\border_0^{(3)} = \loopmat_{3}\border_0 =  [\ptcoord_{D}^T;\ptcoord_{E}^T;\ptcoord_{F}^T;\ptcoord_{A}^T;\ptcoord_{B}^T;\ptcoord_{C}^T]$. Conceptually, we have $\pgon_0^{\prime\prime} = (\border_0, \emptyset) \doteq \pgon_0^{(\loopdelta)} = ( \loopmat_{\loopdelta}\border_0, \emptyset) \; \forall \loopdelta \in \{1,2,...,6\}$ where $\doteq$ indicates two geometries represent equivalent shape information. Loop \rvtwo{origin} invariance expects 
$\pgonenc(\pgon_0^{\prime\prime}) = \pgonenc(\pgon_0^{(\loopdelta)})$.
    
    \item \textbf{Trivial vertex invariance (TriV)}: The encoding result of a polygonal geometry should be invariant when we add/delete trivial vertices to/from its exterior or interiors. Trivial vertices are unimportant vertices such that adding or deleting them from polygons' exteriors or interiors does not change their overall shape and topology. For example, 6 red vertices - $A^\prime, B^\prime, C^\prime, D^\prime, E^\prime, F^\prime$ - (Figure \ref{fig:pgon_shape}) are trivial vertices of Polygon $\pgon_0^{\prime}$ since deleting them yield Polygon $\pgon_0$ which has the same shape as $\pgon_0^{\prime}$. We expect $\pgonenc(\pgon_0^{\prime}) = \pgonenc(\pgon_0)$. \rvtwo{It} is particularly difficult for 1D CNN- or RNN-based polygon encoders to achieve this since trivial vertices will significantly change the input polygon boundary coordinate sequences.
    
    \item \textbf{Part permutation invariance (ParP)}: The encoding result of a multipolygon $\mtpgon_{\pgonidx}$ should be invariant when permuting the feed-in order of its parts. For instance, the encoding result of $\pgonenc(\mtpgon)$ (Figure \ref{fig:pgon}) should not change when changing the feed-in order of $\pgon_0, \pgon_1$.
    
    \item \textbf{Topology awareness (Topo)}: The polygon encoder $\pgonenc(\geom_{\pgonidx})$ should be aware of the topology of the polygonal geometry $\geom_{\pgonidx}$. $\pgonenc(\geom_{\pgonidx})$ should not only encode the boundary information of $\geom_i$ but also be aware of the exterior and interior relationship\rvtwo{s}. For example, as shown in Figure \ref{fig:pgon} and \ref{fig:pgon1}, $\mtpgon = \{\pgon_0, \pgon_1\}$ and  $\mtpgon^{\prime\prime} = \{\pgon_0^{\prime\prime}, \pgon_1, \pgon_2\}$ are two multipolygons. $\pgon_0^{\prime\prime}$ and $\pgon_2$ are two simple polygons and $\pgon_2$ is inside of $\pgon_0^{\prime\prime}$. Although $\mtpgon$ and $\mtpgon^{\prime\prime}$ have the same boundary information, the encoding results of them should be different given their different topological information.
\end{enumerate}

As can be seen, these properties are unique requirements for encoding polygonal geometries. 
In certain scenarios, \textbf{translation invariance}, \textbf{scale invariance}, and \textbf{rotation invariance}, which require the encoding results of a polygon encoder unchanged when polygons are gone through translation/scale/rotation translations, are also expected in many shape related tasks such as shape classification, shape matching, shape retrieval~\citep{jiang2019ddsl}. However, in other tasks such as spatial relation prediction (including topological relations and cardinal direction relations) and GeoQA, translation/scale/rotation invariance are unwanted. For example, after a translation transformation on $\pgon_0$ (Figure \ref{fig:pgon}), the cardinal direction between $\pgon_0$ and $\pgon_1$ changes. So in this work, we primarily focus on the \reviseone{above mentioned}  four properties.

 \section{Related Work}  \label{sec:related}

\subsection{Object Instance Segmentation and Polygon Decoding} \label{sec:pgon_dec_related}
Most existing machine learning research involving polygons mainly focus on object instance segmentation and localization tasks. They aim at constructing a simple polygon as a localized object mask on an image. 
\rvtwo{Most existing works adopt a polygon refinement approach.}
For example, \rvtwo{both Zhang et al. \citep{zhang2012superedge} and Sun et al. \citep{sun2014free} proposed to first detect} the boundary fragments of polygons from images and then extract a simple polygon by finding the optimal circle linking these fragments into the object contours.

A recent deep learning approach, Polygon-RNN~\citep{castrejon2017annotating}, first encodes a given image with a \reviseone{deep CNN structure (similar to VGG \citep{simonyan2014very})} and then decodes the polygon mask of an object with a two-layer convolutional LSTM with skip connections. The RNN polygon decoder decodes one polygon vertex at each time step until \rvtwo{the end-of-sequence is decoded which indicates that} the polygon is closing. The first vertex is predicted with another CNN using a multi-task loss. 
Polygon-RNN++~\citep{acuna2018efficient} improves Polygon-RNN by adding a Gated Graph Neural Network (GGNN) \citep{li2016gated} after the RNN polygon decoder to increase the spatial resolution of the output polygon. 
Since Polygon-RNN uses cross-entropy loss which over-penalizes the model and is different from the evaluation metric,
Polygon-RNN++ changes the learning objective to reinforcement learning to directly optimize on the evaluation metric.

Similar to the polygon refinement idea, PolyTransform~\citep{liang2020polytransform} first uses a instance initialization module to provide a good polygon initialization for each individual object. And then a feature extraction network is used to extract embeddings for each polygon vertex. Next, PolyTransform uses a self-attention Transformer network to encode the exterior of each polygon and deform the initial polygon. This deforming network predicts the offset for each vertex on the polygon exterier such that the resulting polygon snaps better to the ground truth polygon (the object mask).

Compared with our polygon encoding model, these polygon decoding models treat polygons as output instead of input to the model. In addition, although Polygon-RNN++ and PolyTransform have sub-network modules that take a polygon as the input and refine/deform it into a more fine-grained polygon shape, both their modules -- GGNN for Polygon-RNN++ and Transformer for PolyTransform -- can only handle simple polygons \reviseone{but not complex plolygonal geometries. }Moreover, the GGNN for Polygon-RNN++ satisfies loop \reviseone{origin} invariance but not other properties while the Transformer module of PolyTransform can not satisfy any of those four properties in Section \ref{sec:prob_stat}.

\subsection{2D Shape Classification} \label{sec:shape_cla_related}

\reviseone{Stemming from }image analysis, shape representation is a fundamental problem in computer vision~\citep{wang2014bag}, since shape and texture are two most important aspects for object analysis. 
Shape classification aims at classifying an object (e.g., animal, leaf) represented as either a polygon or an image into \rvtwo{its} corresponding class.
For example, given an image of the silhouette of an animal, shape classification aims to decide the type of this animal~\citep{bai2009integrating}. Given a building footprint represented as a polygon, cartographers are interested in knowing which type of building it falls into such as E-type, Y-type~\citep{yan2021graph}. 
Further more, given a neighborhood represented as a collection of building polygons as shown in Figure \ref{fig:build_pattern}, we would like to know the type/land use of this neighborhood.

Traditional shape classification models are based on handcrafted or learned shape descriptors. Wang et al. \citep{wang2014bag} proposed a Bag of Contour Fragment (BCF) method by decomposing one shape polygon into contour fragments each of which is described by a shape descriptor. Then a compact shape representation is max-pooled from them based on a spatial pyramid method. 
With the development of deep learning technology, many image-based shape classification models \rvtwo{skip} the image vectorization step and directly \rvtwo{apply} CNN on those input images for shape classification~\citep{atabay2016binary,atabay2016convolutional}. 
Later on, instead of directly applying CNN \reviseone{to} images, Hofer et al. \citep{hofer2017deep} first converted 2D object shapes (images) into topological signatures and \rvtwo{inputted} 
them into a CNN-based model. 
However, several recent work showed that when using deep convolutional networks for object classification, surface texture plays a larger role than shape information. Although CNN models can access some local shape features such as local orientations, they have no sensitivity to the overall shape of objects~\citep{baker2018deep}. This indicates that it is meaningful to develop a polygon encoding model \rvtwo{that is sensitive to shape information} for shape classification.

Table \ref{tab:shape_dataset_stat} provides \reviseone{statistic on multiple existing shape classification datasets}. We can see that except for Yan et al's building dataset~\citep{yan2021graph}, most shape classification datasets provide images as shape samples and are open access. \reviseone{Yan et al~\citep{yan2021graph} created a building shape classification dataset where each building is represented as one simple polygon, not image. But this dataset is not open sourced. }
In addition, most datasets 
are rather \reviseone{small} (less than 5K training samples), which is very challenging for deep learning models. 
For example, according to \rvtwo{Kurnianggoro et al.} \citep{kurnianggoro2018survey}, BCF \citep{wang2014bag}, a feature engineering model, 
is still the state-of-the-art model on MPEG-7 and outperforms all deep learning models. 
\reviseone{Our \mnistcomplex~dataset is based on MNIST. See Section \ref{sec:mnist_data} for a detailed description\rvtwo{, and Figure \ref{fig:shape_cla_necesity_illu} for shape examples}. }

\begin{table}[!ht]
\caption{\reviseone{Statistics of shape classification datasets.}
``\#C'' and ``\#S/C'' indicate the number of categories and the number of samples per category in each dataset.
``$\sim$'' indicates an estimation \rvtwo{of \reviseone{``\#S/C''} for datasets which are not balanced.}
``\#Train'' and ``\#Test'' indicate the number of training and testing samples in each dataset. ``-'' indicates that there is no common agreement on train/test split on this dataset.
``Topic'' \rvtwo{indicates the type of object} each shape sample stands for.
``OA'' indicates whether this dataset is open access.
	}
	\label{tab:shape_dataset_stat}
	\centering
\setlength{\tabcolsep}{2pt}
\begin{tabular}{l|c|c|c|c|p{2.3cm}|p{2.5cm}|l}
\toprule
Dataset         & \#C & \#S/C & \#Train & \#Test & Data Format               & Topic                                          & OA \\ \hline
MPEG-7~\citep{latecki2000shape}          & 70            & 20                     & -           & -          & Silhouette images         & Various objects                                & Yes         \\ \hline
Animal~\citep{bai2009integrating}          & 20            & 100                    & -           & -          & Silhouette images         & Animals                                        & Yes         \\ \hline
Swedish leaf~\citep{soderkvist2001computer}    & 15            & 75                     & -           & -          & Colorful images           & Leaves                                         & Yes         \\ \hline
ETH-80~\citep{leibe2003analyzing}          & 8             & 410                    & 3,239       & 41         & Colorful images           & Various objects & Yes         \\ \hline
100 leaves~\citep{mallah2013plant}      & 100           & 16                     & -           & -          & Colorful images           & Leaves                                         & Yes         \\ \hline
Kimia-216~\citep{sebastian2005curves}       & 18            & 12                     & -           & -          & Silhouette images         & Objects/Animals
& Yes         \\ \hline
MNIST~\citep{lecun1998gradient}           & 10            &  $\sim$7000            & 60,000      & 10,000     & Silhouette images         & Handwritten digits                             & Yes         \\ \hline
Yan et al~\citep{yan2021graph} & 10            & $\sim$775                      & 5,000       & 2,751      & Simple polygons & Building footprints        & No      \\
\bottomrule
\end{tabular}
\end{table}

\subsection{Polygon Encoding} \label{sec:pgon_enc_related}

\reviseone{Following Section \ref{sec:intro}, here, we review several existing work about spatial domain and spectral domain polygon encoders.
}

\reviseone{Spatial domain polygon encoders \citep{veer2018deep,yan2021graph} directly consume polygon vertex coordinates in the spatial domain for polygon encoding. Most of them only consider simple polygons .
Veer et al. \citep{veer2018deep} proposed two spatial domain polygon encoders: a recurrent neural network (RNN) based model and a 1D CNN based model. The RNN model directly feeds the polygon exterior coordinate sequence into a bi-directional LSTM and takes the last state as the polygon embedding. The CNN model feeds the polygon exterior sequence into a series of 1D convolutional layers with zero padding followed by a global average pooling. Veer et al. \citep{veer2018deep} applied both models to three polygon-shape-based tasks: neighbourhood population prediction, building footprint classification, and archaeological ground feature classification.
Results show that the CNN model is better than the RNN model on all three tasks. In this work, the CNN model denoted as \veercnn~ is used as one of our baselines.
}

\reviseone{Another example of spatial domain polygon encoders is the Graph Convolutional AutoEncoder model (GCAE) \citep{yan2021graph}. GCAE learns a polygon embedding for each building footprint represented as a simple polygon in an unsupervised learning manner.
}
The exterior of each building (a simple polygon) \reviseone{is converted to} an undirected \reviseone{weighted} graph in which \reviseone{exterior vertices} are the \reviseone{graph} nodes which are connected by \reviseone{exterior segments (graph edges). 
Each edge is weighted by its length.
Each vertex (node) is associated with a node embedding initialized by some predefined local or regional shape descriptors.
}
The GCAE follows a U-Net \citep{ronneberger2015u} like architecture which uses graph convolution layers and graph pooling \citep{defferrard2016convolutional} in the graph encoder and upscaling layers in the \reviseone{graph} decoder. 
\reviseone{The intermediate representation between the encoder and decoder is the learned polygon embedding. } 
The effectiveness of GCAE is demonstrated qualitatively and quantitatively on shape similarity and shape retrieval task.

\reviseone{Instead of encoding a polygonal geometry directly in the spatial domain, spectral domain polygon encoders first transform it into the spectral space by using Fourier transformation and then design a model to consume these spectral features.
One example is Jiang et al \citep{jiang2019convolutional} which first perform Non-Uniform Fourier Transforms (NUFT) to transform a given polygon geometry (or more generally, 2-/3-simplex meshes) into the spectral domain. And then Jiang et al \citep{jiang2019convolutional} perform a inverse Fast Fourier transformation (IFFT) to convert these spectral features into 2D images or 3D voxels. 
}
The result is an image of the polygonal geometry (or a 3D voxel for a 3D shape) which can be easily consumed by different CNN models such as LeNet5 \citep{lecun1998gradient},  ResNet \citep{he2016deep}, and Deep Layer Aggregation (DLA) \citep{yu2018deep}.
\ddsl~\citep{jiang2019ddsl} further extends this NUFT-IFFT operation into a differentiable layer which is more flexible for shape optimization (through back propagation). 
The effectiveness of DDSL has been shown in shape classification task (MNIST),  3D shape retrieval task, and 3D surface reconstruction task. 
However, the NUFT-IFFT operation is essentially a polygon rasterization approach and sacrifice 
\reviseone{an information loss which depends on the pixel size. As shown in Figure \ref{fig:scale}, when the pixel size is too large, the red polygon can not cover even one \rvtwo{pixel} which might lead to wrong prediction. On the other hand, when the pixel size is too small, the image become unnecessary large and lead to huge computation cost. 
}
\reviseone{Inspired \rvtwo{by} DDSL, our \nuftmlp~model adopts the NUFT idea. Instead of \reviseone{performing} an IFFT, we directly learn the polygon embeddings in the spectral domain.
Without the restriction of IFFT, we have more flexibility in terms of the choice of Fourier frequency map (See Section \ref{sec:nuft_method}).}
So \nuftmlp~ \reviseone{is expected to have lower information loss and a better performance.}
We will discuss the NUFT method in detail in Section \ref{sec:nuftmlp_method}. 
 \vspace{-0.3cm}
\section{Method}   \label{sec:method}
In this section, we first 
present two polygon encoders: \reviseone{\resnetoned~and \nuftmlp. 
Figure \ref{fig:restnet1d_model} and \ref{fig:nuftmlp_model} show the general model architectures of \resnetoned~and \nuftmlp~respectively. 
}
\rvtwo{Then we discuss how these encoders can be used to form shape classification models and spatial relation prediction models.}
We will compare \rvtwo{the encoders} with other baselines and discuss their properties in Section \ref{sec:compare}.
\rvtwo{Finally, we provide proofs for their key properties in Section \ref{sec:proofs}.}

\vspace{-0.3cm}
\subsection{\reviseone{\resnetoned~Encoder}}  \label{sec:resnet1d}

\reviseone{We first propose a spatial domain polygon encoder called \resnetoned~which \rvtwo{uses} a modified 1D ResNet model with circular padding to encode the polygon exterior vertices.
}

\reviseone{
Given a simple polygon $\geom = (\border, \emptyset)$ where $\border = [\ptcoord_{0}^T;\ptcoord_{1}^T;...;\ptcoord_{\ptidx}^T; ...;\ptcoord_{\numpgonpt{}-1}^T] \in \Real^{\numpgonpt{} \times 2}$, \resnetoned~treats the exterior $\border$ of $\geom$ as a 1D coordinate sequence. Before feeding $\border$ into the 1D ResNet layer, we first compute a point embedding $\ptemb_{\ptidx} \in \Real^{4\kdelta+2}$ for the $\ptidx$th point $\ptcoord_{\ptidx}$ by concatenating $\ptcoord_{\ptidx}$ with its spatial affinity with its neighboring $2\kdelta$ points:
\begin{align}
    \ptemb_{\ptidx} = [\ptcoord_{\ptidx}; \ptcoord_{\ptidx-\kdelta} - \ptcoord_{\ptidx}; ...;\ptcoord_{\ptidx-1} - \ptcoord_{\ptidx};\ptcoord_{\ptidx+1} - \ptcoord_{\ptidx};...;\ptcoord_{\ptidx+\kdelta} - \ptcoord_{\ptidx}]
    \label{equ:resnet1d_ptemb}
\end{align}
We call  Equation \ref{equ:resnet1d_ptemb}  \textbf{\kdeltaenc~point encoder}, which  adds  neighborhood structure information into each point embedding and helps to reduce the need to train very deep encoders. 
Here, if $\ptidx-\kdelta < 0$ or $\ptidx+\kdelta \geq \numpgonpt{}$, we get its coordinates by \textit{circular padding} given the fact that $\border$ represents a circle. The resulting embedding matrix $\ptembmat = [\ptemb_0^{T}; \ptemb_1^{T};...; \ptemb_{\ptidx}^{T};...; \ptemb_{\numpgonpt{}-1}^{T}] \in \Real^{\numpgonpt{} \times (4\kdelta+2)}$ is the input of a modified 1D ResNet model which uses \textit{circular padding} instead of zero padding in 1D CNN and max pooling layers to ensure loop origin invariance. The whole \resnetoned~architecture is illustrated as Equation \ref{equ:resnet1d_model}.
\vspace{-0.3cm}
\begin{align}
\begin{split}
    \pgonenc_{\resnetoned}(\geom) = 
    \Big[ \ptembmat &\to \cnnonedlayer_{3\times1}^{\pgonembdim,1,1} \to \batchnormonedlayer \to \relu \to \maxpoolonedlayer_{2\times1}^{2,0} \\ &
    \to \big( \resnetonedlayer \big)_{\resnetonedlayeridx=1}^{\resnetonedlayernum} 
    \to \globalmaxpoolonedlayer \to \dropout \to \pgonemb \Big]
\end{split}
    \label{equ:resnet1d_model}
\end{align}
$\cnnonedlayer_{3\times1}^{\pgonembdim,1,1}$ indicates a 1D CNN layer with 1 stride, 1 padding (circular padding) and $\pgonembdim$ number of $3\times1$ kernel (1D kernel). $\batchnormonedlayer$ and $\relu$ indicate 1D batch normalization layer and a ReLU activation layer. $\maxpoolonedlayer_{2\times1}^{2,0}$ indicates a 1D Max Pooling layer with 2 stride, 0 padding, and kernel size $2$. $\big( \resnetonedlayer \big)_{\resnetonedlayeridx=1}^{\resnetonedlayernum}$ indicates $\resnetonedlayernum$ standard 1D ResNet layers \rvtwo{with circular padding}. $\globalmaxpoolonedlayer$ and $\dropout$ are a global max pooling layer and dropout layer. The final output $\pgonenc_{\resnetoned}(\geom) = \pgonemb \in \Real^{\pgonembdim}$ is the polygon embedding of the simple polygon $\geom$.}

\subsection{\nuftmlp~\rvtwo{Encoder}}  \label{sec:nuftmlp_method}
As shown in Figure \ref{fig:nuftmlp_model}, \nuftmlp~first applies Non-Uniform Fourier Transforms (NUFT) to convert a polygonal geometry $\geom$ into the spectral domain. Then it directly feeds these spectral features into a multi-layer perceptron to obtain the polygon embedding $\pgonemb$ of $\geom$. Polygonal geometry $\geom$ can be either a polygon with/without holes, or a multipolygon. 

\subsubsection{Converting Polygon Geometries to $\simplexdim$-Simplex Meshes} \label{sec:nuft_convert_simplex}
\begin{figure*}[ht!]
	\centering \tiny
	\vspace*{-0.4cm}
		\centering 
		\includegraphics[width=1.0\textwidth]{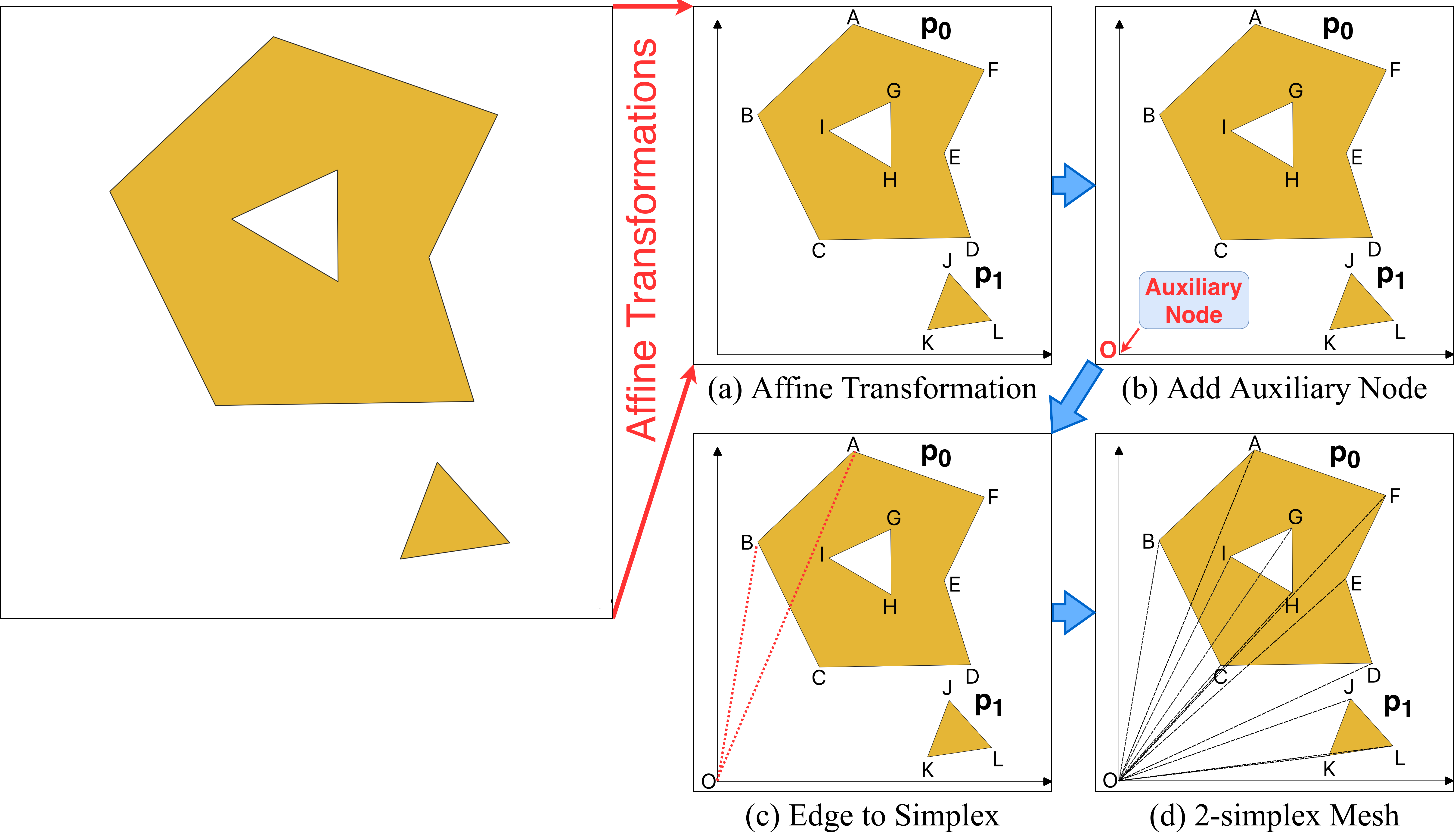}
	\caption{\reviseone{An illustration of the auxiliary node method.
	(a) Applying a series of affine transformations to a polygonal geometry $\geom$;
	(b) Adding auxiliary node ``O'';
	(c) Converting each polygon edge to a 2-simplex;
	(d) Constructing a 2-simplex mesh $\simplexmesh^{(\simplexdim)}$.}
	} 
	\label{fig:aux_node}
	\vspace{-0.7cm}
\end{figure*}

Following DDSL's \textit{auxiliary node method}~\citep{jiang2019ddsl}, we first convert a given polygonal geometry $\geom$ into a $\simplexdim$-simplex mesh $\simplexmesh^{(\simplexdim)} = \{\simplex^{(\simplexdim)}_\simplexidx\}_{\simplexidx=1}^{\numpgonpt{}} = (\vtxmat, \edgmat, \desmat)$ (here $\simplexdim = 2$) by adding one auxiliary node (the origin point $\ptcoord_O$). A $2$-simplex is simply a triangle in the 2D space.
The polygon-to-$\simplexdim$-simplex-mesh operation \reviseone{is illustrated in Figure \ref{fig:aux_node} and summarized as below:}
\begin{enumerate}
    \item \reviseone{As shown in Figure \ref{fig:aux_node}a, }we first apply a series of affine transformations including scaling and translation to transform $\geom$ into a unit space. 
    First, a translation operation is applied to move $\geom$ such that the center of its bounding box is the origin point $\ptcoord_O = [0,0]$. 
    Then, a scaling operator is applied to scale $\geom$ into the unit space $[-1,1] \times [-1,1]$.
Finally, another translation operation is used to move $\geom$ into the space $[0,2] \times [0,2]$, since positive coordinates are required for NUFT.
These transformation operations are used to allow each polygon lays in the same \reviseone{relative} space which is critical for neural network learning. 
Polygon encoding mainly aims at encoding the shape of each polygon. The global position of $\geom$ (e.g., the real geographic coordinates of each vertex of California's polygon) should be encoded seperately through location encoding method~\citep{mai2020multiscale,mai2021review}. If we would like to do spatial relation prediction between a polygon pair, they should be transformed into the same unit space $[0,2] \times [0,2]$.
    \item We add the origin point $\ptcoord_O = [0,0]$ as an \textit{auxiliary node} which helps us to convert Polygonal Geometry $\geom$ into $\simplexdim$-simplex mesh $\simplexmesh^{(\simplexdim)}$ \reviseone{(See Figure \ref{fig:aux_node}b)}.
    \item We loop through all edges in  $\{\edge_\simplexidx\}_{\simplexidx=1}^{\numpgonpt{}}$ (See Definition \ref{def:pgon_geometry}) of $\geom$. For the $\simplexidx$th edge $\edge_\simplexidx$, we connect its two vertices to the auxiliary node $\ptcoord_O$ (origin) to construct a 2-simplex (triangle) $\simplex^{(\simplexdim)}_\simplexidx$. For instance, for Edge $\edge_{AB}$, we construction a Triangle $\simplex^{(\simplexdim)}_{ABO} = \triangle_{ABO}$ \reviseone{(See Figure \ref{fig:aux_node}c)}.
    \item For a polygonal geometry $\geom$ with in total $\numpgonpt{}$ vertices, 
we can get $\numpgonpt{}$ 2-simplexes which form a 2-simplex mesh 
\reviseone{$\simplexmesh^{(\simplexdim)} = \{\simplex^{(\simplexdim)}_\simplexidx\}_{\simplexidx=1}^{\numpgonpt{}}$ (See Figure \ref{fig:aux_node}d)}.
\end{enumerate}

A 2-simplex mesh $\simplexmesh^{(\simplexdim)} = \{\simplex^{(\simplexdim)}_\simplexidx\}_{\simplexidx=1}^{\numpgonpt{}} = (\vtxmat, \edgmat, \desmat)$ is represented as three matrices - vertex matrix $\vtxmat \in \Real^{(\numpgonpt{}+1) \times 2}$ for polygon/simplex vertex coordinates, edge matrix $\edgmat \in \NonNegInt^{\numpgonpt{} \times 3}$ for 2-simplex connectivity, and density matrix $\desmat \in \Real^{\numpgonpt{} \times \desdim}$ for per-simplex density. 
$\vtxmat$ is a float matrix that contains the coordinates of all vertices of $\geom$ 
as well as the origin $\ptcoord_O = [0,0]$ (the last row). 
The edge matrix $\edgmat$ contains non-negative integers. The $\simplexidx$th row of $\edgmat$ corresponds to the $\simplexidx$th simplex $\simplex^{(\simplexdim)}_{\simplexidx}$ in $\simplexmesh^{(\simplexdim)}$ whose values indicate the indices of vertices of $\simplex^{(\simplexdim)}_{\simplexidx}$ in $\vtxmat$.
For the density matrix $\desmat \in \Real^{\numpgonpt{} \times \desdim}$, the $\simplexidx$th row indicates a $\desdim$-dimensional density features associated with the $\simplexidx$th edge/simplex (e.g., edge color, density, and etc.).
In our case, we assume each edge $\edge_\simplexidx$ has a constant 1D density feature -- $[1]$, i.e., $\desmat = \onemat \in \Real^{\numpgonpt{} \times 1}$, a constant one matrix.

We should make sure that the boundary of $\geom$ is oriented correctly by following Definition \ref{def:pgon} - the exteriors of all sub-polygons should be oriented in a \rvtwo{counterclockwise}
fashion while all interiors should be oriented in a clockwise fashion. This makes sure that we can use the \textit{right-hand rule}\footnote{\url{https://mapster.me/right-hand-rule-geojson-fixer/}} to infer the correct orientation of each edge $\edge_\simplexidx$. So the orientation of the boundary of the corresponding simplex $\simplex^{(\simplexdim)}_{\simplexidx}$ can be determined (e.g., counterclockwise or clockwise).
Based on that, we can compute the signed content (area) of each simplex $\simplex^{(\simplexdim)}_{\simplexidx}$ so that the topology of $\geom$ is preserved - \textit{topology awareness}.

To concretely show how to convert a polygonal geometry $\geom$ into a 2-simple mesh $\simplex^{(2)} = (\vtxmat, \edgmat, \desmat)$, we use the example of Multipolygon $\geom$ shown in Figure \ref{fig:nuftmlp_model}. It can be converted into Simplex $\simplex^{(2)} = (\vtxmat, \edgmat, \desmat)$ where $\vtxmat = [\ptcoord_{A}^T;\ptcoord_{B}^T;\ptcoord_{C}^T;\ptcoord_{D}^T;\ptcoord_{E}^T;\ptcoord_{F}^T;\ptcoord_{G}^T;\ptcoord_{H}^T;\ptcoord_{I}^T;\ptcoord_{J}^T;\ptcoord_{K}^T;\ptcoord_{L}^T;\ptcoord_{O}^T] \in \Real^{13 \times 2}$, $\edgmat = [[0,1,12], [1,2,12],[2,3,12],[3,4,12],[4,5,12],[5,0,12],[6,7,12], [7,8,12], $ $
[8,6,12], [9,10,12],$ $[10,11,12],[11,9,12]] \in \NonNegInt^{12 \times 3}$, and $\desmat$ is a $12 \times 1$ constant one matrix.

\subsubsection{Non-Uniform Fourier Transforms} \label{sec:nuft_method}
Next, we perform NUFT on this $\simplexdim$-simplex mesh 
$\simplexmesh^{(\simplexdim)} = \{\simplex^{(\simplexdim)}_\simplexidx\}_{\simplexidx=1}^{\numpgonpt{}}$
(here $\simplexdim = 2$). 
Compared with the conventional Discrete Fourier transform (DFT) whose input signal is sampled at equally spaced points or frequencies (or both), NUFT can deal with input signal sampled at non-equally spaced points or transform the input into non-equally spaced frequencies. This makes NUFT very suitable for irregular structured data such as point cloud, line meshes, polygonal geometries, and so on \citep{jiang2019convolutional,jiang2019ddsl}. In contrast, DFT is more suitable for regular structured data such as images, videos \citep{rippel2015spectral}.

\vspace{-0.5cm}
\begin{definition}[Density Function on $\simplexdim$-simplex] \label{def:simplex_func}
    For the $\simplexidx$th $\simplexdim$-simplex $\simplex^{(\simplexdim)}_{\simplexidx}$
in a $\simplexdim$-simplex mesh $\simplexmesh^{(\simplexdim)} = \{\simplex^{(\simplexdim)}_\simplexidx\}_{\simplexidx=1}^{\numpgonpt{}} = (\vtxmat, \edgmat, \desmat)$, we define a density function $\pcf_{\simplexidx}^{(\simplexdim)}(\ptcoord)$ on $\simplex^{(\simplexdim)}_{\simplexidx}$ as Equation \ref{equ:density_func} in which $\desfunc_{\simplexidx}$ is the signal density defined on $\simplex^{(\simplexdim)}_{\simplexidx}$.  \begin{align}
        \pcf_{\simplexidx}^{(\simplexdim)}(\ptcoord) = 
        \begin{cases}
        \desfunc_{\simplexidx}, & \ptcoord \in \simplex^{(\simplexdim)}_{\simplexidx} \\
        0, & \ptcoord \notin \simplex^{(\simplexdim)}_{\simplexidx}
        \end{cases}
        \label{equ:density_func}
    \end{align}

\end{definition}
\vspace{-0.5cm}

\vspace{-0.5cm}
\begin{definition}[Piecewise-Constant Function over \rvtwo{a} simplex mesh] \label{def:pcf}
The Piecewise-Constant Function (PCF) over \rvtwo{a} simplex mesh $\simplexmesh^{(\simplexdim)}$ is the superposition of the density function $\pcf_{\simplexidx}^{(\simplexdim)}(\ptcoord)$ for each simplex $\simplex^{(\simplexdim)}_{\simplexidx}$:
\begin{align}
    \pcf_{\simplexmesh}^{(\simplexdim)}(\ptcoord) = \sum_{\simplexidx=1}^{\numpgonpt{}} \pcf_{\simplexidx}^{(\simplexdim)}(\ptcoord)
    \label{equ:pcf}
\end{align}
\end{definition}
\vspace{-0.5cm}

\vspace{-0.5cm}
\begin{definition}[NUFT of PCF $\pcf_{\simplexmesh}^{(\simplexdim)}(\ptcoord)$] \label{def:nuftpcf}
The NUFT of PCF $\pcf_{\simplexmesh}^{(\simplexdim)}(\ptcoord)$ over a $\simplexdim$-simplex mesh $\simplexmesh^{(\simplexdim)} = \{\simplex^{(\simplexdim)}_\simplexidx\}_{\simplexidx=1}^{\numpgonpt{}} = (\vtxmat, \edgmat, \desmat)$ on a set of $\fftfreqnum$ Fourier base frequencies $\fftfreqmat = \{\fftfreq_{\fftfreqidx}\}_{\fftfreqidx=1}^{\fftfreqnum}$ is a sequence of $\fftfreqnum$ complex numbers:

\vspace{-0.5cm}
\begin{align}
    \fftpcf_{\simplexmesh}^{(\simplexdim)}(\ptcoord) = [\fftpcf_{\simplexmesh,1}^{(\simplexdim)}(\ptcoord), \fftpcf_{\simplexmesh,2}^{(\simplexdim)}(\ptcoord),..., \fftpcf_{\simplexmesh,\fftfreqidx}^{(\simplexdim)}(\ptcoord),...,
    \fftpcf_{\simplexmesh,\fftfreqnum}^{(\simplexdim)}(\ptcoord)
    ]
    \label{equ:nuftpcfall}
\end{align}

where the NUFT of $\pcf_{\simplexmesh}^{(\simplexdim)}(\ptcoord)$ on each base frequency $\fftfreq_{\fftfreqidx} \in \Real^{2}$ can be written as the weighted sum of the Fourier transform on each $\simplexdim$-simplex $\simplex^{(\simplexdim)}_\simplexidx$:

\vspace{-0.2cm}
\begin{align}
\begin{split}
    \fftpcf_{\simplexmesh,\fftfreqidx}^{(\simplexdim)}(\ptcoord) 
    & = \idotsint_{-\infty}^{\infty} \pcf_{\simplexmesh}^{(\simplexdim)}(\ptcoord) \expbase^{-i\langle \fftfreq_{\fftfreqidx}, \ptcoord \rangle} d\ptcoord \\
    & = \sum_{\simplexidx=1}^{\numpgonpt{}} \desfunc_{\simplexidx} \idotsint_{\simplex^{(\simplexdim)}_\simplexidx} \expbase^{-i\langle \fftfreq_{\fftfreqidx}, \ptcoord \rangle} d\ptcoord  \\
    & = \sum_{\simplexidx=1}^{\numpgonpt{}} \desfunc_{\simplexidx} \fftpcf_{\simplexidx,\fftfreqidx}^{(\simplexdim)}(\ptcoord)
\end{split}
    \label{equ:nuftpcf}
\end{align}
\end{definition}
\vspace{-0.5cm}

$\fftpcf_{\simplexidx,\fftfreqidx}^{(\simplexdim)}(\ptcoord)$ is the NUFT on the $\simplexidx$th simplex $\simplex^{(\simplexdim)}_\simplexidx$ with base frequency $\fftfreq_{\fftfreqidx}$ which can be defined as follow:

\vspace{-0.4cm}
\begin{align}
    \fftpcf_{\simplexidx,\fftfreqidx}^{(\simplexdim)}(\ptcoord) = 
    i^{\simplexdim} \simplexsign(\simplex^{(\simplexdim)}_\simplexidx) \simplexcontratio^{(\simplexdim)}_\simplexidx \sum_{t=1}^{\simplexdim+1} \dfrac{\expbase^{-i \langle \fftfreq_{\fftfreqidx}, \ptcoord_t \rangle} }{\prod_{l=1}^{\simplexdim+1} ( \expbase^{-i \langle \fftfreq_{\fftfreqidx}, \ptcoord_t \rangle} - \expbase^{-i \langle \fftfreq_{\fftfreqidx}, \ptcoord_l \rangle} )},
    \label{equ:nuftpcf_simplex}
\end{align}
\vspace{-0.4cm}

where $\simplexsign(\cdot): \simplexmesh^{(\simplexdim)} \to \{-1, 1\}$ is a sign function which determines the sign of the content (area) of $\simplex^{(\simplexdim)}_\simplexidx$. $\simplexcontratio^{(\simplexdim)}_\simplexidx$ is the content distortion factor, which is the ratio of the unsigned content of $\simplex^{(\simplexdim)}_\simplexidx$ - $\simplexcont^{(\simplexdim)}_\simplexidx$ - and the content of the unit orthogonal $\simplexdim$-simplex -  $\simplexcont^{(\simplexdim)}_{I} = 1/\simplexdim!$. 
$\simplexsign(\simplex^{(\simplexdim)}_\simplexidx) \simplexcontratio^{(\simplexdim)}_\simplexidx$ is the signed content distortion factor of $\simplex^{(\simplexdim)}_\simplexidx$ which 
can be computed based on the determinant of the Jacobian matrix\footnote{\url{https://en.m.wikipedia.org/wiki/Simplex\#Volume}} $\jacobmat_{\simplexidx}$ of simplex $\simplex^{(\simplexdim)}_\simplexidx$. \reviseone{Let $\ptcoord_{\simplexidx,1},\ptcoord_{\simplexidx,2},\ptcoord_{O}$ the three vertices of $\simplex^{(\simplexdim)}_\simplexidx$,} we have:

\vspace{-0.4cm}
\begin{align}
    \begin{split}
     \simplexsign(\simplex^{(\simplexdim)}_\simplexidx) \simplexcontratio^{(\simplexdim)}_\simplexidx 
     & = \dfrac{\simplexsign(\simplex^{(\simplexdim)}_\simplexidx)   \simplexcont^{(\simplexdim)}_\simplexidx}{\simplexcont^{(\simplexdim)}_{I}}
     = \dfrac{1/\simplexdim! \det(\jacobmat_{\simplexidx})}{1/\simplexdim!} \\
& =\det([\ptcoord_{\simplexidx,1} - \ptcoord_{O},\ptcoord_{\simplexidx,2}- \ptcoord_{O}]) = \det([\ptcoord_{\simplexidx,1},\ptcoord_{\simplexidx,2}])
     \end{split}
    \label{equ:simplex_content_ratio}
\end{align}
\vspace{-0.4cm}

The proof of Equation \ref{equ:nuftpcf}, \ref{equ:nuftpcf_simplex}, and \ref{equ:simplex_content_ratio} can be found in \citep{jiang2019convolutional}.

Note that in Definition \ref{def:nuftpcf}, we have not specified the way in which we select the set of $\fftfreqnum$ Fourier base frequencies $\fftfreqmat = \{\fftfreq_{\fftfreqidx}\}_{\fftfreqidx=1}^{\fftfreqnum}$ where $\fftfreq_{\fftfreqidx} \in \Real^{2}$. 
We can either use a series of equally spaced frequencies along each dimension, denoted as \textit{linear grid frequency map $\fftfreqmat^{(\linearfreqmtd)}$}, or a series of non-equally spaced frequencies along each dimension. For the second option, in this work, we choose to use a geometric series as the Fourier frequencies in the X and Y dimension, denoted as \textit{geometric grid frequency map $\fftfreqmat^{(\linearfreqmtd)}$}, which has been widely used in many location encoding literature~\citep{mai2020multiscale,mildenhall2020nerf,tancik2020fourier,mai2021review} and Transformer architecture~\citep{vaswani2017attention}. 
We formally define these two Fourier frequency maps as below.

\begin{definition}[Linear Grid Frequency Map]  \label{def:fft_freq}
The linear grid frequency map $\fftfreqmat^{(\linearfreqmtd)}$ is just simply the normal \textit{integer} Fourier frequency bases used by the Fast Fourier Transform (FFT). It is defined as a Cartesian product between the linear frequency sets along the X and Y axis -- $\fftfreqvec_{x}^{(\linearfreqmtd)}$ and $\fftfreqvec_{y}^{(\linearfreqmtd)}$ -- which contain $\fftfreqnumX$ and $\fftfreqnumY$ integer values respectively:
\begin{align}
    \fftfreqmat^{(\linearfreqmtd)} = \fftfreqvec_{x}^{(\linearfreqmtd)} \times \fftfreqvec_{y}^{(\linearfreqmtd)}  \in \Int^{\fftfreqnumX} \times \Int^{\fftfreqnumY}
    \label{equ:fftfreq_fft}
\end{align}
Here, $\fftfreqnum = \fftfreqnumX\fftfreqnumY$. $\fftfreqvec_{x}^{(\linearfreqmtd)}$ and $\fftfreqvec_{y}^{(\linearfreqmtd)}$ are defined as Equation \ref{equ:fftfreq_fft_w}.
\begin{align}
    \begin{split}
    \fftfreqvec_{x}^{(\linearfreqmtd)} & = 
    \begin{cases}
     \{-\nscale,...,-1,0,1,\nscale\}, \; & if \; \fftfreqnumX = 2\nscale + 1 \\
    \{-\nscale,...,-1,0,1,\nscale-1\}, \; & if \; \fftfreqnumX = 2\nscale
    \end{cases}
    \\
    \fftfreqvec_{y}^{(\linearfreqmtd)} & = 
    \begin{cases}
     \{0,1,\nscale\}, \; & if \; \fftfreqnumY = \nscale + 1 \\
    \{0,1,\nscale-1\}, \; & if \; \fftfreqnumY = \nscale
    \end{cases}
    \end{split}
    \label{equ:fftfreq_fft_w}
\end{align}
Where $\nscale \in \Int^{+}$ is \reviseone{half}
number of frequencies we use which decides $\fftfreqnumX$ and $\fftfreqnumY$. Note that a normal practice of FFT is to use half frequency bases in the last dimension. So here $\fftfreqvec_{y}^{(\linearfreqmtd)}$ has roughly \reviseone{half of} $\fftfreqvec_{x}^{(\linearfreqmtd)}$'s frequencies
\end{definition}

\begin{definition}[Geometric Grid Frequency Map] \label{def:geometric_freq}
Since polygonal geometries are non-uniform signals, we do not need to use the normal integer FFT frequency map $\fftfreqmat^{(\linearfreqmtd)}$. Instead, we can use real value\rvtwo{s} as Fourier frequency bases to increase the data variance we can capture.
By following this idea, the geometric grid frequency map $\fftfreqmat^{(\geometricfreqmtd)}$ is defined as a Cartesian product between the selected geometric series based frequency sets along the X and Y axis -- $\fftfreqvec_{x}^{(\geometricfreqmtd)}$ and $\fftfreqvec_{y}^{(\geometricfreqmtd)}$: \begin{align}
    \fftfreqmat^{(\geometricfreqmtd)} = \fftfreqvec_{x}^{(\geometricfreqmtd)} \times \fftfreqvec_{y}^{(\geometricfreqmtd)}  \in \Real^{\fftfreqnumX} \times \Real^{\fftfreqnumY}
    \label{equ:fftfreq_geom}
\end{align}
Here, $\fftfreqvec_{x}^{(\geometricfreqmtd)}$ and $\fftfreqvec_{y}^{(\geometricfreqmtd)}$ are defined as Equation \ref{equ:fftfreq_geom_w}.
\vspace{-0.1cm}
\begin{align}
    \begin{split}
    \fftfreqvec_{x}^{(\geometricfreqmtd)} & = 
    \begin{cases}
     \{-\fftfreqitem{\certainscale}\}_{\certainscale=0}^{\nscale-1}
     \cup \{0\} \cup 
     \{\fftfreqitem{\certainscale}\}_{\certainscale=0}^{\nscale-1}, \; & if \; \fftfreqnumX = 2\nscale + 1 \\
    \{-\fftfreqitem{\certainscale}\}_{\certainscale=0}^{\nscale-1}
     \cup \{0\} \cup 
     \{\fftfreqitem{\certainscale}\}_{\certainscale=0}^{\nscale-2}, \; & if \; \fftfreqnumX = 2\nscale
    \end{cases}
    \\
    \fftfreqvec_{y}^{(\geometricfreqmtd)} & = 
    \begin{cases}
     \{0\} \cup 
     \{\fftfreqitem{\certainscale}\}_{\certainscale=0}^{\nscale-1}, \; & if \; \fftfreqnumY = \nscale + 1 \\
    \{0\} \cup 
     \{\fftfreqitem{\certainscale}\}_{\certainscale=0}^{\nscale-2}, \; & if \; \fftfreqnumY = \nscale
    \end{cases}
    \end{split}
    \label{equ:fftfreq_geom_w}
\end{align}
Here, $\fftfreqvec_{y}^{(\geometricfreqmtd)}$ also has roughly a half of $\fftfreqvec_{x}^{(\geometricfreqmtd)}$'s frequencies. In Equation \ref{equ:fftfreq_geom_w}, $\{\fftfreqitem{\certainscale}\}_{\certainscale=0}^{\nscale-1} = \{ \fftfreqitem{0}=\minscale, \fftfreqitem{1},...,\fftfreqitem{\certainscale},...,\fftfreqitem{\nscale-1}=\maxscale \}$ is defined as a geometric series, where
\begin{align}
\fftfreqitem{\certainscale}  = \minscale \cdot \scaleratio^{\certainscale/(\nscale-1)} ;  \;
where \; \scaleratio  = \frac{\maxscale}{\minscale} ;  \;
\certainscale \in \{0,1,...,\nscale-1\}
\label{equ:fftfreq_geom_def}
\end{align}
$\minscale, \maxscale \in \Real^{+}$ are the \reviseone{minimum and maximum frequency} (hyperparameters). \end{definition}

\begin{figure*}[ht!]
	\centering \tiny
	\vspace*{-0.2cm}
	\begin{subfigure}[b]{0.495\textwidth}  
		\centering 
		\includegraphics[width=\textwidth]{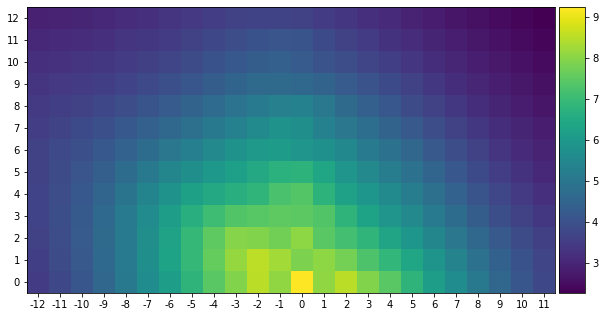}\vspace*{-0.2cm}
		\caption[]{{Linear grid $\fftfreqmat^{(\linearfreqmtd)}$
		}}    
		\label{fig:fft_freq}
	\end{subfigure}
	\hfill
	\begin{subfigure}[b]{0.495\textwidth}  
		\centering 
		\includegraphics[width=\textwidth]{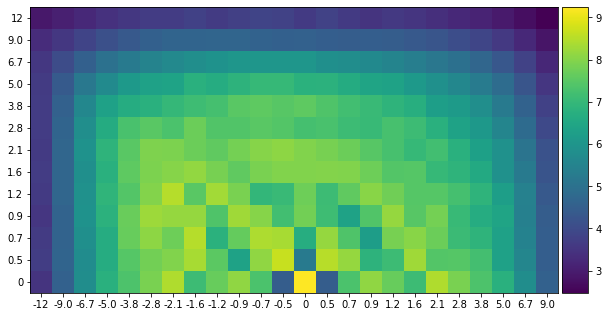}\vspace*{-0.2cm}
		\caption[]{{Geometric grid $\fftfreqmat^{(\geometricfreqmtd)}$
		}}    
		\label{fig:geometric_freq}
	\end{subfigure}
	\caption{An illustration of the full set of the linear grid frequency map $\fftfreqmat^{(\linearfreqmtd)}$ and geometric grid frequency map $\fftfreqmat^{(\geometricfreqmtd)}$. 
} 
	\label{fig:freq_map}
\end{figure*}

A visualization of $\fftfreqmat^{(\linearfreqmtd)}$ and $\fftfreqmat^{(\geometricfreqmtd)}$ can be seen in Figure \ref{fig:fft_freq} and \ref{fig:geometric_freq}.
To investigate the effect of NUFT frequency base selection on the effectiveness of polygon embedding, we compute the data variance on each Fourier frequency base across all 60K training polygon samples in \mnistcomplex~training set.
More specifically, for each polygonal geometry $\geom_i$ in the \mnistcomplex~training set, we first perform NUFT. Each $\geom_i$ yield $\fftfreqnum = \fftfreqnumX \fftfreqnumY$ complex valued NUFT features each of which corresponds to one frequency item $\fftfreq_{\fftfreqidx} \in \Real^{2}$ in $\fftfreqmat^{(\linearfreqmtd)}$ or $\fftfreqmat^{(\geometricfreqmtd)}$. We compute the data variance\footnote{We only compute the data variance for the real value part for each NUFT complex feature.} of each $\fftfreq_{\fftfreqidx}$ across all 60K training polygons in \mnistcomplex~and \rvtwo{visualize them} 
in Figure \ref{fig:fft_freq} and \ref{fig:geometric_freq}. We can see that for linear grid frequency map $\fftfreqmat^{(\linearfreqmtd)}$, most of the data variance is captured by the low frequency components while the high frequencies are less informative. When we switch to the geometric grid frequency map $\fftfreqmat^{(\geometricfreqmtd)}$, \reviseone{more} frequencies have higher data variance which is easier for the following MLP to learn from.

\subsubsection{NUFT-based Polygon Encoding} \label{sec:nuftspec}
Finally, we can define the $\nuftmlp$ polygon encoder $\pgonenc_{\nuftmlp}(\geom)$ as:

\vspace{-0.5cm}
\begin{align}
    \pgonenc_{\nuftmlp}(\geom) = \nuftspecmlpfunc(\nuftspecnorm \big(\fftpcf_{\simplexmesh}^{(\simplexdim)}(\ptcoord) \big))
    \label{equ:nuftmlp_pgon_enc}
\end{align}
Here, $\nuftspecmlpfunc(\cdot)$ is a $\nuftspecmlplayernum$ layer multi-layer perceptron in which each layer is a linear layer followed by a nonlinearity (e.g., ReLU), a skip connection, and a layer normalization layer \citep{ba2016layer}. 
$\nuftspecnorm(\cdot)$ first extract a $2\fftfreqnum$ dimension real value vector from $\fftpcf_{\simplexmesh}^{(\simplexdim)}(\ptcoord)$ and 
\reviseone{then normalize this spectral representation, e.g., L2, batch normalization.}

\vspace{-0.3cm}
\subsection{Shape Classification Model} \label{sec:shape_cla_model}
Given a polygonal geometry $\geom$, we encode it with a polygon encoder $\pgonenc(\cdot)$ followed by a \reviseone{multilayer perceptron (MLP) $\shapeclamlpfunc(\cdot)$} and a softmax layer to predict its shape class $\shapecla$:
\vspace{-0.3cm}
\begin{align}
\rvtwo{P}(\relat \mid \geom) 
    = \softmax(\shapeclamlpfunc(\pgonenc(\geom))).
    \label{equ:shape_cla_model}
\end{align}
Here $\pgonenc(\cdot)$ can be \reviseone{any baseline or our proposed encoders and $\softmax(z_i) = e^{z_i}/\sum_{k=1}^{K} e^{z_k}$.}

\vspace{-0.3cm}
\subsection{Spatial Relation Prediction Model} \label{sec:spa_rel_model}
Given a polygon pair $(\geom_{\subject}, \geom_{\object})$, the spatial relation prediction task aims at predicting a spatial relation between them such as topological relations, cardinal direction relations, and so on. In this work, we adopt a MLP-based spatial relation prediction model
\begin{align}
\rvtwo{P}(\relat \mid \geom_{\subject}, \geom_{\object}) 
    = \softmax(\sparelmlpfunc([\pgonenc(\geom_{\subject}); \pgonenc(\geom_{\object})])),
    \label{equ:spa_rel}
\end{align}
where $[\pgonenc(\geom_{\subject}); \pgonenc(\geom_{\object})] \in \Real^{2\pgonembdim}$ indicates the concatenation of the polygon embeddings of $\geom_{\subject}$ and $\geom_{\object}$. $\sparelmlpfunc(\cdot)$ takes this as input and outputs raw logits over all $\relnum$ possible spatial relations. $\softmax(\cdot)$ normalizes it into a probability distribution 
\rvtwo{$P(\relat \mid \geom_{\subject}, \geom_{\object})$ }
over $\relnum$ relations. %

\subsection{Model Property Comparison}  \label{sec:compare}

\begin{theorem}
	\label{the:resnetoned_proof}
	\resnetoned~is loop origin invariant for simple polygons $\geom = (\border, \emptyset)$.
	\vspace{-0.4cm}
\end{theorem}

\begin{theorem}
	\label{the:nuftmlp_proof}
	\nuftmlp~is (1) loop origin invariant, (2) trivial vertex invariant, (3) part permutation invariant, and (4) topology aware for any polygonal geometry $\geom$.
	\vspace{-0.4cm}
\end{theorem}

Now we compare different polygon encoders discussed in Section \ref{sec:related}  as well as 
\resnetoned~and \nuftmlp~based on\reviseone{: 
1) their encoding capabilities -- whether it can handle holes and multipolygons; 2) four polygon encoding properties (See Section \ref{sec:prob_stat}). }

\reviseone{By definition, \veercnn, \gcae, and \resnetoned~can not handle polygons with holes nor multipolygons. 
To allow these three models handle complex polygonal geometries, we need to concatenate the vertex sequences of different sub-polygons' exteriors and interiors. After doing that, the feed-in order of sub-polygons will affect the encoding results and the topological relations between exterior rings and interiors are lost. 
So none of them satisfy the part permutation invariance and topology awareness. 
Since \veercnn~consumes the polygon exterior as a coordinate sequence and encodes it with zero padding CNN layers, 
the origin vertex and the length of the sequence will affect its encoding results. So it is not loop origin invariant nor trivial vertex invariant. 
Both \gcae~and \resnetoned~are sensitive to trivial vertices which means they are not trivial vertex invariance.
However, \gcae~and \resnetoned~are both loop origin invariant when encoding simple polygons.
\gcae~achieves this by representing the polygon exterior as an undirected graph where no origin is defined. \resnetoned~achieves this 
by using circular padding in each 1D CNN and max pooling layer.
However, this property can not be held for \gcae~and \resnetoned~if the input is complex polygonal geometries. We use ``Yes*'' in Table \ref{tab:pgon_enc_property_compare}.
Only \ddsl~\citep{jiang2019convolutional,jiang2019ddsl} and our \nuftmlp~can handle polygons with holes and multipolygons. They also satisfy all four properties. This is because the nature of NUFT.
\rvtwo{For \resnetoned~ and \nuftmlp, we declare Theorem \ref{the:resnetoned_proof} and \ref{the:nuftmlp_proof}  
whose proofs can be seen in Section \ref{sec:resnetoned_proof} and \ref{sec:nuftmlp_proof}.}
}
Table \ref{tab:pgon_enc_property_compare} shows the full comparison result.

Except for \gcae~\citep{yan2021graph}, the performance of all other polygon encoders listed in Table \ref{tab:pgon_enc_property_compare} are compared and analyzed on the shape classification (Section \ref{sec:exp_shape_clas}) and spatial relation prediction (Section \ref{sec:exp_spa_rel}) task. We do not include \gcae~\reviseone{in} our baseline models since \reviseone{its implementation} is not open sourced.

\vspace{-0.5cm}
\begin{table}[h!]
\caption{The comparison among different polygon encoders by properties such as whether it can handle polygon with holes \reviseone{or} multipolygons as well as the four polygon encoding properties we discuss in Section \ref{sec:prob_stat}.
``Yes*'' indicates that loop \reviseone{origin} invariance only holds for \gcae~and \resnetoned~if the input is simple polygons.
	}
	\label{tab:pgon_enc_property_compare}
	\centering
\reviseone{
\begin{tabular}{l|c|c|c|c|c|c|c}
\toprule
Property      & Type & Holes & Multipolygons & Loop & TriV & ParP & Topo \\ \hline
\veercnn~\citep{veer2018deep}       & Spatial & No                 & No            & No              & No                       & No                          & No                 \\ 
\gcae~\citep{yan2021graph}          & Spatial & No                 & No            & Yes*             & No                       & No                          & No                 \\ 
\ddsl~\citep{jiang2019convolutional,jiang2019ddsl}     & Spectral & Yes                & Yes           & Yes             & Yes                      & Yes                         & Yes                \\ \hline
\resnetoned    & Spatial & No                 & No            & Yes*             & No                       & No                          & No                 \\ 
\nuftmlp & Spectral & Yes                & Yes           & Yes             & Yes                      & Yes                         & Yes          \\
\bottomrule
\end{tabular}
}
\end{table}

\subsection{\rvtwo{Theoretical Proofs of the Model Properties}} \label{sec:proofs}
\subsubsection{\rvtwo{Proofs of Theorem \ref{the:resnetoned_proof}}}  \label{sec:resnetoned_proof}

\rvtwo{
In \resnetoned, circular padding is used in the convolution layers with stride 1. Given a polygon $\pgon = (\border, \holeset = \emptyset)$, circular padding wraps the vector $\border$ on one end around to the other end to provide the
missing values in the convolution computations near the boundary. Thus, $\cnnonedlayer_{3\times1}^{\pgonembdim,1,1} (\loopmat_{\loopdelta}\border)=\loopmat_{\loopdelta} \cnnonedlayer_{3\times1}^{\pgonembdim,1,1} (\border)$ for any input $\border$. Max pooling layer with stride 1 and circular padding has the similar property. $\maxpoolonedlayer_{2\times1}^{1,1}(\loopmat_{\loopdelta}\border)=\loopmat_{\loopdelta}(\maxpoolonedlayer_{2\times1}^{1,1}\border)$. Trivially, $\batchnormonedlayer(\loopmat_{\loopdelta}\border)=\loopmat_{\loopdelta}\batchnormonedlayer(\border)$ and $\relu(\loopmat_{\loopdelta}\border)=\loopmat_{\loopdelta}\relu(\border)$. In the end, there is a global maxpooling $\globalmaxpoolonedlayer(\loopmat_{\loopdelta}\border)=\globalmaxpoolonedlayer(\border)$. With these layers as components, \resnetoned{~} would keep the loop origin invariance.
}

\subsubsection{\rvtwo{Proofs of Theorem \ref{the:nuftmlp_proof}}}   \label{sec:nuftmlp_proof}

\begin{figure*}[!ht]
	\centering \tiny
\begin{subfigure}[b]{0.495\textwidth}  
		\centering 
		\includegraphics[width=\textwidth]{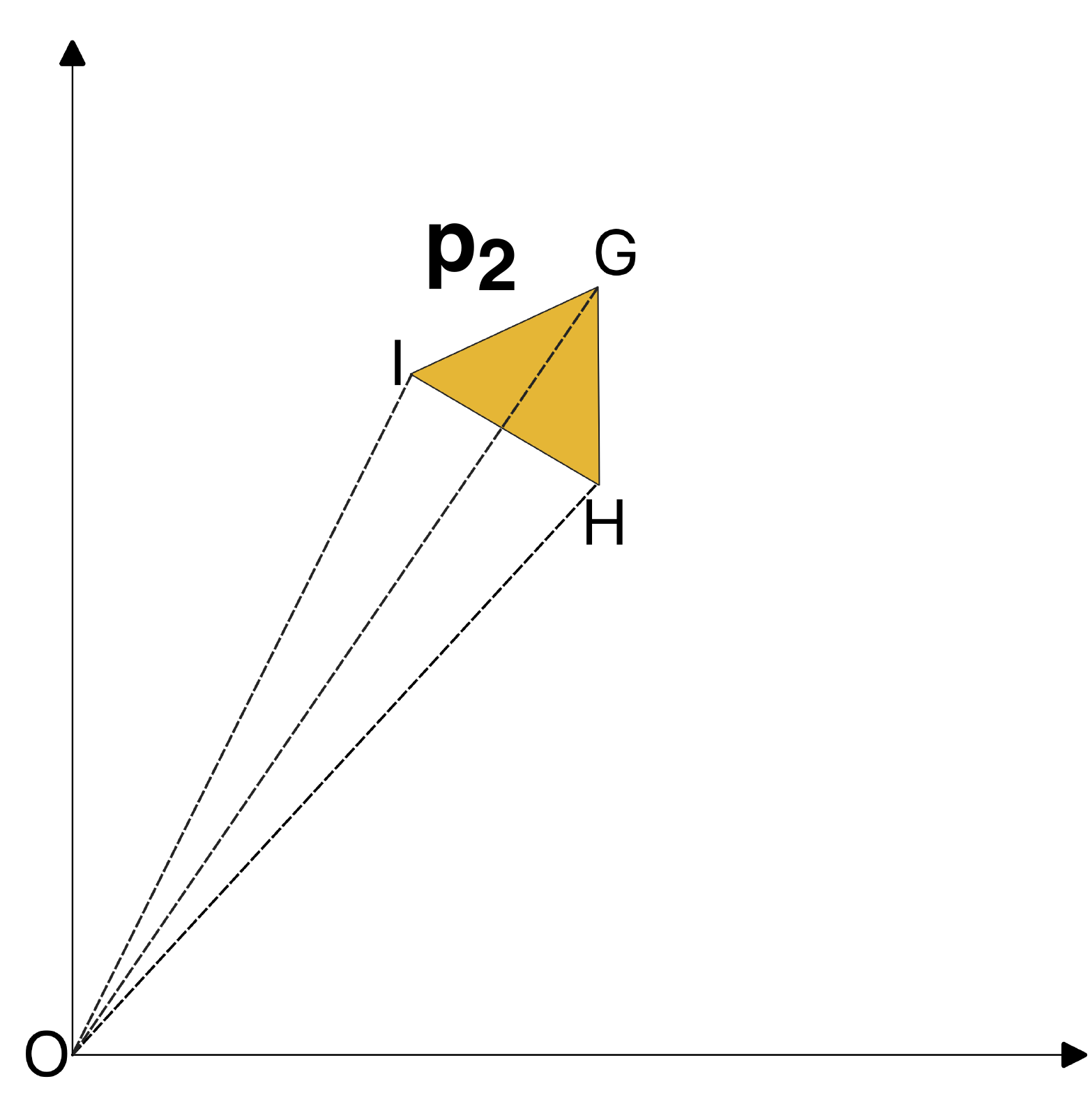}\vspace*{-0.2cm}
		\caption[]{
		\tiny{\rvtwo{\reviseone{$\pgon_2$ and $\simplexmesh_{2}^{(\simplexdim)} = \{ \simplex^{(\simplexdim)}_{GIO}, \simplex^{(\simplexdim)}_{IHO},
	    \simplex^{(\simplexdim)}_{HGO}\}$
		}}}}    
		\label{fig:pgon_prove_1}
	\end{subfigure}
	\hfill
	\begin{subfigure}[b]{0.495\textwidth}  
		\centering 
		\includegraphics[width=\textwidth]{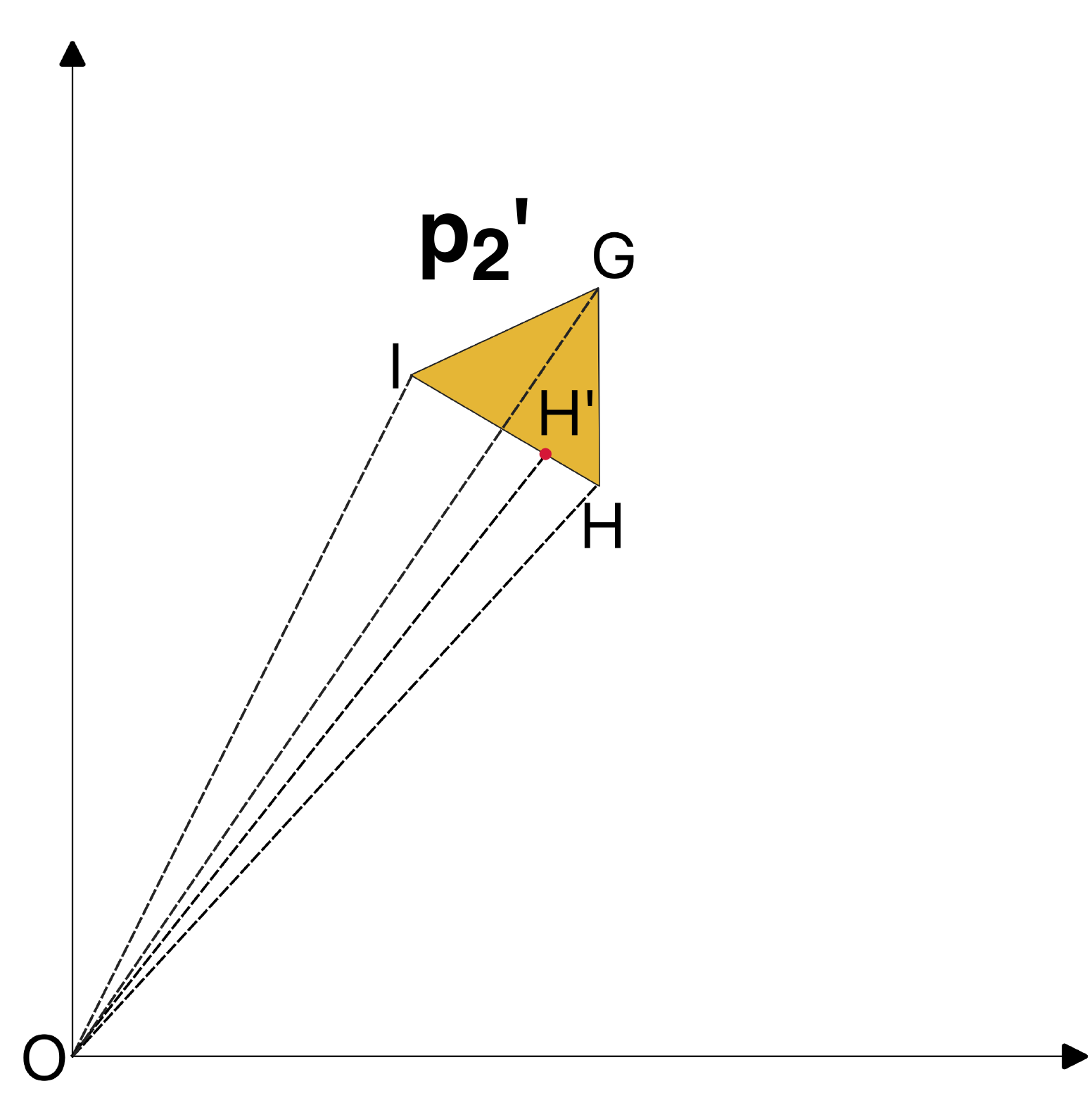}\vspace*{-0.2cm}
		\caption[]{\tiny{\rvtwo{\reviseone{$\pgon_2^{\prime}$ and $\simplexmesh_{2^\prime}^{(\simplexdim)} = \{ \simplex^{(\simplexdim)}_{GIO}, \simplex^{(\simplexdim)}_{IH^{\prime}O},
    	\simplex^{(\simplexdim)}_{H^{\prime}HO},
    	\simplex^{(\simplexdim)}_{HGO}\}$
		}}} }   
		\label{fig:pgon_prove}
	\end{subfigure}
	\caption{
	\rvtwo{
	A illustration to facilitate the proof of Theorem \ref{the:nuftmlp_proof} . 
	(a) Polygon $\pgon_2$ is a simple polygon from Figure \ref{fig:pgon1}. We convert it to a 2-simplex mesh $\simplexmesh_{2}^{(\simplexdim)} = \{ \simplex^{(\simplexdim)}_{GIO}, \simplex^{(\simplexdim)}_{IHO},
	\simplex^{(\simplexdim)}_{HGO}\}$.
	(a) Polygon $\pgon_2^{\prime}$ is also a simple polygon which has the same shape as $\pgon_2$ but add an additional trivial vertex $H^{\prime}$. We convert it to a 2-simplex mesh $\simplexmesh_{2^\prime}^{(\simplexdim)} = \{ \simplex^{(\simplexdim)}_{GIO}, \simplex^{(\simplexdim)}_{IH^{\prime}O},
	\simplex^{(\simplexdim)}_{H^{\prime}HO},
	\simplex^{(\simplexdim)}_{HGO}\}$.
	}
	}
	\label{fig:pgon_prove_trivial}
\end{figure*}
\vspace{-0.5cm}

\vspace{-0.2cm}
\rvtwo{
\paragraph{Proof of Theorem \ref{the:nuftmlp_proof} (1) - Loop Origin Invariance.}
Given a simple polygon $\pgon = (\border, \emptyset)$, we convert it to a 2-simplex mesh $\simplexmesh^{(\simplexdim)} = \{\simplex^{(\simplexdim)}_\simplexidx\}_{\simplexidx=1}^{\numpgonpt{}}$ similar to what is shown in Figure \ref{fig:pgon_prove_1}. Since $\{\simplex^{(\simplexdim)}_\simplexidx\}_{\simplexidx=1}^{\numpgonpt{}}$ is an \textit{unordered set} of 2-simplexes, for any loop matrix $\loopmat_{\loopdelta}$, Polygon $\pgon^{(\loopdelta)} = ( \loopmat_{\loopdelta}\border, \emptyset)$ will have exactly the same 2-simplex mesh as $\pgon$ - $\simplexmesh^{(\simplexdim)}$. 2-simplex mesh $\simplexmesh^{(\simplexdim)}$ is the input of our \nuftmlp. So the output polygon embedding $\pgonenc_{\nuftmlp}(\pgon)$ is invariant to any loop transformation $\loopmat_{\loopdelta}$ on Polygon $\pgon$'s exterior $\border$. Similarly, we can prove that the encoding results of \nuftmlp~ is also invariant to a loop transformation on the boundary of one of holes of the complex polygon geometry $\geom$.
}

\vspace{-0.5cm}
\rvtwo{
\paragraph{Proof of Theorem \ref{the:nuftmlp_proof} (2) - Trivial Vertex Invariance.}
We use Polygon $\pgon_2$ and $\pgon_2^{\prime}$ shown in Figure \ref{fig:pgon_prove_1} and \ref{fig:pgon_prove} as an example to demonstrate the proof. The only difference between them is that $\pgon_2^{\prime}$ has an additional trivial vertex $H^{\prime}$ while $\pgon_2$ and $\pgon_2^{\prime}$ have the same \textit{shape}. They have different 2-simplex meshes: $\simplexmesh_{2}^{(\simplexdim)} = \{ \simplex^{(\simplexdim)}_{GIO}, \simplex^{(\simplexdim)}_{IHO},
\simplex^{(\simplexdim)}_{HGO}\}$ 
and
$\simplexmesh_{2^\prime}^{(\simplexdim)} = \{ \simplex^{(\simplexdim)}_{GIO}, \simplex^{(\simplexdim)}_{IH^{\prime}O},
\simplex^{(\simplexdim)}_{H^{\prime}HO},
\simplex^{(\simplexdim)}_{HGO}\}$.
The Piecewise-Constant Function (PCF) $\pcf_{\simplexmesh}^{(\simplexdim)}(\ptcoord) = \sum_{\simplexidx=1}^{\numpgonpt{}} \pcf_{\simplexidx}^{(\simplexdim)}(\ptcoord)$ defined on the simplex mesh $\simplexmesh_{2}^{(\simplexdim)}$ is essentially a \textit{summation} over the individual density function $\pcf_{\simplexidx}^{(\simplexdim)}(\ptcoord)$ defined on each simplex of $\simplexmesh_{2}^{(\simplexdim)}$. 
Since $\pgon_2$ and $\pgon_2^{\prime}$ have the same \textit{shape}, the PCF defined on them should be exactly the same. Since \nuftmlp~ polygon embedding is derived from the NUFT of the PCF over a 2-simple mesh. So we can conclude that the \nuftmlp~ polygon embeddings of $\pgon_2$ and $\pgon_2^{\prime}$ should be the same. In other words, \nuftmlp~ is trivial vertex invariant.
}

\vspace{-0.5cm}
\rvtwo{
\paragraph{Proof of Theorem \ref{the:nuftmlp_proof} (3) - Part Permutation Invariance.} 
Given a multipolygon $\mtpgon = \{ \pgon_{\pgonidx} \}$, its 2-simplex mesh $\simplexmesh^{(\simplexdim)} = \{\simplex^{(\simplexdim)}_\simplexidx\}_{\simplexidx=1}^{\numpgonpt{}}$ (similar to Figure \ref{fig:pgon_simplex}) is an unordered set of signed 2-simplexes/triangles. Changing the feed-in order of polygon set $\{ \pgon_{\pgonidx} \}$ will not affect the resulting 2-simplex mesh. So \nuftmlp~ is part permutation invariant.
}

\vspace{-0.5cm}
\rvtwo{
\paragraph{Proof of Theorem \ref{the:nuftmlp_proof} (4) - Topology Awareness.} 
Given a polygon $\pgon = (\border, \holeset = \{\hole_{\holeidx}\})$ with holes, its 2-simplex mesh $\simplexmesh^{(\simplexdim)} = \{\simplex^{(\simplexdim)}_\simplexidx\}_{\simplexidx=1}^{\numpgonpt{}}$ is consist of oriented 2-simplexes. Since we require each polygonal geometry is oriented correctly, the right-hand rule can be used to compute the signed content of each 2-simplex. So the topology of polygon $\pgon = (\border, \holeset = \{\hole_{\holeidx}\})$ is preserved during the polygon-simple mesh conversion. Thus, \nuftmlp~ is aware of the topology of the input polygon geometry.
}

\section{Shape Classification Experiments}   
\label{sec:exp_shape_clas}
\reviseone{
Shape classification is an essential task for many computer vision \citep{kurnianggoro2018survey} and cartographic applications \citep{yan2019graph, yan2021graph}. In this study we focus on a large scale 
polygon classification 
dataset \mnistcomplex, which is based on the commonly used MNIST dataset.
}

\subsection{\mnistcomplex~Dataset}  \label{sec:mnist_data}
According to Table \ref{tab:shape_dataset_stat}, MNIST is the only large-scale open-sourced shape classification dataset \reviseone{with 60K training samples}. 
The original MNIST data \reviseone{\rvtwo{was originally designed} for optical character recognition (OCR) which uses images to present shapes, not polygons. Jiang et al. \citep{jiang2019ddsl} convert MNIST into a polygon shape dataset for shape classification purpose. Following their practice, we construct a polygon-based shape classification dataset -- \mnistcomplex.}

\begin{figure*}[ht!]
	\centering \tiny
	\vspace*{-0.2cm}
	\begin{subfigure}[b]{0.48\textwidth}  
		\centering 
		\includegraphics[width=\textwidth]{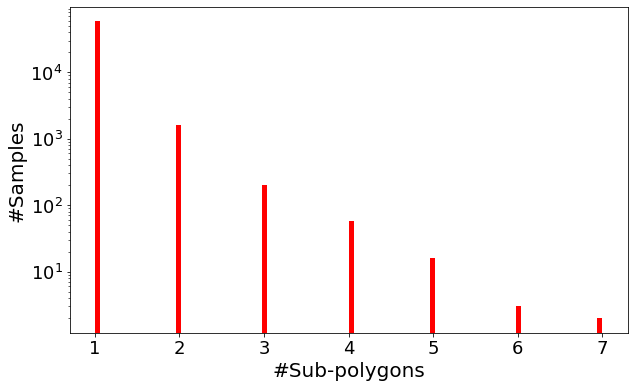}\vspace*{-0.2cm}
		\caption[]{\rvtwo{\#Sub-polygons in the training set
		}}    
		\label{fig:mnist_train_pgon_stat}
	\end{subfigure}
	\hfill
	\begin{subfigure}[b]{0.485\textwidth}  
		\centering 
		\includegraphics[width=\textwidth]{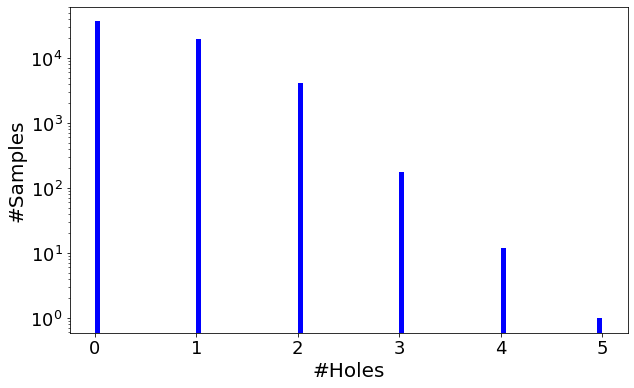}\vspace*{-0.2cm}
		\caption[]{\rvtwo{\#Holes in the training set
		}}    
		\label{fig:mnist_train_hole_stat}
	\end{subfigure}
	\hfill
	\begin{subfigure}[b]{0.48\textwidth}  
		\centering 
		\includegraphics[width=\textwidth]{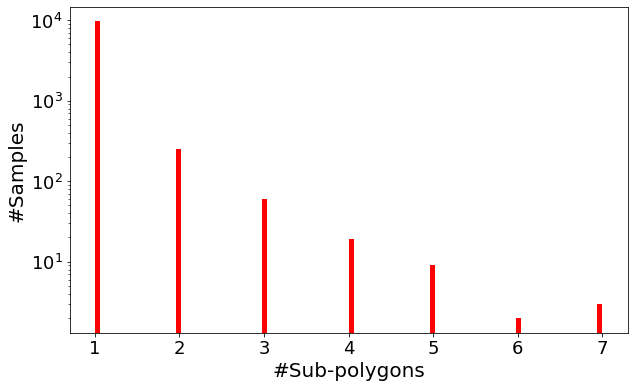}\vspace*{-0.2cm}
		\caption[]{\rvtwo{\#Sub-polygons in the testing set
		}}    
		\label{fig:mnist_test_pgon_stat}
	\end{subfigure}
	\hfill
	\begin{subfigure}[b]{0.485\textwidth}  
		\centering 
		\includegraphics[width=\textwidth]{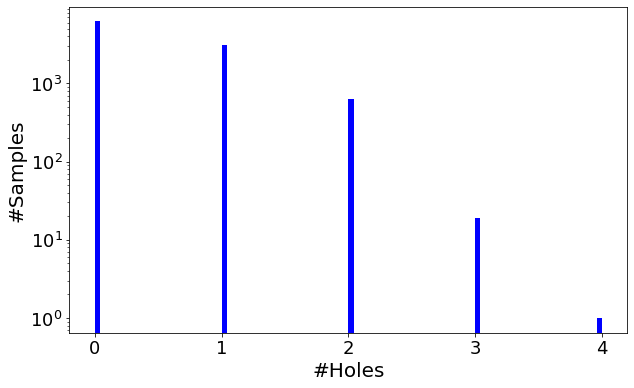}\vspace*{-0.2cm}
		\caption[]{\rvtwo{\#Holes in the testing set
		}}    
		\label{fig:mnist_test_hole_stat}
	\end{subfigure}
	\caption{\rvtwo{The statistic of the complexity of polygonal geometries in the \mnistcomplex~training and testing set.}
	} 
	\label{fig:mnist_stat}
\end{figure*}

The original MNIST pixel image is up-sampled using interpolation and contoured to get a polygonal representation of the digit by using the functions provided by Jiang et al. \cite{jiang2019ddsl}. Then we simplify each geometry \rvtwo{by: 1) making sure that} each polygon sample contains 500 unique vertices in order to do mini-batch training; 2) \rvtwo{dropping} very small holes and sub-polygons while \rvtwo{keeping} large ones. 
The result is a shape classification dataset of 70k examples (see Table \ref{tab:shape_dataset_stat} for its detailed statistics).
Figure \ref{fig:mnist_stat} shows the histogram statistic of the number of sub-polygons and number of holes of each samples in the resulting \mnistcomplex~training and \rvtwo{testing} set.

\subsection{Baselines and Model Variations} \label{sec:shp_cla_model}

We use the same shape classification module as shown in Equation \ref{equ:shape_cla_model} but vary the polygon encoders we used.
We consider multiple polygon encoders for the shape classification task on \mnistcomplex~dataset:

\begin{enumerate}
    \item \textbf{\veercnn}~\cite{veer2018deep}: We strictly follow the TensorFlow implementation\footnote{\url{https://github.com/SPINlab/geometry-learning}} of Veer et al. \citep{veer2018deep} and re-implement it in PyTorch. The polygon embedding is used for shape classification.
    
    \item \textbf{\resnetoned}: We implement \resnetoned~as described in Section \ref{sec:resnet1d} and use it for shape classification.
    
    \item \textbf{\ddsl+MLP}: This is a model developed based on the \ddsl~model proposed by Jiang et al. \cite{jiang2019ddsl}. Similar to \nuftmlp, \ddsl+MLP first transforms a polygonal geometry into the spectral space with NUFT. Instead of learning embedding directly on these NUFT features, it converts this NUFT representation back to the spatial space as an image by Inverse Fast Fourier Transform (IFFT). Note that in order to make sure the NUFT features can be later used for IFFT, we can only use linear grid feature map $\fftfreqmat^{(\linearfreqmtd)}$ (See Definition \ref{def:fft_freq}) as the Fourier feature bases for NUFT in \ddsl+MLP.
    This NUFT-IFFT transformation is essentially a vector-to-raster operation. \ddsl+MLP converts the resulting 2D image into a 1D feature vector and directly applies a multi-layer percetron (MLP) on it to learn polygon embeddings.
    
    \item \textbf{\ddsl+PCA+MLP}: Directly applying an MLP on 1D image feature vector can lead to overfitting. In order to prevent overfitting, \ddsl+PCA+MLP applies a Principal Component Analysis (PCA) on the 1D NUFT-IFFT image feature vectors across all training shapes in \mnistcomplex. The first $\pcadim$ PCA dimensions (which accounts for 80+\% of the data variances) of the NUFT-IFFT image feature vector are extracted and fed into an MLP.
    
    \item \textbf{\nuftmlp[\linearfreqmtd]+MLP}: We first use NUFT to transform a polygonal geometry into the spectral space with linear grid $\fftfreqmat^{(\linearfreqmtd)}$ (indicated by ``[\linearfreqmtd]''). Then an MLP is directly applied on the resulting NUFT features.
    
    \item \textbf{\nuftmlp[\linearfreqmtd]+PCA+MLP}: Similarly, in order to prevent overfitting, we apply a projection of NUFT features down to the first $\pcadim$ PCA dimensions (which account for 80+\% of the data variance) before the MLP.
    
    \item \textbf{\nuftmlp[\geometricfreqmtd]+MLP}: One advantage of \nuftmlp~is that since \nuftmlp~does not require IFFT, we do not need to use regularly spaced frequencies ($\fftfreqmat^{(\linearfreqmtd)}$) but choose whatever frequency map works best for a given task. So compared with \nuftmlp[\linearfreqmtd]+MLP, we switch to the geometric grid frequency map $\fftfreqmat^{(\geometricfreqmtd)}$ \rvtwo{as proposed in Mai et al. \cite{mai2020multiscale}.}
    
    \item \textbf{\nuftmlp[\geometricfreqmtd]+PCA+MLP}: Similar to \nuftmlp[\geometricfreqmtd]+MLP~but we perform an extra PCA before feeding into the MLP.
    
\end{enumerate}

\reviseone{
Both \veercnn~and \resnetoned~can only encode simple polygons by design while \mnistcomplex~contains complex polygonal geometries. So given a complex polygon geometry, we concatenate the vertex coordinate sequences of all sub-polygons' exterior and interior rings before feeding it into \veercnn~or \resnetoned. We call this preprocess step as \textit{boundary concatenation operation}.}
For example, Geometry $\mtpgon = \{\pgon_0, \pgon_1\}$ shown in Figure \ref{fig:pgon} is represented as $[\ptcoord_{A}^T;\ptcoord_{B}^T;\ptcoord_{C}^T;\ptcoord_{D}^T;\ptcoord_{E}^T;\ptcoord_{F}^T;\ptcoord_{G}^T;\ptcoord_{H}^T;\ptcoord_{I}^T;\ptcoord_{J}^T;\ptcoord_{K}^T;\ptcoord_{L}^T]$ before feeding into \veercnn~and \resnetoned. 
\reviseone{After boundary concatenation operation, a complex polygonal geometry is converted to a single boundary vertex sequence. It is like using one stroke to draw a complex polygonal geometry~\citep{ha2018sketchrnn} whereas all topology information is lost.
In this situation, circular padding does not ensure loop origin invariance for \resnetoned~anymore. 
}

\reviseone{
\veercnn~and \resnetoned~are spatial domain polygon encoders while the last six models are spectral domain polygon encoders. }
The last six NUFT-based polygon encoders essentially provide a framework for the ablation study on the usability of NUFT features for polygon representation learning. We vary several components of the polygon encoder: 1) \textbf{\ddsl~v.s. \nuftmlp} -- whether to learn representation in the spectral domain (\nuftmlp) or the spatial domains (\ddsl);
2) \textbf{\linearfreqmtd~v.s. \geometricfreqmtd} -- whether to use linear grid or geometric grid frequency map; 
3) \textbf{MLP v.s. PCA+MLP}: whether to apply PCA for feature projection before the MLP layers.
Here, we keep the learning neural network component to be an MLP to make sure the model performance differences are all from those three model variations but not from different neural architectures.

\rvtwo{The whole model architectures of \resnetoned, \nuftmlp~ as well as all baselines are implemented in PyTorch. All models are trained and evaluated on one Linux machine with 256 GB memory, 56 CPU cores, and 2 GeForce GTX 1080 Ti CUDA GPU (12 GB memory each). }

\subsection{Main Evaluation Results}  \label{sec:shp_cla_eval}

\begin{figure*}[b]
	\centering \tiny
	\vspace*{-0.2cm}
		\centering 
		\includegraphics[width=0.75\textwidth]{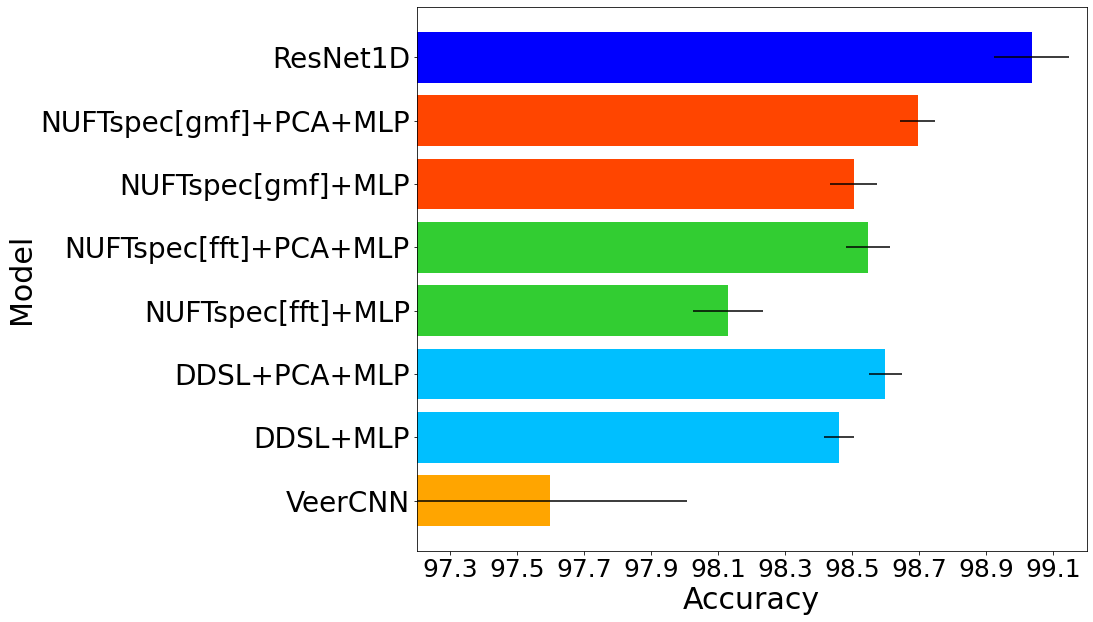}
	\caption{Overall classification accuracy comparison among different polygon encoders on the \mnistcomplex~\rvtwo{testing} set.
We also plot the standard deviation of each model's performance as black lines on each bar based 10 runs of each model. 
	We test multiple mode variations of \ddsl~and \nuftmlp.
	} 
	\label{fig:shp_cla_eval}
	\vspace{-0.7cm}
\end{figure*}

Figure \ref{fig:shp_cla_eval} compares the shape classification accuracy of the eight models described in Section \ref{sec:shp_cla_model} on the \mnistcomplex~\rvtwo{testing} set. 
Each bar indicates the performance of one model, where we mark the estimated standard deviations from 10 runs for each model setting. 
\rvtwo{
The hyperparameter tuning detailed are discussed in Appendix \ref{sec:para_tune}. The best hyperparameter combinations for different models are listed in Table \ref{tab:mnist_hyperpara}.
}
From Figure \ref{fig:shp_cla_eval}, we can see that:

\begin{enumerate}
    \item \veercnn~performs the worse. Our \resnetoned~achieves the best performance while our \nuftmlp[\geometricfreqmtd]+PCA+MLP comes as the second best.
    
    \item PCA significantly improves the accuracy of all three settings. This is especially evident for \nuftmlp[\linearfreqmtd], which supports our intuition that a linear grid introduces too many higher frequency features which are noisy and \rvtwo{make} learning hard (See Figure \ref{fig:fft_freq}). For more analysis on the PCA features, please see \rvtwo{Section} \ref{sec:shp_cla_pca}.
    
    \item When using the linear grid frequency map $\fftfreqmat^{(\linearfreqmtd)}$, two \ddsl~models outperform their corresponding \nuftmlp~counterparts, i.e., \ddsl+MLP $>$ \nuftmlp[\linearfreqmtd]+MLP, and \ddsl+PCA+MLP $>$ \nuftmlp[\linearfreqmtd]+PCA+MLP.
    
    \item However, when we switch to the geometric grid $\fftfreqmat^{(\geometricfreqmtd)}$, \nuftmlp~outperforms \ddsl, i.e., \nuftmlp[\geometricfreqmtd]+MLP $>$ \ddsl+MLP, and \nuftmlp[\geometricfreqmtd]+PCA+MLP $>$ \ddsl+PCA+MLP. These differences are statistically significant given the estimated standard deviations.
    Note that for the geometric grid $\fftfreqmat^{(\geometricfreqmtd)}$, IFFT is no longer applicable so that we can not use \ddsl~logic to transform those spectral features into the spatial domain. This shows the flexibility of \nuftmlp~in terms of the choice of Fourier frequency maps which have a significant impact on the model performance, while \ddsl~is restricted to uniform frequency sampling.
\end{enumerate}

In order to understand how different polygon encoders handle polygons with different complexity, we split the 10K polygon samples in the \mnistcomplex~\rvtwo{testing} dataset into 6 groups using the number of sub-polygons of each sample, and compare the performances 
of \veercnn, \ddsl+PCA+MLP, \resnetoned, and \nuftmlp[\geometricfreqmtd]+PCA+MLP 
on each group. The numbers in the parenthesis are the estimated standard deviations. The results are shown in Table \ref{tab:shp_cla_eval_split}. We can see that:

\begin{enumerate}
    \item Compared with \ddsl+PCA+MLP and \nuftmlp[\geometricfreqmtd]+PCA+MLP, \resnetoned~does poorly on multi-polygon samples\reviseone{, especially on samples with more than 3 subpolygons. }This indicates that the superority of \resnetoned~on \mnistcomplex~dataset is mainly because most of samples (96.56\%) in \mnistcomplex~testing set 
    are single polygons.
    \rvtwo{Therefore, for shape classification tasks} with a higher proportion of complex polygonal geometry samples, we expect  \nuftmlp[\geometricfreqmtd]+PCA+MLP \rvtwo{to perform better than \resnetoned.}
    
    \item Comparing \ddsl+PCA+MLP and \nuftmlp[\geometricfreqmtd]+PCA+MLP, we see that they both can handle multi-polygon samples well \reviseone{due to} the NUFT component. However, \nuftmlp[\geometricfreqmtd]+PCA+MLP performs better at 1 and 2 sub-polygon groups. 

    \item We find out that \ddsl+PCA+MLP and \nuftmlp[\geometricfreqmtd]+PCA+MLP have the same performance on the 3, 4, and 5 sub-polygon groups. This indicates that both models fail on the same number of samples in each group. After looking into the actual predictions, we find out that they actually fail \rvtwo{for} different samples.
\end{enumerate}

\begin{table}[t]
\caption{Shape classification \rvtwo{results}
on different polygon groups of the \mnistcomplex~\rvtwo{testing} set. 
\rvtwo{The ``\#Sub-Polygon'' row indicates the }number of sub-polygons each shape sample have \rvtwo{in each group}. 
``6+'' indicates all shape samples who have at least 6 sub-polygons.
``ALL'' indicates the evaluation on the whole \rvtwo{testing} set.
``\#Samples'' row shows the number of shape samples in each groups.
``()'' indicates the standard deviations of 10 runs of each model.
}
	\label{tab:shp_cla_eval_split}
	\centering
	\setlength{\tabcolsep}{3pt}
\begin{tabular}{l|c|c|c|c|c|c|c}
\toprule
\#Sub-Polygon                   & 1     & 2     & 3     & 4     & 5     & 6+    & ALL            \\ \hline
\#Samples                       & 9,656  & 250   & 61    & 19    & 9     & 5     & 10,000          \\ \hline
VeerCNN                         & 98.06 & 93.60 & 88.52 & 68.42 & 55.56 & 60.00 & 97.60 (0.43) \\
\ddsl+PCA+MLP                    & 98.78 & 95.20 & \textbf{96.72} & \textbf{84.21} & \textbf{77.78} & \textbf{80.00} & 98.60 (0.05) \\ \hline
\resnetoned                        & \textbf{99.25} & \textbf{96.00} & \textbf{96.72} & 63.16 & 66.67 & 40.00 & \textbf{99.00} (0.05) \\
\nuftmlp[\geometricfreqmtd]+PCA+MLP & 98.87 & \textbf{96.00} & \textbf{96.72} & \textbf{84.21} & \textbf{77.78} & 60.00 & 98.70 (0.05) \\ \bottomrule
\end{tabular}
\end{table}

\begin{table}[t]
\caption{
\reviseone{
Shape classification \rvtwo{results}
on different polygon groups of the 
\mnistcomplex~\rvtwo{testing} set (the same as Table \ref{tab:shp_cla_eval_split}) after training  models on the \mnistcomplexaug~dataset, which adds data augmentation for samples with more than 3 subpolygons.
}
	}
	\label{tab:shp_cla_eval_split_noise_aug}
	\centering
	\setlength{\tabcolsep}{4pt}
	\reviseone{
\begin{tabular}{l|c|c|c|c|c|c|c}
\toprule
\#Sun-Polygon                   & 1              & 2              & 3              & 4              & 5              & 6              & ALL            \\ \hline
\#Samples                       & 9,656           & 250            & 61             & 19             & 9              & 5              & 10,000          \\ \hline
\veercnn                         & 97.41          & 87.60          & 83.61          & 73.68          & 66.67          & 60.00          & 96.99          \\
\ddsl+PCA+MLP                    & 98.76          & 96.00          & 95.08          & 78.95          & \textbf{88.89} & \textbf{80.00} & 98.61          \\ \hline
\resnetoned                        & \textbf{99.10} & \textbf{96.80} & 96.72          & 68.42          & 77.78          & 60.00          & \textbf{98.93} \\
\nuftmlp[\geometricfreqmtd]+PCA+MLP & 98.81          & \textbf{96.80} & \textbf{98.36} & \textbf{89.47} & \textbf{88.89} & \textbf{80.00} & 98.76          \\ \bottomrule
\end{tabular}
}
\end{table}

\reviseone{
Since there are much less samples with more than 3 sub-polygons in \mnistcomplex~\rvtwo{training and testing} dataset (see Figure \ref{fig:mnist_train_pgon_stat} and \ref{fig:mnist_test_pgon_stat}), it might be questionable to draw a conclusion that compared with \ddsl+PCA+MLP and \nuftmlp[\geometricfreqmtd]+PCA+MLP, \resnetoned~does poorly on samples with more than 3 subpolygons. The lower performance of \resnetoned~on samples with more than 3 subpolygons might be due to \textit{the unbalanced training data with respect the number of samples with different sub-polygons}. 
}
\reviseone{
To mitigate the unbalanced nature of the training data, we perform data augmentation to increase the number of multi-polygon samples in the \rvtwo{training} set.
For each training sample with more than 3 subpolygons, in addition to the original sample, we generate 10 extra samples by adding random Gaussian noise to the vertices of current polygonal geometry. We denote this augmented training set as \mnistcomplexaug.
We train four polygon encoders on \mnistcomplexaug~dataset and evaluate them on the original \mnistcomplex~\rvtwo{testing} set. 
}

\reviseone{
The evaluation results on different polygon groups are shown in Table \ref{tab:shp_cla_eval_split_noise_aug}. We can see that, after \rvtwo{adding extra} multipolygon samples, the overall accuracy of \resnetoned~decrease whereas the over accuracy of \nuftmlp[\geometricfreqmtd]+PCA+MLP increase when we compare them with those in Table \ref{tab:shp_cla_eval_split}. 
Moreover, we can still observe that both \ddsl+PCA+MLP and \nuftmlp[\geometricfreqmtd]+PCA+MLP can outperform \resnetoned~on 4, 5, 6+ sub-polygon groups. 
This result demonstrates that \resnetoned~has an intrinsic model bias to perform poorly on multi-polygon samples as opposed to model variance (lack of training examples).
}

\subsection{\rvtwo{Analysis of PCA-based Models}}  \label{sec:shp_cla_pca}
\begin{figure*}[ht!]
	\centering \tiny
	\vspace*{-0.2cm}
	\begin{subfigure}[b]{0.48\textwidth}  
		\centering 
		\includegraphics[width=\textwidth]{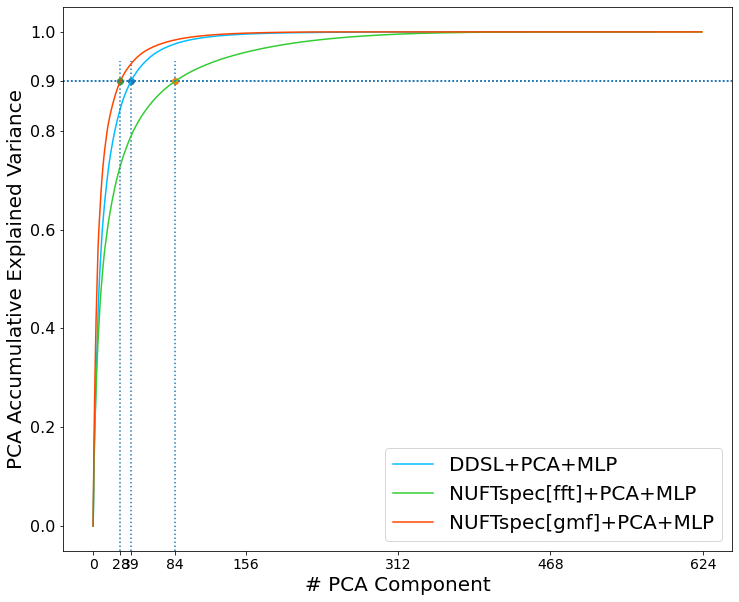}\vspace*{-0.1cm}
		\caption[]{{PCA explained variance curve ($\fftfreqnumX=24$)
		}}    
		\label{fig:pca_exp_var}
	\end{subfigure}
	\hfill
	\begin{subfigure}[b]{0.49\textwidth}  
		\centering 
		\includegraphics[width=\textwidth]{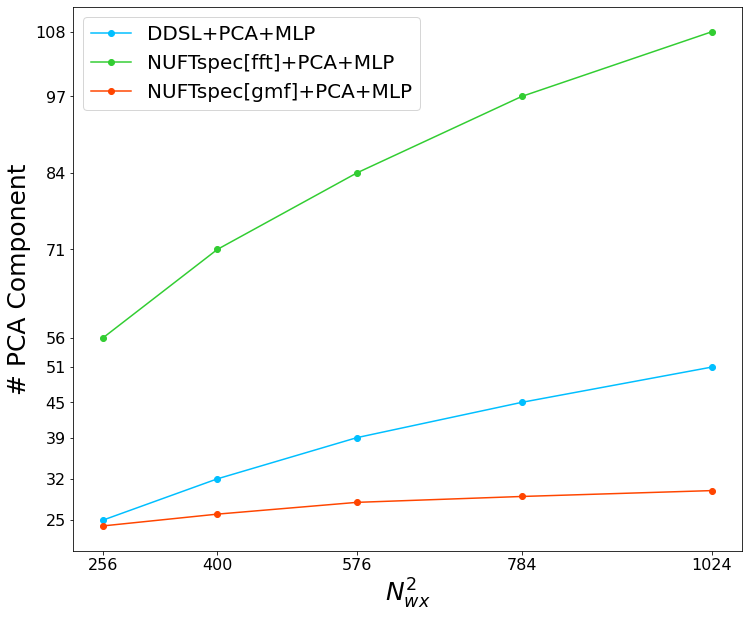}\vspace*{-0.2cm}
		\caption[]{{Top $\pcadim$ PCA components for 90\% data variance
		}}    
		\label{fig:pca_var_curve}
	\end{subfigure}
	\caption{\rvtwo{The comparison of \ddsl+PCA+MLP, \nuftmlp[\linearfreqmtd]+PCA+MLP and \nuftmlp[\geometricfreqmtd]+PCA+MLP on how their PCA components explain the data variance.
	(a) The PCA explained variance curves of three models when $\fftfreqnumX=24$. 
	(b) The curve of the number of top $\pcadim$ PCA components that can explain 90\% of data variance for \mnistcomplex~training set with different $\fftfreqnumX$.
	We can see that compared with the other two models, \nuftmlp[\geometricfreqmtd]+PCA+MLP has a more compact set of PCA features.}
	} 
	\label{fig:pgon_enc_pca_curve}
\end{figure*}

\rvtwo{
We also investigate the nature of the feature representations of those three PCA models -- \ddsl+PCA+MLP, \nuftmlp[\linearfreqmtd]+PCA+MLP and \nuftmlp[\geometricfreqmtd]+PCA+MLP. The results are show in Figure \ref{fig:pgon_enc_pca_curve}.
Figure \ref{fig:pca_exp_var} compares the PCA explained variance curves of three models when $\fftfreqnumX=24$. 
Instead of fixing $\fftfreqnumX$, Figure \ref{fig:pca_var_curve} compares three PCA models based on the number of PCA components needed to explain 90\% of the data variance with different $\fftfreqnumX$. We can see that for any given $\fftfreqnumX$, these three models need different numbers of PCA components with \nuftmlp[\geometricfreqmtd]+PCA+MLP being the most compact and \nuftmlp[\linearfreqmtd]+PCA+MLP being the least compact. The curve of \nuftmlp[\geometricfreqmtd]+PCA+MLP is very flat indicating that geometric frequency introduces much less noise when $\fftfreqnumX$ increases. }

\subsection{Understanding the Four Polygon Encoding Properties} \label{sec:shp_cla_invar_prop}

Section \ref{sec:shp_cla_eval} shows that \nuftmlp~and \resnetoned~are two best models on \mnistcomplex~dataset. 
In order to deeply understand how those four polygon encoding properties discussed in Section \ref{sec:prob_stat} affect the performance of \nuftmlp~and \resnetoned, 
\reviseone{we conduct four follow-up experiments. 
To test the rotation invariance property of their proposed models, both Deng et al. \citep{deng2021vector} and Esteves et al. \citep{esteves2018learning} kept the training dataset unchanged and modified the \rvtwo{testing} data by rotating each \rvtwo{testing} shape. They investigated how the \rvtwo{testing} shape rotations affect the model performance. Inspired by \rvtwo{them}, we also keep the training dataset of \mnistcomplex~unchanged and adopt four ways to modify the polygon representations in the \mnistcomplex~\rvtwo{testing} dataset. We compare the performances of \nuftmlp~and \resnetoned~on these modified datasets.
}

The first three properties are invariance properties. So what we expect for a polygon encoder is that when we 1) loop around the exterior/interiors \rvtwo{by starting} with different vertices, or 2) add/delete trivial vertices, or 3) permute the feed-in order of different sub-polygons, the resulting polygon embedding is invariant since the shape of the input polygon does not change. So for the first three properties, we modify the polygon representations in \mnistcomplex~\rvtwo{testing} set while keeping their shape invariant. We call them \textit{shape-invariant geometry modifications}.

The topology awareness property is not an invariance property. So our expectation is that when the topology of a polygon changes, the polygon embedding should be different even if \reviseone{they have the same vertex information such as $\mtpgon = \{\pgon_0, \pgon_1\}$ and  $\mtpgon^{\prime\prime} = \{\pgon_0^{\prime\prime}, \pgon_1, \pgon_2\}$ shown in Figure \ref{fig:pgon} and \ref{fig:pgon1}.} 

\begin{figure*}[ht!]
	\centering \tiny
	\vspace*{-0.2cm}
	\begin{subfigure}[b]{0.48\textwidth}  
		\centering 
		\includegraphics[width=\textwidth]{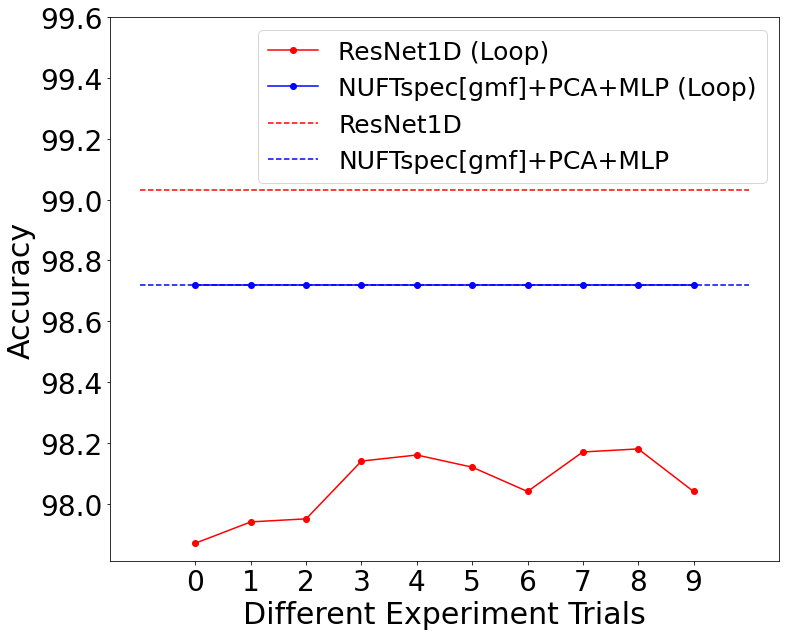}\vspace*{-0.2cm}
		\caption[]{{Loop \reviseone{origin} invariance
		}}    
		\label{fig:loop_exp}
	\end{subfigure}
	\hfill
	\begin{subfigure}[b]{0.49\textwidth}  
		\centering 
		\includegraphics[width=\textwidth]{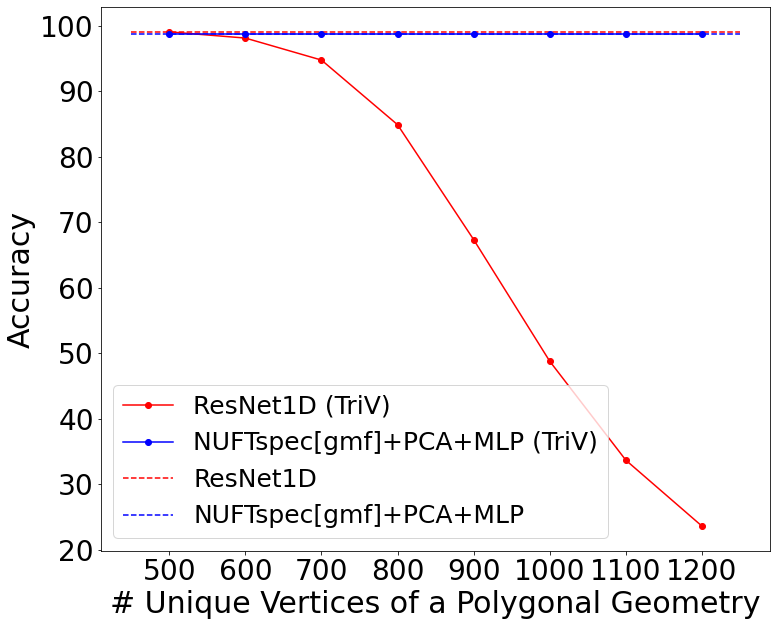}\vspace*{-0.2cm}
		\caption[]{{Trivial vertex invariance
		}}    
		\label{fig:triv_exp}
	\end{subfigure}
	\hfill
	\begin{subfigure}[b]{0.48\textwidth}  
		\centering 
		\includegraphics[width=\textwidth]{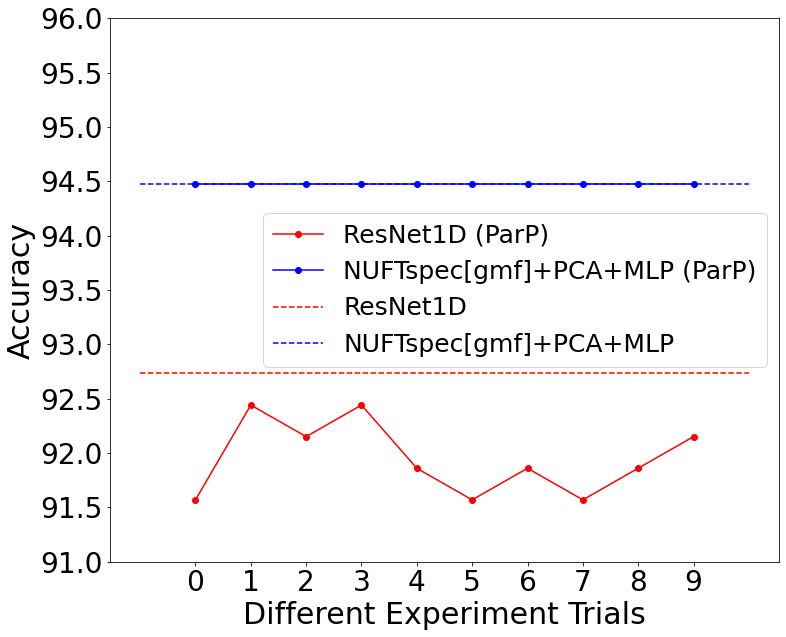}\vspace*{-0.2cm}
		\caption[]{{Part permutation invariance
		}}    
		\label{fig:parp_exp}
	\end{subfigure}
	\hfill
	\begin{subfigure}[b]{0.45\textwidth}  
		\centering 
		\includegraphics[width=\textwidth]{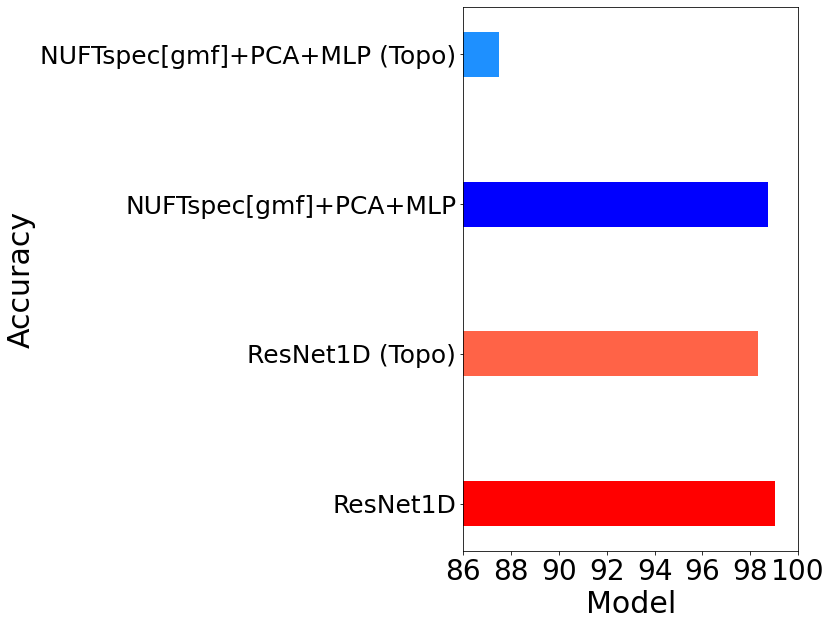}\vspace*{-0.2cm}
		\caption[]{{Topology awareness
		}}    
		\label{fig:topo_exp}
	\end{subfigure}
	\caption{
\reviseone{
    The experiments on the four polygon encoding properties of  \nuftmlp[geometric]+PCA+MLP and \resnetoned.
    ``\resnetoned'' and ``\nuftmlp[geometric]+PCA+MLP'' indicate the \rvtwo{evaluations} of these two models on the original \mnistcomplex~\rvtwo{testing} set while model names with ``(*)'' indicate their results on modified datasets. ``(*)'' indicates different modification methods for different properties:
    (a) Loop origin invariance: looping around the \rvtwo{testing} polygons' exterior/interiors with 10 different origins;
    (b) Trivial vertex invariance: adding different numbers of trivial vertices;
    (c) Part permutation invariance: permuting the feed-in order of different sub-polygons;
    (d) Topology awareness: converting polygon holes into subpolygons.
    }
	} 
	\label{fig:pgon_enc_property_study}
\end{figure*}

Figure \ref{fig:pgon_enc_property_study} visualizes the evaluate results for each \rvtwo{property. }

\begin{enumerate}
    \item \textbf{Loop \reviseone{origin} invariance (Loop)}: As for all training and testing polygons in original \mnistcomplex~dataset, we always start looping around each polygon's exterior/interiors with its upper most vertex.
    To study the loop origin invariance property, We randomize the starting point of the coordinate sequence of each sub-polygon's exterior or interiors. We refer to this method as \textit{loop \rvtwo{origin} randomization}. For example, as shown in Figure \ref{fig:pgon}, the exterior of Polygon $\pgon_0$, i.e., $\border_0$, can be written as $\border_0 = [\ptcoord_{A}^T;\ptcoord_{B}^T;\ptcoord_{C}^T;\ptcoord_{D}^T;\ptcoord_{E}^T;\ptcoord_{F}^T] \in \Real^{6 \times 2}$ with $\ptcoord_{A}$ as the start point. When we select $\ptcoord_{D}$ as the start point, the exterior of Polygon $\pgon_0$ becomes $\border_0^{(3)} = \loopmat_{3}\border_0 =  [\ptcoord_{D}^T;\ptcoord_{E}^T;\ptcoord_{F}^T;\ptcoord_{A}^T;\ptcoord_{B}^T;\ptcoord_{C}^T]$ which has \rvtwo{an} identical shape as $\border_0$.
In total, we generate 10 independent \rvtwo{testing} datasets based on the \mnistcomplex~original \rvtwo{testing} set by using the above method. We \rvtwo{evaluate} the trained \resnetoned~and \nuftmlp~model on these different modified \rvtwo{testing} sets whose evaluation results are shown as two curves -- ``ResNet1D (Loop)'' and ``NUFTspec[geometric]+PCA+MLP (Loop)'' in Figure \ref{fig:loop_exp}. We also show their performance on the original \rvtwo{testing} set as the dotted lines.
We can see that while \nuftmlp’s performance is unaffected under loop origin randomization, \resnetoned~is affected with roughly absolute 1\% performance decrease, which is statistically significant.
    Note that \resnetoned~is loop origin invariance only if the polygon is a simple polygon. However, \mnistcomplex~contains multiple complex polygonal geometries. In these cases, loop origin invariance property of \resnetoned~does not hold. That is why we see this performance drop.
    
    \item \textbf{Trivial vertex invariance (TriV)}: For each polygonal geometry in the \mnistcomplex~\rvtwo{testing} dataset, we upsample its vertices to have more than the initial 500 points (600 - 1200) by adding trivial vertices. Figure \ref{fig:pgon_shape} shows the example of trivial vertices (red points). We compare the performance of \resnetoned~and \nuftmlp~on these upsampled \rvtwo{testing} datasets which indicate as ``ResNet1D (TriV)'' and ``NUFTspec[geometric]+PCA+MLP (TriV)'' in Figure \ref{fig:triv_exp}. 
We can see that while the performance of \nuftmlp~is unaffected no matter how many vertices we upsampled, the performance of \resnetoned~decreases dramatically when the number of vertices increases.

    \item \textbf{Part permutation invariance (ParP)}: The feed-in order of sub-polygons in each polygonal geometry sample in the original \mnistcomplex~dataset follows the normal raster scan order -- from up to down and from left to right. To test the part permutation invariance property, we do random part permutation for each multipolygon in  \mnistcomplex~\rvtwo{testing} set. Since this part permutation operation will only affect the multipolygon shape samples, we only compare the performance of \resnetoned~and \nuftmlp~on this subset of the \rvtwo{testing} set which consists of 344 multipolygon samples. 
    Similar to Figure \ref{fig:loop_exp}, we also do 10 different experiment trials. The performances of \resnetoned~and \nuftmlp~on these permuted \rvtwo{testing} datasets are indicated as ``ResNet1D (ParP)'' and ``NUFTspec[geometric]+PCA+MLP (ParP)''. Dotted lines indicates their performance on the original dataset. We can see that while \nuftmlp~is unaffected, \resnetoned's performance decreased by 0.8\%.

    \item \textbf{Topology awareness (Topo)}: From a theoretical perspective, it is easy to see that \nuftmlp~is topology aware \reviseone{since NUFT knows which points are inside the polygon and which outside whereas \resnetoned~is not as shown in \rvtwo{Section \ref{sec:nuftmlp_proof}.} }However, since topology awareness is not an invariance property, it is rather difficult to show with experiments. Nevertheless, to align with the experiments of other three properties, we did a similar experiment by modifying the polygon representations. More specifically, \reviseone{for each polygonal geometry with holes, }
    we delete the holes and use the holes' coordinate sequence to construct a new sub-polygon for the current geometry. 
    One example consists of the multipolygon $\mtpgon = \{\pgon_0, \pgon_1\}$ and  multipolygon $\mtpgon^{\prime\prime} = \{\pgon_0^{\prime\prime}, \pgon_1, \pgon_2\}$ shown in Figure \ref{fig:pgon} and \ref{fig:pgon1}. The hole of sub-polygon $\pgon_0$ is deleted. Instead, it is instantiated as a new sub-polygon $\pgon_2$. By doing that, we change the topology of a geometry while keeping the vertices unchanged. Note that when a hole become the exterior of a sub-polygon, the coordinate sequence should switch from clockwise rotation to \rvtwo{counterclockwise} rotation.
    We evaluate \resnetoned~and \nuftmlp~on the original and modified \mnistcomplex~\rvtwo{testing} set as shown in Figure \ref{fig:topo_exp}. ``ResNet1D (Topo)'' and ``NUFTspec[\geometricfreqmtd]+PCA+MLP (Topo)'' indicate the results on the modified dataset. We can see that when changing the polygon topology, the performances of both models are affected while \nuftmlp~is affected more severely. This is actually expected, or even desired since when the topology of a polygonal geometry changes, the shape also changes and the decision of shape classification should also change. Figure \ref{fig:topo_exp} shows that 
    \nuftmlp~are more sensitive to topological changes of polygons. \end{enumerate}

In conclusion, based on the above study, we can see that compared with \resnetoned, \nuftmlp~is \textit{more robust to \reviseone{loop origin randomization}, vertex upsampling, and part permutation operation \reviseone{whereas it} is more sensitive to topological changes}. In other words, \textit{\nuftmlp~retains identical performance \reviseone{when polygons undergo shape-invariant geometry modifications} due to the invariance \reviseone{inherented from} 
the NUFT representation, whereas models that directly utilize the polygon vertex features such as \resnetoned~suffer significant performance degradations.}

\begin{figure*}[ht!]
	\centering \tiny
	\vspace*{-0.2cm}
	\begin{subfigure}[b]{1.0\textwidth}  
		\centering 
		\includegraphics[width=\textwidth]{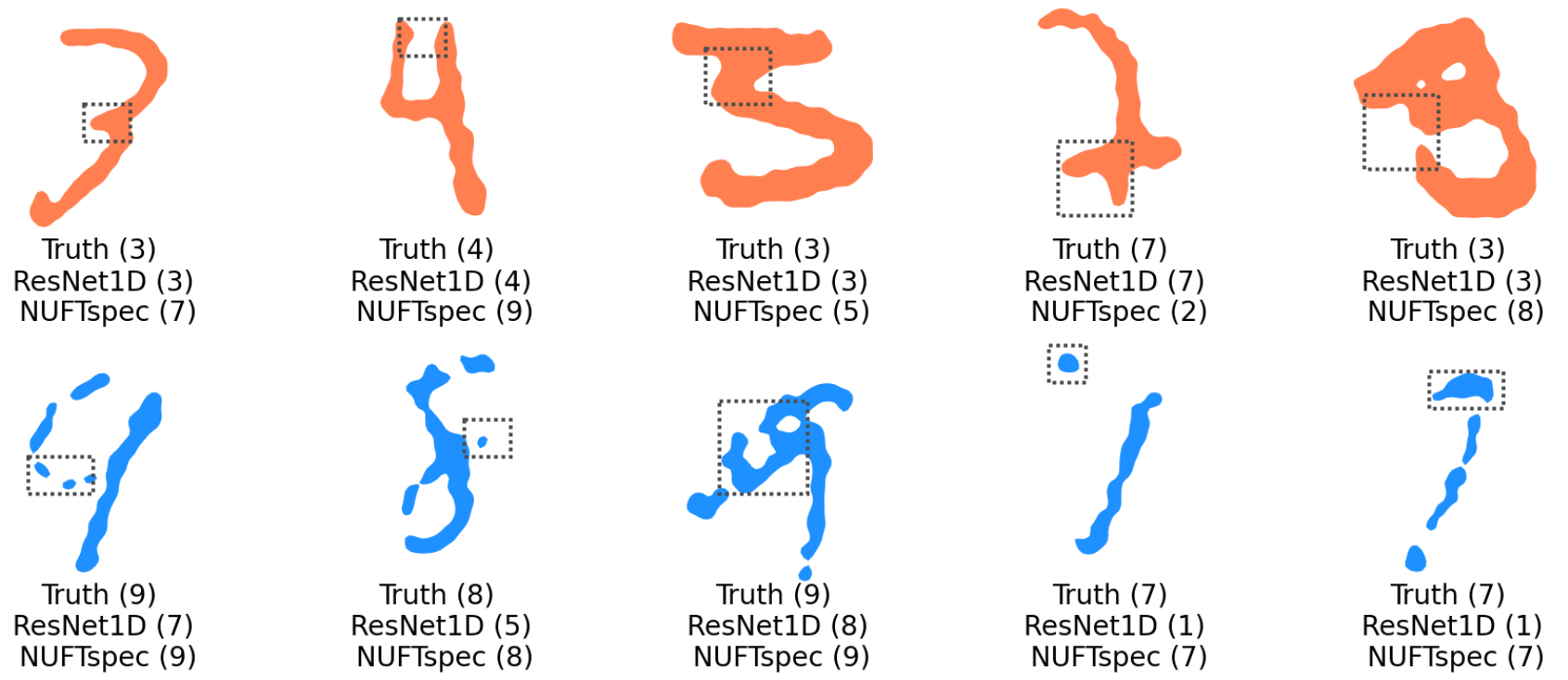}
		\caption[]{{\resnetoned~v.s. \nuftmlp[\geometricfreqmtd]+PCA+MLP
		}}    
		\label{fig:rs_vs_nuft}
	\end{subfigure}
	\hfill
	\begin{subfigure}[b]{1.0\textwidth}  
		\centering 
		\includegraphics[width=\textwidth]{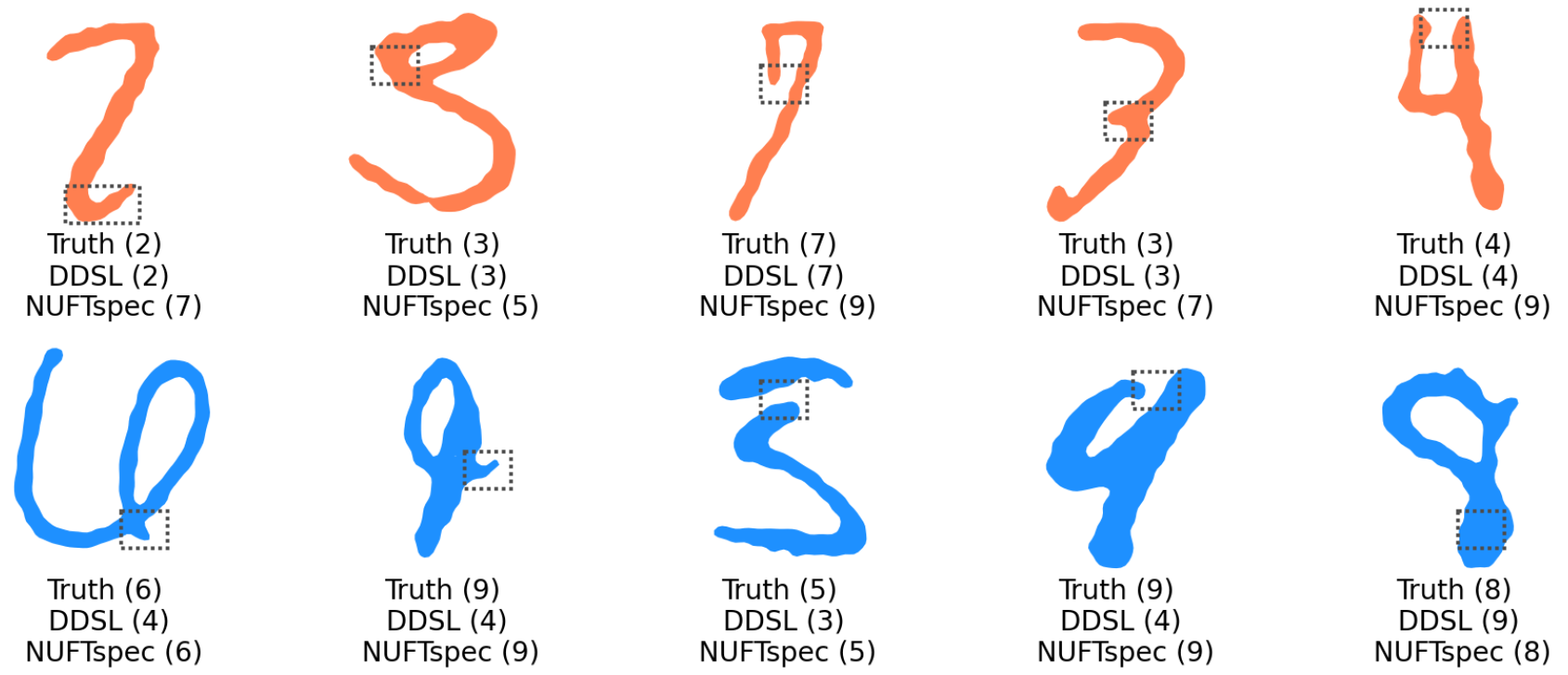}
		\caption[]{{\ddsl+PCA+MLP v.s. \nuftmlp[\geometricfreqmtd]+PCA+MLP
		}}    
		\label{fig:ddsl_vs_nuft}
	\end{subfigure}
	\caption{
	A qualitative analysis on the performance of \ddsl+PCA+MLP, \resnetoned, and \nuftmlp[\geometricfreqmtd]+PCA+MLP on the \mnistcomplex~\rvtwo{testing} set.
In each example's caption, ``*'' in ``Truth (*)'' indicates the ground truth labels while the other numbers in ``()'' indicates the predicted labels for different models.
	\reviseone{The dashed boxes highlight the parts which might lead to the wrong predictions of polygon encoders. }
	(a) The upper row contains examples where \resnetoned~predicts correctly while \nuftmlp[\geometricfreqmtd]+PCA+MLP fails. The lower row contains examples in which these two models \reviseone{perform otherwise. }(b) The upper row contains examples where \ddsl+PCA+MLP predicts correctly while \nuftmlp[\geometricfreqmtd]+PCA+MLP fails. The lower row contains examples in which these two models \reviseone{perform otherwise. }} 
	\label{fig:shp_cla_illu_example}
\end{figure*}

\subsection{Qualitative Analysis}  \label{sec:shp_cla_example}

Figure \ref{fig:shp_cla_illu_example} shows the qualitative analysis results of \ddsl+PCA+MLP, \resnetoned, and \nuftmlp[\geometricfreqmtd]+PCA+MLP.
We visualize some illustrated examples in which these models win or fail.

Figure \ref{fig:rs_vs_nuft} shows sampled winning and failing examples comparing \nuftmlp[\geometricfreqmtd]+PCA+MLP and \resnetoned. From the lower row of Figure \ref{fig:rs_vs_nuft}, we can see that \nuftmlp[\geometricfreqmtd]+PCA+MLP is able to jointly consider all sub-polygons of a multipolygon sample and the spatial relations among them during prediction while \resnetoned~fails to do so. From the upper row of Figure \ref{fig:rs_vs_nuft}, we can see that \nuftmlp[\geometricfreqmtd]+PCA+MLP sometimes fails because it can not recognize the shape details which are sometimes important for classification (such as the 1st case -- ``7'' vs ``3'').

Figure \ref{fig:ddsl_vs_nuft} shows sampled positive and negative examples comparing \nuftmlp[\geometricfreqmtd]+PCA+MLP and \ddsl+PCA+MLP. 
\reviseone{From the model design choice, we expect}
\nuftmlp~works better than \ddsl~in certain cases because \ddsl~is essentially a vector-to-raster operation that may lose global information that is important for a given task. 
As shown in the first 2 examples in the lower row, \ddsl+PCA+MLP may classify shapes as ``4'' because topologically the shapes do conform to ``4''. However, based on the overall shape \nuftmlp[\geometricfreqmtd]+PCA+MLP is able to determine that they are more likely ``6'' and ``9''. The small ``local'' protruding elements \reviseone{highlighted in dashed boxes} in these examples are not intended by the writer, and have a relatively small impact on \nuftmlp's features, since all Fourier features are global statistics. Similarly having no hole in the ``8'' example (the last example in the lower row \rvtwo{of Figure \ref{fig:ddsl_vs_nuft}}) prevents \ddsl+PCA+MLP from classifying it as ``8'', while \nuftmlp[\geometricfreqmtd]+PCA+MLP \reviseone{can well handle this topological abnormality. }On the other hand, this insensitivity to local features may also hurt \nuftmlp[\geometricfreqmtd]+PCA+MLP when they are important. This is evident in the upper row in Figure \ref{fig:ddsl_vs_nuft} for number ``2'', ``3'', and ``7'' where a small protruding structure, or a small gap indicates the topological changes that are intended by the writer.

Based on the above experiment results, we can see that compared with \resnetoned, both \ddsl~and \nuftmlp~are better at handling multi-polygon samples. While \ddsl~focuses on the localized features, \nuftmlp~pays attention to the global shape information. Moreover, since \nuftmlp~does not have the IFFT step, it is more flexible in terms of choosing the frequency maps.

\section{Spatial Relation Prediction Experiments}   \label{sec:exp_spa_rel}
\reviseone{The polygon-based spatial relation prediction is an important component for GeoQA \citep{mai2021geographic}. To study the effectiveness of \nuftmlp~and \resnetoned~on this task, we construct two real-world datasets \dbtopo~and \dbtopocomplex~for the evaluation purpose based on DBpedia and OpenStreetMap.}

\subsection{\dbtopo~and \dbtopocomplex~Dataset}
\label{sec:dataset}

\begin{figure}[ht!]
	\centering \tiny
\begin{subfigure}[b]{0.329\textwidth}  
		\centering 
		\includegraphics[width=\textwidth]{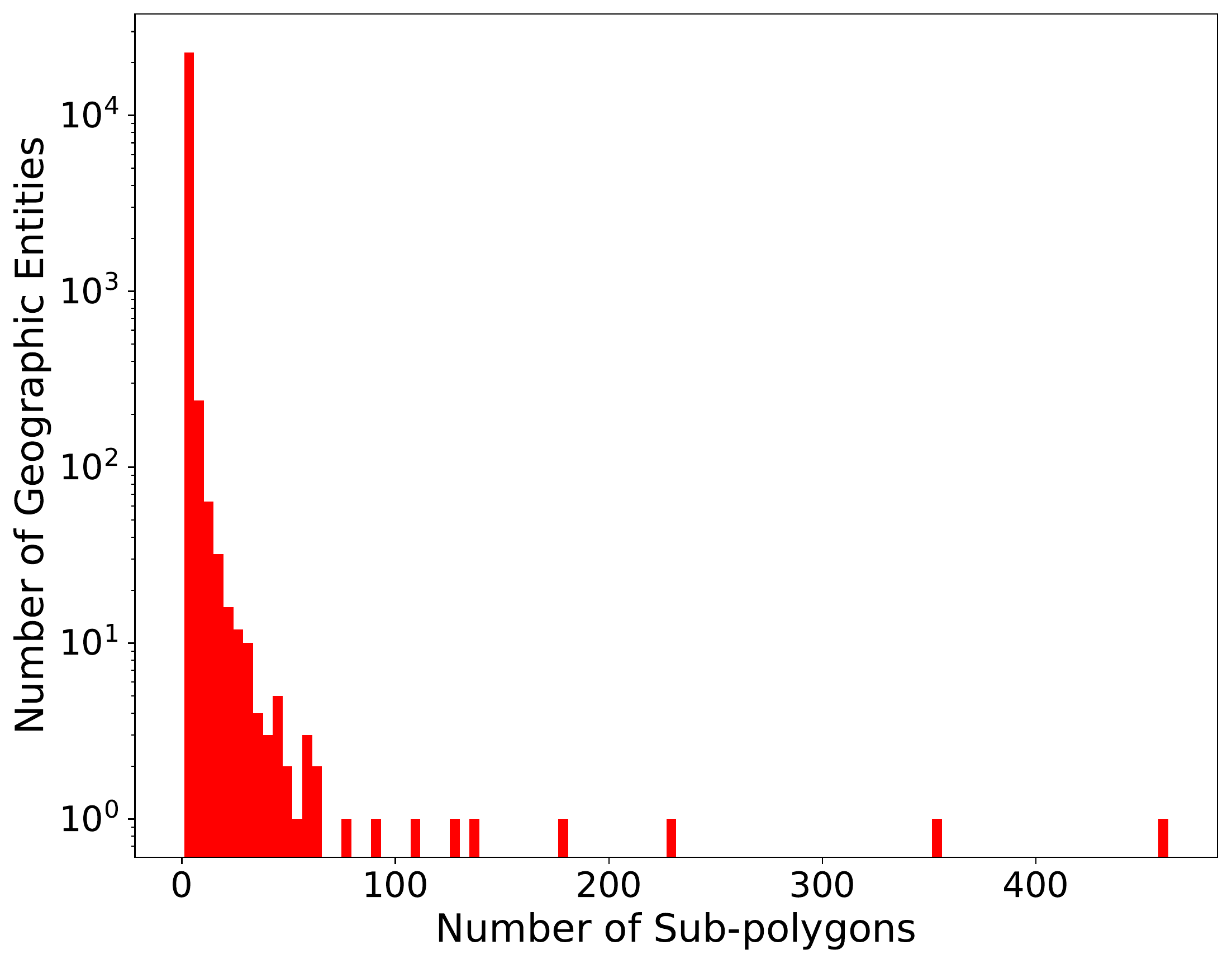}\caption[]{\rvtwo{\# Sub-polygons
		}}    
		\label{fig:dbtopo_part_hist}
	\end{subfigure}
	\hfill
	\begin{subfigure}[b]{0.329\textwidth}  
		\centering 
		\includegraphics[width=\textwidth]{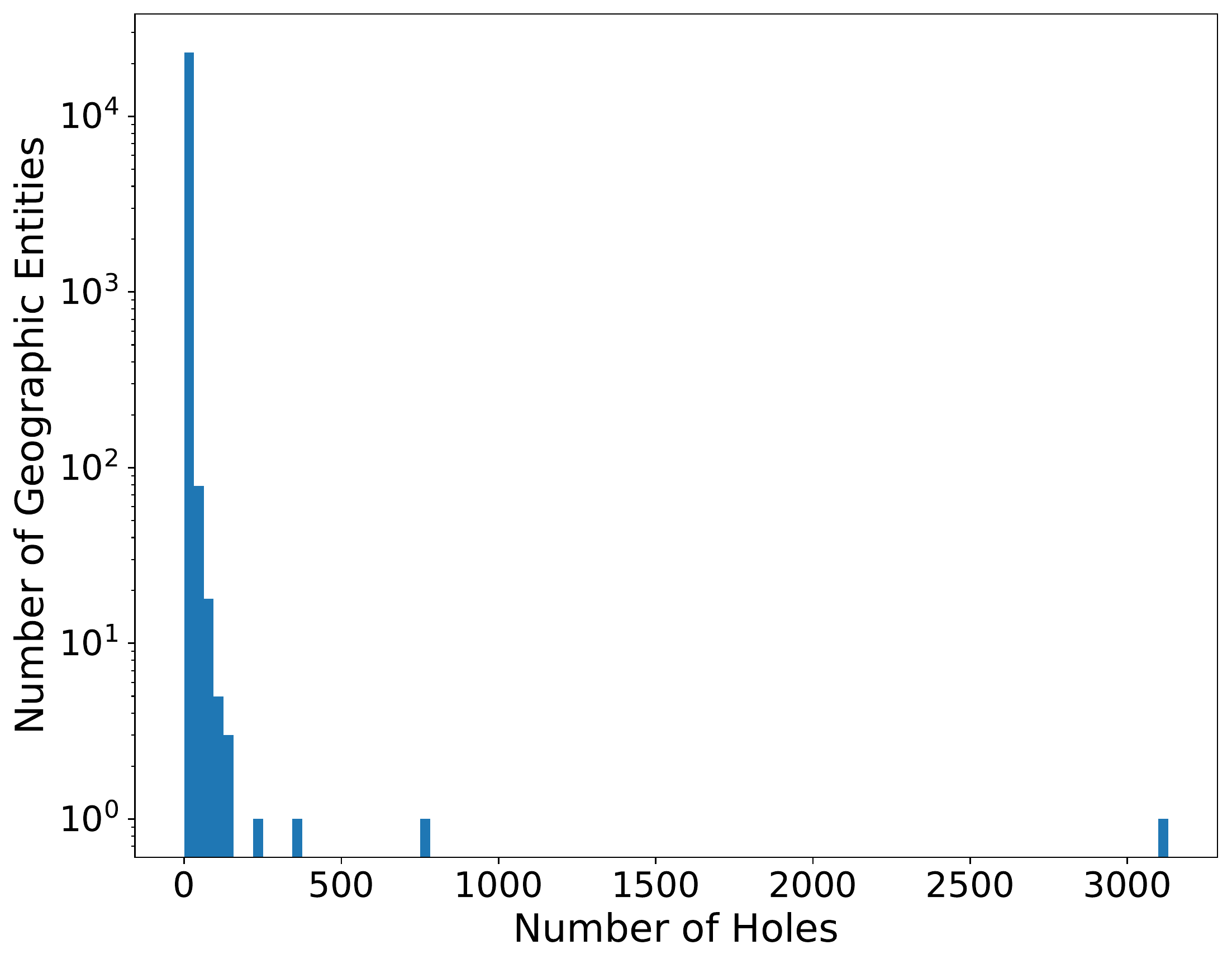}\caption[]{\rvtwo{\# Holes
		}}    
		\label{fig:dbtopo_hole_hist}
	\end{subfigure}
	\hfill
	\begin{subfigure}[b]{0.329\textwidth}  
		\centering 
		\includegraphics[width=\textwidth]{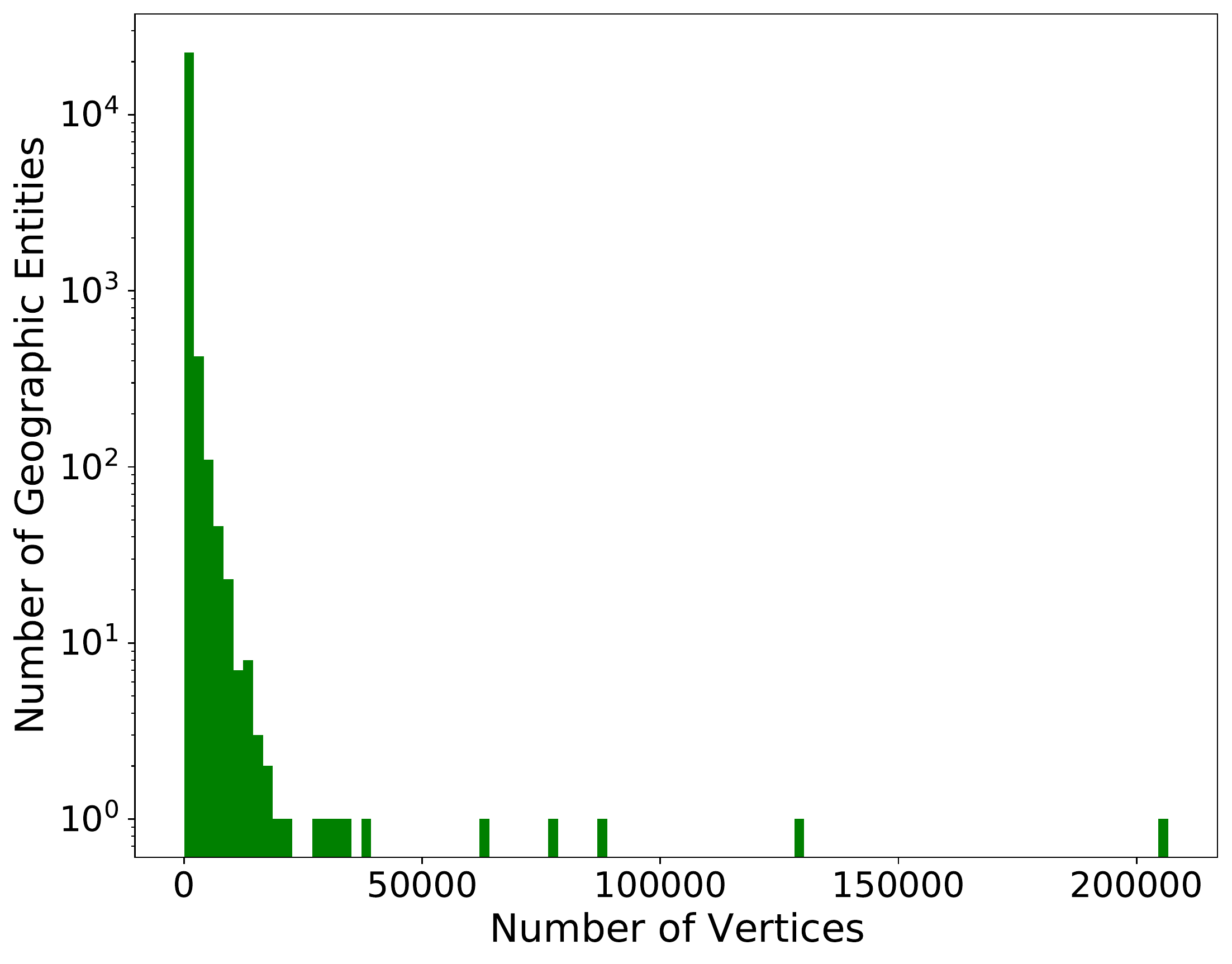}\caption[]{\rvtwo{\# Vertices
		}}    
		\label{fig:dbtopo_vert_hist}
	\end{subfigure}
	\hfill
	\begin{subfigure}[b]{0.329\textwidth}  
		\centering 
		\includegraphics[width=\textwidth]{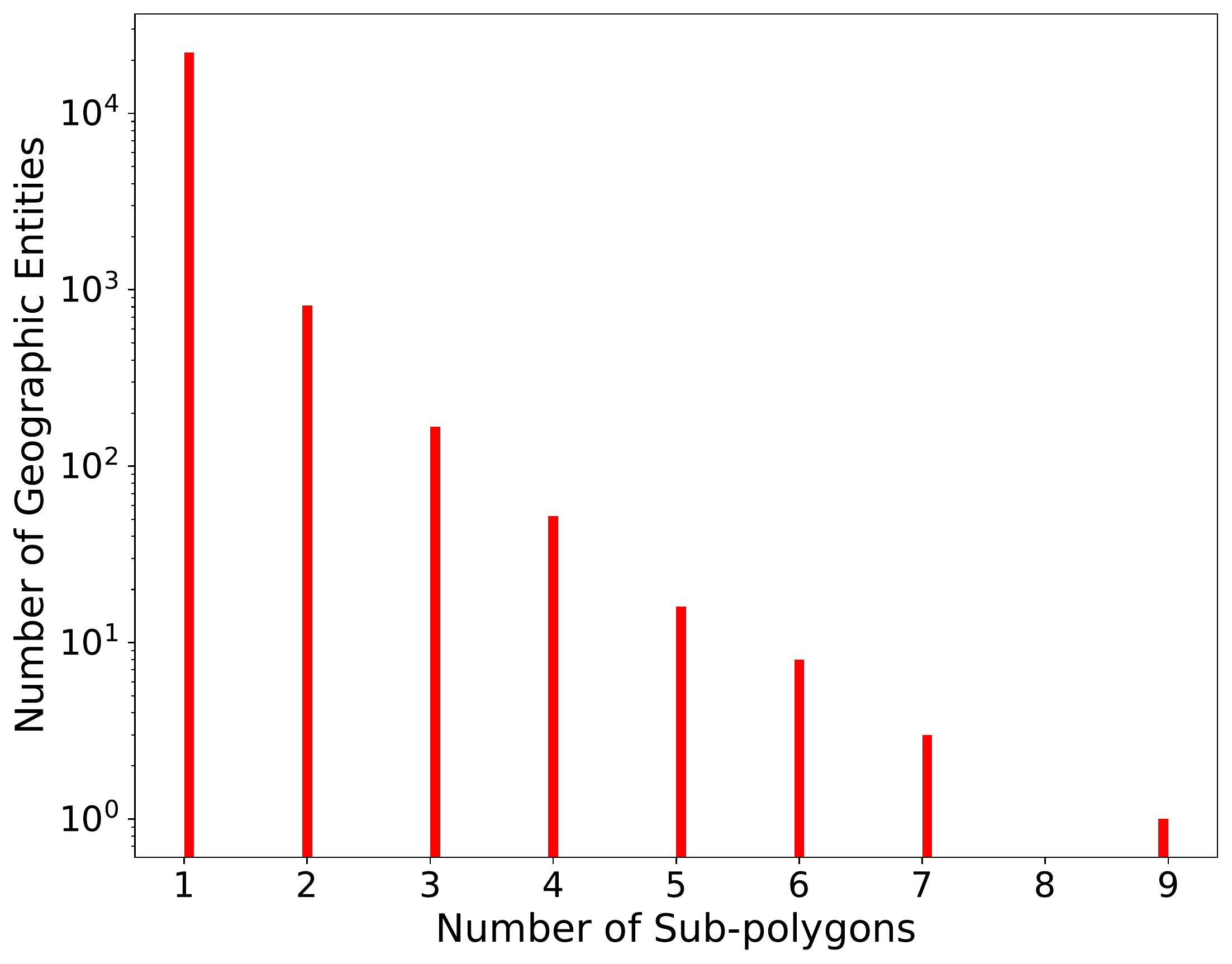}\caption[]{\rvtwo{\# Sub-polygons
		}}    
		\label{fig:dbtopo_part_hist_clx}
	\end{subfigure}
	\hfill
	\begin{subfigure}[b]{0.329\textwidth}  
		\centering 
		\includegraphics[width=\textwidth]{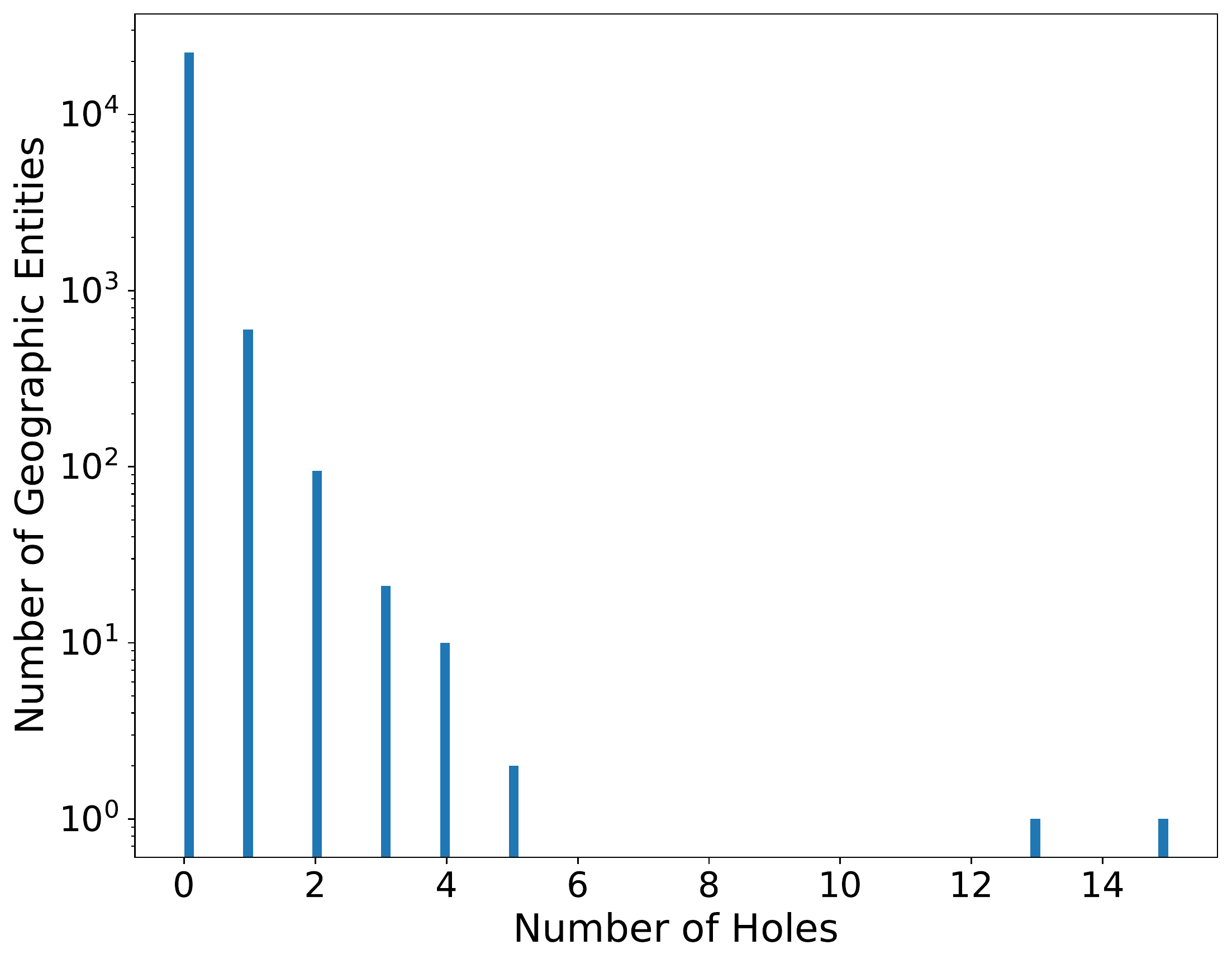}\caption[]{\rvtwo{\# Holes
		}}    
		\label{fig:dbtopo_hole_hist_clx}
	\end{subfigure}
	\hfill
	\begin{subfigure}[b]{0.329\textwidth}  
		\centering 
		\includegraphics[width=\textwidth]{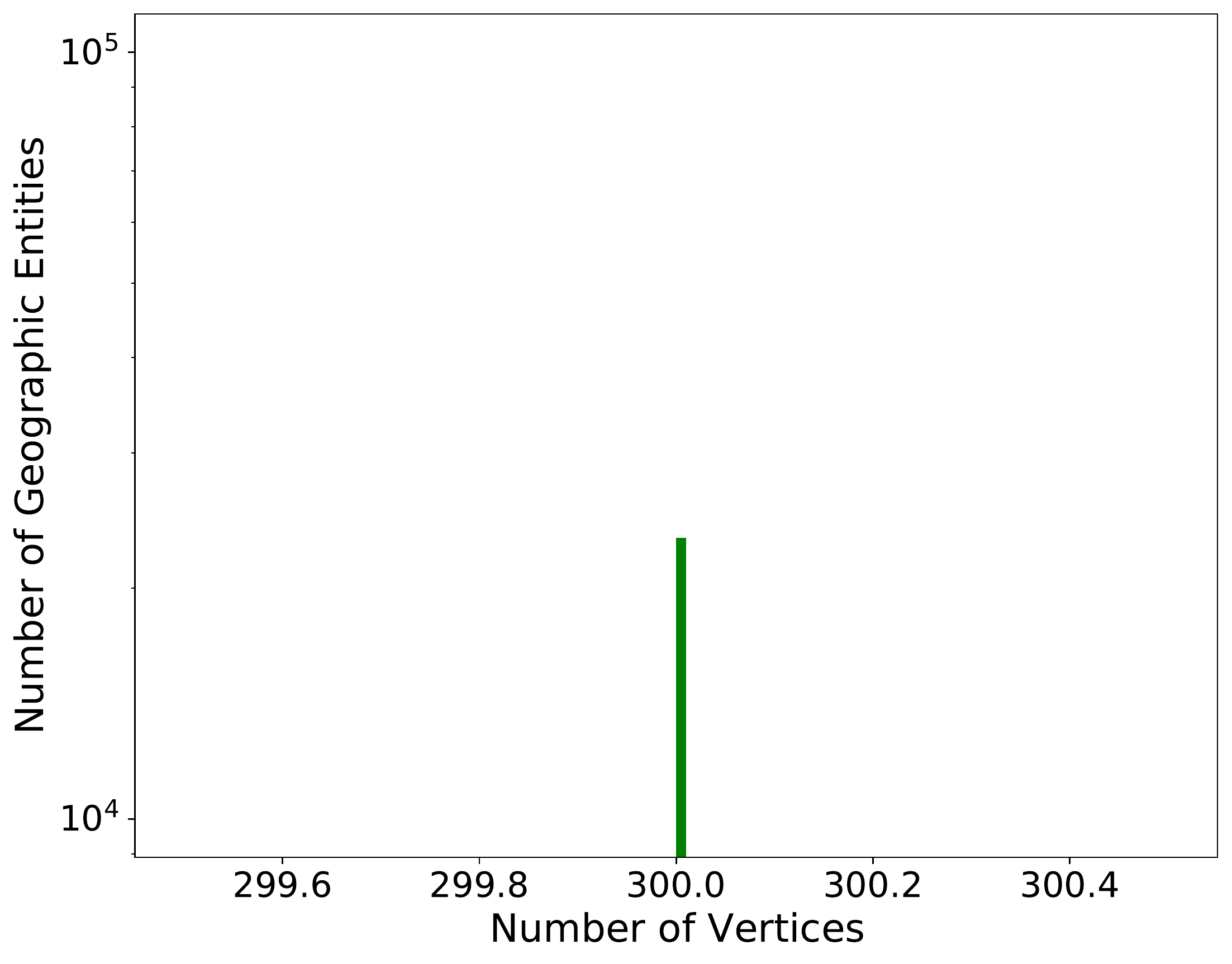}\caption[]{\rvtwo{\# Vertices
		}}    
		\label{fig:dbtopo_vert_hist_clx}
	\end{subfigure}
	\caption{\rvtwo{
	A statistic of the complexity of the raw polygonal geometries as well as the simplfied geometries in \dbtopocomplex~ from OpenStreetMap. (a)-(c) indicate the statistics on the raw polygonal geometries retrieved from OpenStreetMap while (d)-(f) are the same statistics on \dbtopocomplex.
	(a) \& (d) A histogram of the number of sub-polygons per geographic entities.
	(b) \& (e) A histogram of the total number of holes per geographic entities.
	(c) \& (f) A histogram of the total number of unique vertices on each polygonal geometry's exterior and interiors per geographic entities.}
	} 
	\label{fig:dbtopo_geom_hist}
\end{figure}

\begin{figure}[!ht]
	\centering 
	\includegraphics[width=1\textwidth]{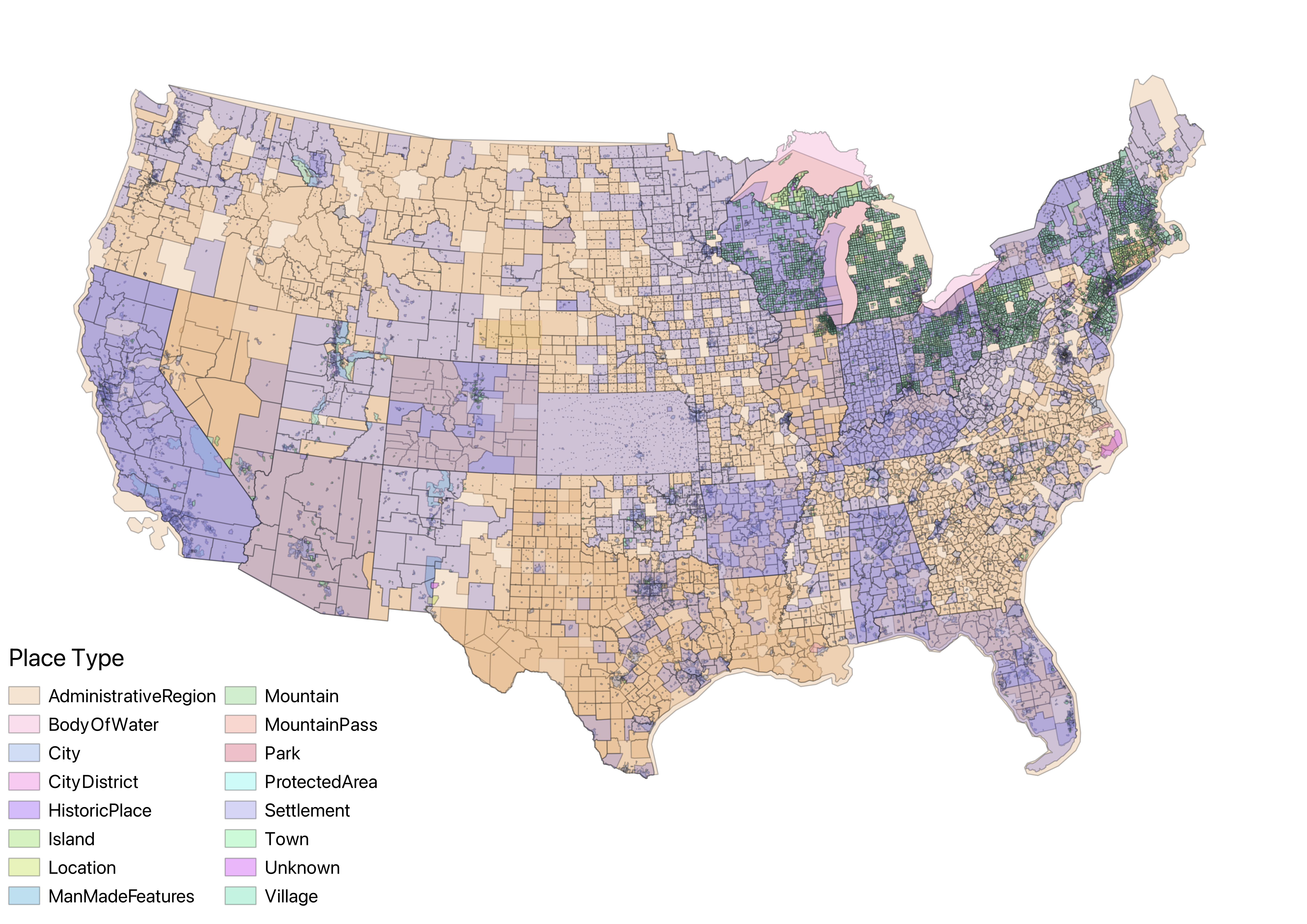}\vspace*{-0.1cm}
	\caption[]{\rvtwo{A map of geographic entities in \dbtopo~ dataset. Some geographic entities' spatial footprints overlays with each other.
	}}    
	\label{fig:place_type_map}
\end{figure}

Since there is no existing benchmark dataset available for this task, we construct two real-world datasets - \dbtopo~and \dbtopocomplex- based on DBpedia Knowledge Graph as well as OpenStreetMap. \dbtopo~and \dbtopocomplex~use the same entity set $\entset$ and triple set $\tripleset$ and the only different is that \dbtopo~uses simple polygons as entities' spatial footprints while \dbtopocomplex~allows complex polygonal geometries. 
The dataset construction steps are described as below:
\begin{enumerate}
    \item We first select a meaningful set of properties $\relset = \{\relat_{\relidx}\}_{\relidx=1}^{\relnum}$ from DBpedia 
    \reviseone{representing }
different spatial relations. 
    \item Then we collect all triples $\{(\ent_{\subject}, \relat_{\relidx}, \ent_{\object})\}$ from DBpedia whose relation $\relat_{\relidx} \in \relset$ and $\ent_{\subject}$, $\ent_{\object}$ are geographic entities. 
    \item Next, we filter out triples whose subject $\ent_{\subject}$ or object $\ent_{\object}$ is located outside the \reviseone{contiguous US. }The resulting triple set $\tripleset = \{(\ent_{\subject}, \relat_{\relidx}, \ent_{\object})\}$ forms a sub-graph of DBpedia with the entity set $\entset$. 
    \item For each entity $\ent \in \entset$, we obtain corresponding Wikidata ID by using \textit{owl:sameas} links.
    \item With each entity $\ent$'s Wikidata ID, we can obtain \reviseone{its polygonal geometry}
    from OpenStreetMap by using Overpass API\footnote{\url{https://wiki.openstreetmap.org/wiki/Overpass_API}}.
    \item The raw polygonal geometries from OpenStreetMap are very detail and complex. For example, Keweenaw County, Michigan is represented as a multipolygon that is consist of 462 sub-polygons. Lake Superior has in total 3130 holes and 206661 vertices in its polygon representation. United States has 128873 vertices. Figure \ref{fig:dbtopo_part_hist}-\ref{fig:dbtopo_vert_hist} \rvtwo{show} statistic about the complexity of these raw geometries.
We simplify those polygonal geometries collected from OpenStreetmap and make two datasets: \dbtopo~and \dbtopocomplex.
    \item As for \dbtopo, we delete all holes and only keep one single simple polygon with the largest area as the geometry for each geographic entity. We also simplify the exterior of each simple polygon \reviseone{by using the Douglas-Peucker algorithm} such that they all have 300 unique vertices. For those polygons \reviseone{with} less than 300 vertices, we do a equal distance interpolation on the exteriors to upsample the vertices to 300. The reason to do so is that same number of vertices \reviseone{make it possible for mini-batching.}
    \item As for \dbtopocomplex, we also simplify the polygonal geometries but we keep holes and multipolygons if necessary. We delete holes if their area are less than 2.5\% of the total area of their corresponding polygonal geometries. Figure \ref{fig:dbtopo_part_hist_clx}-\ref{fig:dbtopo_vert_hist_clx} show the statistics (number of holes, multipolygons, vertices) of the simplified geometries. 
    \reviseone{Similarly, we make the simplified polygonal geometries to have 300 vertices, $\numpgonpt{}=300$, for mini-batching. }
\dbtopo~and \dbtopocomplex~use the same entity set $\entset$ and triple set $\tripleset$ and the only \reviseone{difference} between them is the polygonal geometry of each entity.
    \item Table \ref{tab:dbsparel_type} in Appendix \ref{sec:place_type_map} shows the number of entities with different place types in \dbtopo~and \dbtopocomplex. Figure \ref{fig:place_type_map} shows the polygonal geometries of these geographic entities in these datasets. 
    \item We split $\tripleset$ into \rvtwo{training/validation/testing} dataset by roughly 80:5:15. By following the traditional transductive knowledge graph embedding literature \citep{bordes2013translating,mai2020se}, we make sure all entities \reviseone{in $\entset$} appear in the training dataset. 
    \reviseone{We construct balanced validation and testing dataset with respect to each spatial relation.}
Table \ref{tab:dbsparel_split} shows the statistic of the dataset split. 
\reviseone{In the training dataset,} the balance of triples with different spatial relations can not be achieved at the same time with the need to include all entities in the graph. \reviseone{So in} training dataset we have far more \textit{dbo:isPartOf} triples than other spatial relations. \reviseone{This phenomenon is caused by the nature of DBpedia. So we need to keep this imbalance. }

    \item The spatial relation prediction task does not \reviseone{need} the absolute position of each geographic entity but focuses on their relative spatial relation. So given a triple $(\ent_{\subject}, \relat_{\relidx}, \ent_{\object})$, we first compute the shared minimal bounding box of subject polygon $\geom_{\subject}$ and object polygon $\geom_{\object}$. Then we use this shared bounding box to normalize both $\geom_{\subject}$ and $\geom_{\object}$ into $[-1, 1] \times [-1, 1]$ 2D unit space. Later on, \veercnn~and \resnetoned~directly encodes the normalized polygon geometries for spatial relation prediction. As for \ddsl~and \nuftmlp, since NUFT prefers positive coordinate inputs, we translate those polygons into $[0,2] \times [0,2]$ space before feeding into the NUFT layer.
    
\end{enumerate}

\vspace{-0.5cm}
\begin{table}[!ht]
\caption{The \rvtwo{training/validation/testing} split of \dbtopo/\dbtopocomplex.
	}
	\label{tab:dbsparel_split}
	\centering
\begin{tabular}{l|c|c|c|c}
\toprule
Relation      & All   & Train   & Valid  & Test    \\ \hline
dbo:isPartOf  & 27364 & 26164   & 300    & 900     \\
dbp:north     & 2807  & 1607    & 300    & 900     \\
dbp:east      & 2797  & 1597    & 300    & 900     \\
dbp:south     & 2770  & 1570    & 300    & 900     \\
dbp:west      & 2759  & 1559    & 300    & 900     \\
dbp:northwest & 2063  & 863     & 300    & 900     \\
dbp:southeast & 2024  & 824     & 300    & 900     \\
dbp:southwest & 2000  & 904     & 274    & 822     \\
dbp:northeast & 1994  & 1175    & 205    & 614     \\ \hline
All           & 46578 & 36263   & 2579   & 7736    \\ 
ratio         & 100\% & 77.85\% & 5.54\% & 16.61\% \\
\bottomrule
\end{tabular}
\end{table}

\subsection{Baselines and Model Variations} \label{sec:spa_rel_baselines}

We keep the spatial relation prediction module in Equation \ref{equ:spa_rel} the same but vary the polygon encoders we use. Two exceptions are the \determin~and \ddsllenet. \determin is based on deterministic (no-learning) spatial computation and \ddsllenet~ uses LeNet for spatial relation prediction.
All models we used for the spatial relation prediction tasks are listed as below:
\begin{enumerate}
    \item \textbf{\determin}: We implement a deterministic baseline based on RCC8 topological operators and cardinal direction operations. First, given a triple $(\ent_{\subject}, \relat_{\relidx}, \ent_{\object})$, if the geometry of $\ent_{\subject}$ is inside of the geometry of $\ent_{\object}$, \determin~gives \textit{dbo:isPartOf} as the prediction. Otherwise, \determin~\reviseone{computes} the geometric center of the subject and object geometry and compute their cardinal direction. This is how the current SQL or GeoSPARQL-based geographic question answering models \citep{chen2014parameterized,punjani2018template} \reviseone{use} to answer spatial relation questions.
    
    \item \textbf{\veercnn}: We use \veercnn~to encode both subject and object polygons and feed them into
the spatial relation prediction model \reviseone{(Equation \ref{equ:spa_rel}).}

    \item \textbf{\ddsllenet}~\citep{jiang2019ddsl}: We utilize the original DDSL implementation\footnote{\url{https://github.com/maxjiang93/DDSL}} and modify it for mini-batch training. \ddsllenet~does not include the spatial relation prediction model. Since the output of NUFT-IFFT of a polygonal geometry is an image, we concatenate the output images of the subject and object geometries \reviseone{in} the channel dimension and feed it into the LeNet5, one type of 2D CNN, for spatial relation prediction. 
    
    \item \textbf{\ddsl+MLP}: Unlike \ddsllenet, given a polygonal geometry, \ddsl+MLP first flattens the NUFT-IFFT output image into 1D and feeds it into an MLP to encode it \reviseone{into} an embedding. The embeddings of the subject and object geometries are concatenated and fed into another MLP as described in Equation \ref{equ:spa_rel}.
    
    \item \textbf{\resnetoned}: We implement our \resnetoned~as described in Section \ref{sec:resnet1d} and utilize it in the spatial relation prediction model. 

    \item \textbf{\nuftmlp[\linearfreqmtd]+MLP}: This version of \nuftmlp~is the same as what we described in Section \ref{sec:shp_cla_model}. It uses a linear grid $\fftfreqmat^{(\linearfreqmtd)}$ for NUFT followed by an MLP to encode the polygon into an embedding. The only difference is that it concatenates the embeddings of subject and object geometries and uses them in the spatial relation prediction model (Equation \ref{equ:spa_rel}).
    
    \item \textbf{\nuftmlp[\geometricfreqmtd]+MLP}: Different from \nuftmlp[\linearfreqmtd]+MLP, this version of \nuftmlp~uses the geometric grid $\fftfreqmat^{(\geometricfreqmtd)}$.
\end{enumerate}

Among these seven models, the first \rvtwo{four} models are baseline approaches while \resnetoned, \nuftmlp[\linearfreqmtd]+MLP, and \nuftmlp[\geometricfreqmtd]+MLP are polygon encoders we proposed. \rvtwo{In order to allow \veercnn~and \resnetoned~to handle complex polygonal geometries in the \dbtopocomplex~dataset, we perform the same boundary concatenation operation as described in Section \ref{sec:shp_cla_model}.}

\begin{table}[h!]
\caption{\rvtwo{Spatial relation prediction accuracy}
on the \dbtopo~\rvtwo{(simple polygon) }and \dbtopocomplex~\rvtwo{(complex polygon) datasets}.
	}
	\label{tab:dbsparel_eval}
	\centering
\begin{tabular}{l|c|c|c|c|c|c}
\toprule
              & \multicolumn{3}{c|}{\dbtopo} & \multicolumn{3}{c}{\dbtopocomplex} \\
\hline
              & Train   & Valid   & Test   & Train     & Valid     & Test      \\ \hline
\determin                  & 75.42          & 75.18          & 73.80          & 75.17          & 75.30          & 73.90          \\
\veercnn~\citep{veer2018deep}   & 92.10          & 77.90          & 77.59          & 91.60          & 77.51          & 77.08          \\
\ddsllenet~\citep{jiang2019convolutional,jiang2019ddsl}                    & \reviseone{93.19}          & \reviseone{80.27}          & \reviseone{79.22}          & \reviseone{93.50}          & \reviseone{80.11}          & \reviseone{78.68}          \\ 
\ddsl+MLP                       & \textbf{95.23} & 78.32          & 78.63          & 92.89          & 79.64          & 79.78          \\ \hline
\resnetoned                     & \reviseone{91.98}          & \reviseone{78.13}          & \reviseone{77.79}          & \reviseone{92.02}         & \reviseone{78.52}         & \reviseone{78.24}          \\
\nuftmlp[\linearfreqmtd]+MLP    & 93.02          & 80.85          & 79.20          & 93.77          & 79.33          & 79.12          \\
\nuftmlp[\geometricfreqmtd]+MLP & 93.41          & \textbf{80.92} & \textbf{79.80} & \textbf{94.63} & \textbf{81.04} & \textbf{80.44} \\
\bottomrule
\end{tabular}
\end{table}

\subsection{Main Evaluation Results}  \label{sec:spa_rel_eval_res}
We evaluates all models described in Section \ref{sec:spa_rel_baselines} on our \dbtopo~and \dbtopocomplex~dataset. Table \ref{tab:dbsparel_eval} shows the overall performance (classification accuracy) of different models on the training, validation, and \rvtwo{testing} set of \dbtopo~as well as \dbtopocomplex~dataset. 
\rvtwo{
The hyperparameter tuning detailed can be seen in Appendix \ref{sec:para_tune} and the best hyperparameter combinations for different models on two datasets are listed in Table \ref{tab:dbtopo_hyperpara}.}
From Table \ref{tab:dbsparel_eval}, we can see that:
\begin{enumerate}
    \item \nuftmlp[\geometricfreqmtd]+MLP achieves the best performance on the validation and \rvtwo{testing} set of both \dbtopo~and \dbtopocomplex. It outperforms \nuftmlp[\linearfreqmtd]+MLP, \resnetoned~as well as all four baseline models. This demonstrates the effectiveness and robustness of our \nuftmlp~model.
    
     \item \reviseone{
On both datasets, \resnetoned~outperforms \determin~and \veercnn, but still falls behind \ddsllenet, \ddsl+MLP, \nuftmlp[\linearfreqmtd]+MLP, and \nuftmlp[\geometricfreqmtd]+MLP. 
    This might because that both \veercnn~and \resnetoned~only encode the polygon boundary information but ignore the polygon topological information (i.e., \rvtwo{distinguishing the exterior from the interiors of a polygon}) which is very important for spatial relation prediction (e.g., dbo:isPartOf). Our qualitative analysis in Section \ref{sec:spa_rel_rs_vs_nuft} confirms this assumption.
    }

    \item Similar to the experiment results on \mnistcomplex, \nuftmlp[\geometricfreqmtd]+MLP outperforms \nuftmlp[\linearfreqmtd]+MLP on both \dbtopo~and \dbtopocomplex. This demonstrates that a good choice of the Fourier frequency map \reviseone{can help improve the model performance. } Since \ddsl~requires an IFFT, the non-integer geometric grid frequency map $\fftfreqmat^{(\geometricfreqmtd)}$ is no longer applicable. This also shows the flexibility of \nuftmlp~in terms of frequency choice.

   \item \reviseone{The original \ddsllenet~model
\citep{jiang2019ddsl} 
   and \ddsl+MLP share the same NUFT-IFFT layer while using different learnable component\rvtwo{s} -- LeNet5 v.s. MLP. We can see that \ddsllenet~can outperform \ddsl+MLP on \dbtopo~dataset but underperform \ddsl+MLP on \dbtopocomplex~\rvtwo{testing} dataset. 
  This shows that when predicting spatial relations between complex polygonal geometries, using more complex learnable model (e.g., LeNet5) might not be helpful.
}

    \item \nuftmlp[\geometricfreqmtd]+MLP \reviseone{outperforms} \ddsl+MLP 
    by 2.6\% and 1.2\% on the \reviseone{validation} and \rvtwo{testing} set of \dbtopo (1.4\% and 0.7\% for \dbtopocomplex). 
    Given the fact that they are using the same MLP-based spatial relation module, 
    \reviseone{these performance improvements \rvtwo{are mainly attributed} to}
the advantages of \nuftmlp: learning polygon representations directly from the NUFT spectral features enables us to have more flexible choices of NUFT frequency map $\fftfreqmat$. This allows us to design $\fftfreqmat$ that can effectively capture irregular polygon representations.
\end{enumerate}

\subsection{Ablation Study on \resnetoned}  \label{sec:resnetoned_abla_std}

\begin{table}[h!]
\caption{Ablation study of \resnetoned~on \dbtopo~dataset.
	}
	\label{tab:dbsparel_eval_resnet_alb}
	\centering
	\reviseone{
\begin{tabular}{l|l|l|l}
\toprule
                      & \multicolumn{3}{c}{\dbtopo} \\ \hline
                      & Train   & Valid   & Test   \\ \hline
\resnetoned                & 91.98   & 78.13   & 77.79  \\
\resnetoned (zero padding) & 90.67   & 74.49   & 73.07  \\
\resnetoned (raw pt) & 91.34   & 75.92   & 75.35  \\
\bottomrule
\end{tabular}
}
\end{table}

Despite the fact that \resnetoned~is similar to \veercnn~in the sense that both of them use 1D CNN layers to model polygons' coordinate sequences, \resnetoned~outperforms \veercnn~on both \dbtopo~and \dbtopocomplex~dataset. To understand the superiority of \resnetoned, we do an ablation study \reviseone{which is } shown in Table \ref{tab:dbsparel_eval_resnet_alb}. It shows that when we replace the circular padding with zero padding -- \resnetoned (zero padding), the performance of \resnetoned~drops \reviseone{3.64\% and 4.72\% on the validation and \rvtwo{testing} set. }This \reviseone{demonstrates} that circular padding is critical for \resnetoned~since it preserves the loop \rvtwo{origin} invariance property.
A similar situation happens when we delete the \kdeltaenc~point encoder component and direct feed the raw point coordinates to ResNet layers -- \resnetoned (raw pt). This demonstrates that \kdeltaenc~ 
\rvtwo{is very helpful for polygon encoding and spatial relation prediction since it uses the spatial affinity features (Equation \ref{equ:resnet1d_ptemb}) to enrich the point embedding with its neighborhood information.}

\subsection{The Impact of $\fftfreqnumX$}  \label{sec:spa_rel_nuftfreq_study}

\begin{figure*}[ht!]
	\centering \tiny
	\vspace*{-0.2cm}
\begin{subfigure}[b]{0.49\textwidth}  
		\centering 
		\includegraphics[width=\textwidth]{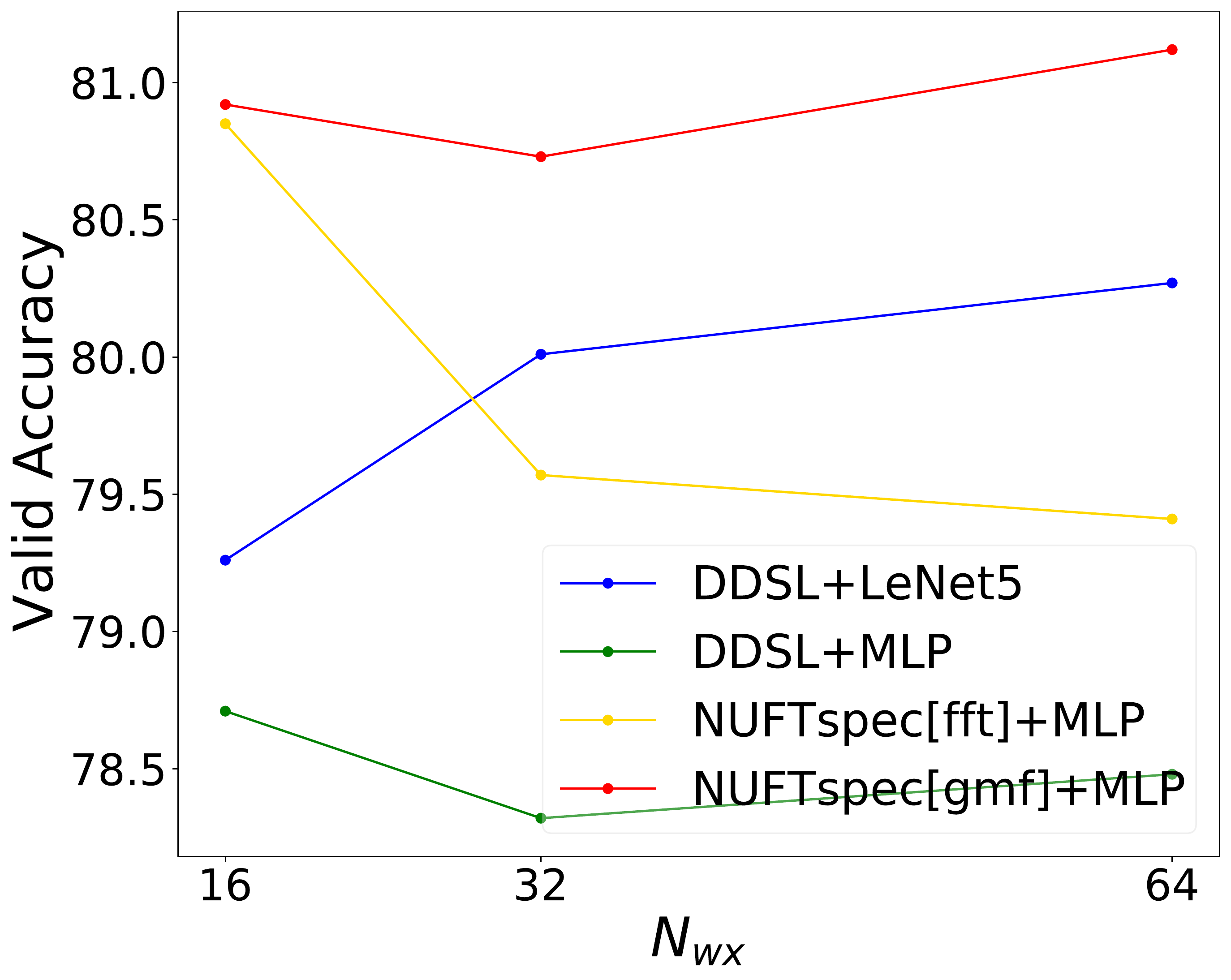}\vspace*{-0.2cm}
		\caption[]{\reviseone{\dbtopo~Validation Set
		}}    
		\label{fig:dbtopo_numfreq_valid}
	\end{subfigure}
	\hfill
	\begin{subfigure}[b]{0.49\textwidth}  
		\centering 
		\includegraphics[width=\textwidth]{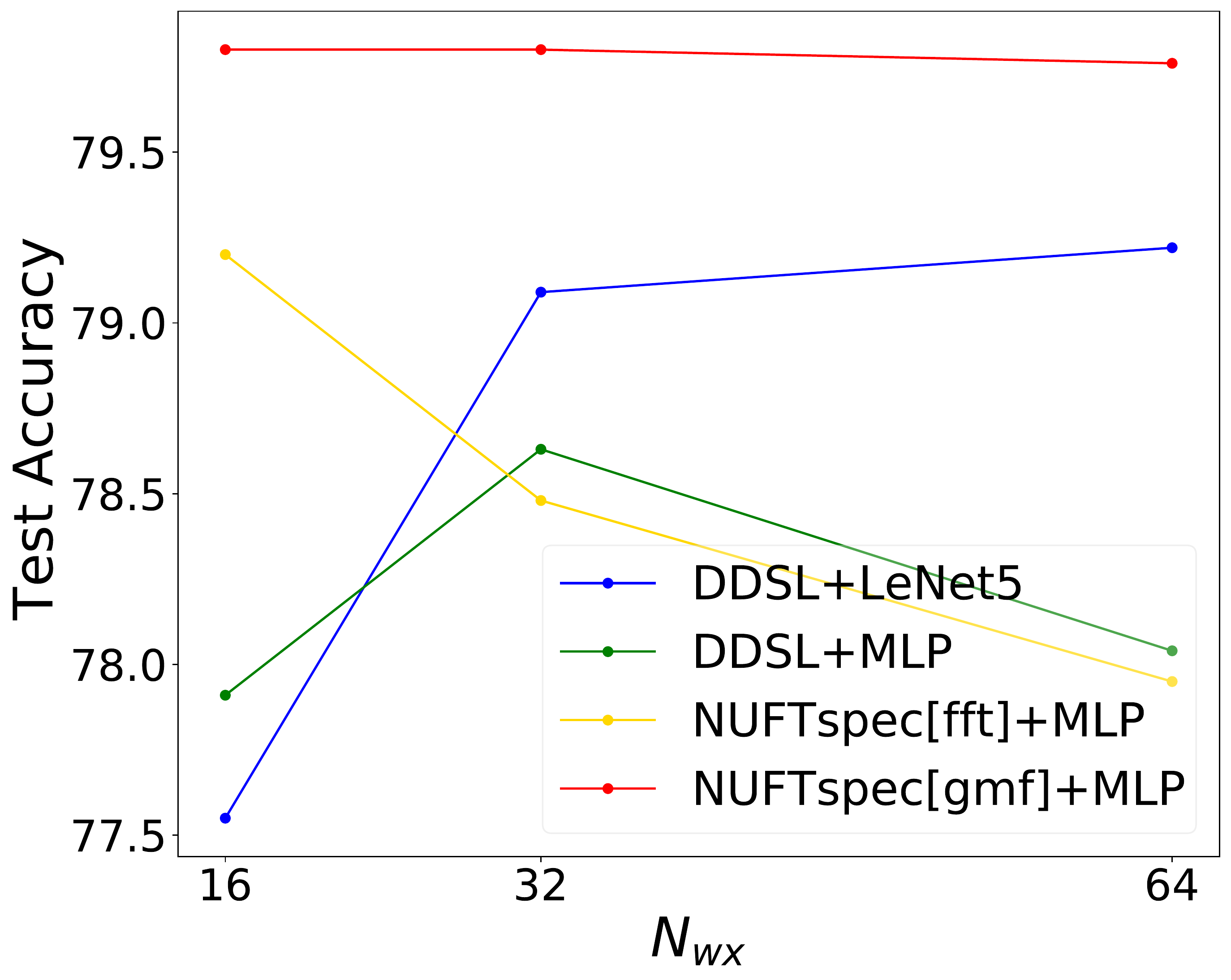}\vspace*{-0.2cm}
		\caption[]{\reviseone{\dbtopo~Testing Set
		}}    
		\label{fig:dbtopo_numfreq_test}
	\end{subfigure}
	\hfill
\begin{subfigure}[b]{0.49\textwidth}  
		\centering 
		\includegraphics[width=\textwidth]{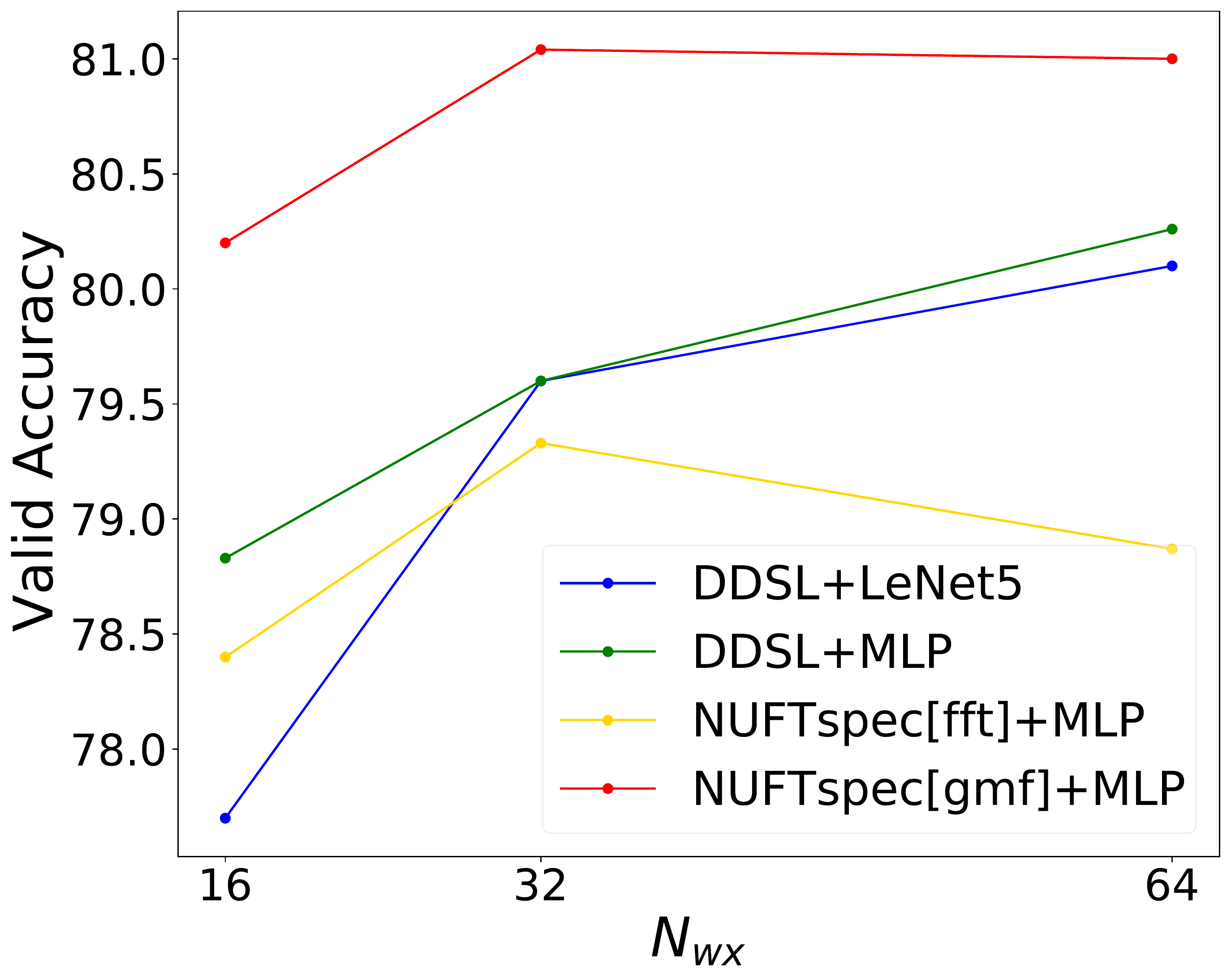}\vspace*{-0.2cm}
		\caption[]{\reviseone{\dbtopocomplex~Validation Set
		}}    
		\label{fig:dbtopocomplex_numfreq_valid}
	\end{subfigure}
	\hfill
	\begin{subfigure}[b]{0.49\textwidth}  
		\centering 
		\includegraphics[width=\textwidth]{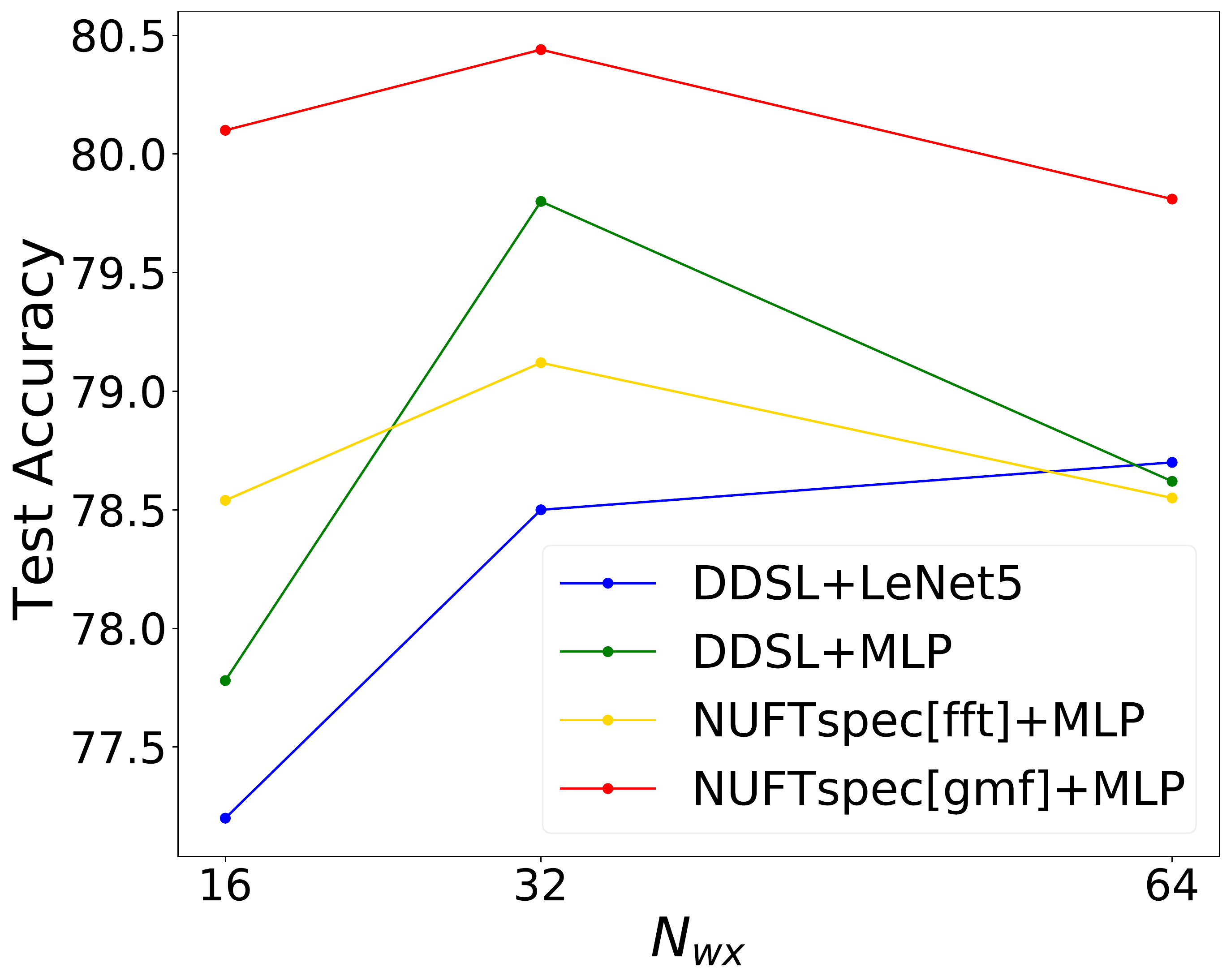}\vspace*{-0.2cm}
		\caption[]{\reviseone{\dbtopocomplex~Testing Set
		}}    
		\label{fig:dbtopocomplex_numfreq_test}
	\end{subfigure}
	\caption{
	\reviseone{
	Performance comparisons among four NUFT-based polygon encoder models on the validation and testing sets of \dbtopo~and \dbtopocomplex~with different numbers of NUFT frequencies used in X - $\fftfreqnumX$. Here, $\fftfreqnumX \propto \fftfreqnumY \propto \nscale \propto \sqrt{\fftfreqnum}$.}
	} 
	\label{fig:spa_rel_numfreq}
\end{figure*}

Among the 7 models listed in Section \ref{sec:spa_rel_baselines}, there are 4 NUFT-based models. In this section, we test the robustness of these 4 models on \dbtopo~and \dbtopocomplex~when we vary the number of frequencies we use in NUFT. Here, we use $\fftfreqnumX$ to indicate the used NUFT frequency numbers since $\fftfreqnumX \propto \fftfreqnumY \propto \nscale \propto \sqrt{\fftfreqnum}$.
Figure \ref{fig:spa_rel_numfreq} compares the model performance of them with different $\fftfreqnumX$ in the validation and testing dataset of \dbtopo~and \dbtopocomplex. 
We can see that \nuftmlp[\geometricfreqmtd]+MLP achieves the best performance on all these four datasets with any given $\fftfreqnumX$ \reviseone{ and show a clear advantage over the other three models. 
}

\subsection{\reviseone{Analysis of \rvtwo{The} Sliver Polygon Problem}} \label{sec:sliver_pgon_analysis}

\reviseone{
To investigate the difficulty of polygon-based spatial relation prediction task and how the performance can be affected by sliver polygons, we compute the topological relations between the subject and object entity of each triple in \dbtopo~based on a deterministic RCC8-based spatial operator\footnote{\url{https://shapely.readthedocs.io/en/stable/manual.html\#binary-predicates}}. The statistics is shown in Table \ref{tab:dbsparel_deter}. We can see almost all triples with cardinal direction relations (the last 8 relations) have subject and object entities that intersect or are disjoint with each other. 
Interestingly, for \textit{dbo:isPartOf} relation, in 76.5\% of the 27,364 triples the subject polygons are inside the corresponding object polygons \reviseone{whereas in 6309 triples (23\%+) the subject and object polygons intersect with each other when computing their relations deterministically. } Based on manual inspection, most of those `intersects' cases are caused by the \textit{sliver polygon} problem shown in Figure \ref{fig:sliver_pogon}. This clearly indicates the necessity of polygon encoding models.
}

\begin{table}[t]
\caption{
\reviseone{Relation classification result on \dbtopo~based on the deterministic RCC8 spatial operator.}
	}
	\label{tab:dbsparel_deter}
	\centering
	\reviseone{
\begin{tabular}{l|c|c|c|c}
\toprule
Relation      & Contains & Intersects & Touches & Disjoint \\ \hline
dbo:isPartOf  & 20,923    & 6,309       & 0       & 132      \\
dbp:north     & 4        & 2,451       & 0       & 352      \\
dbp:east      & 2        & 2,447       & 0       & 348      \\
dbp:south     & 5        & 2,405       & 0       & 360      \\
dbp:west      & 3        & 2,408       & 0       & 348      \\
dbp:northwest & 4        & 1,644       & 0       & 415      \\
dbp:southeast & 2        & 1,618       & 0       & 404      \\
dbp:southwest & 4        & 1,622       & 0       & 374      \\
dbp:northeast & 3        & 1,634       & 0       & 357      \\
\bottomrule
\end{tabular}
}
\end{table}

\begin{figure}[b]
  \begin{minipage}[c]{0.5\linewidth}
    \centering
    \includegraphics[width=1.0\textwidth]{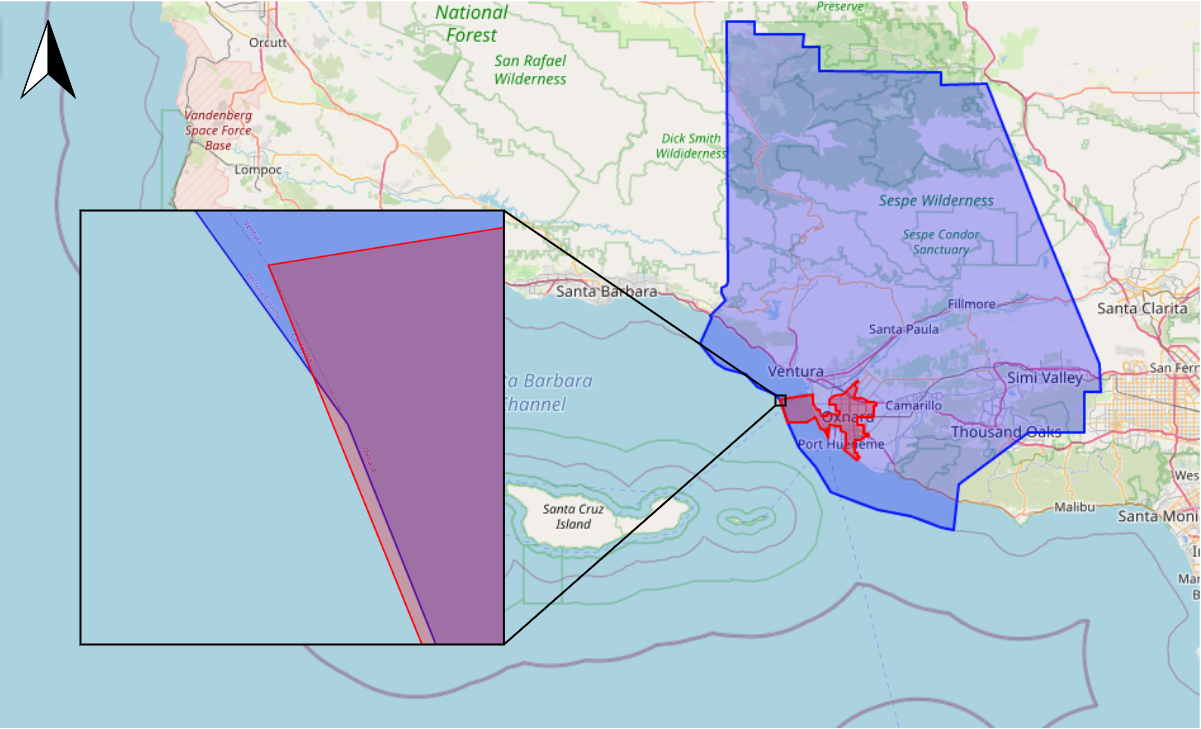}
  \end{minipage}\hspace{0.2cm}
  \begin{minipage}[c]{0.30\linewidth}
    \centering
{\setlength{\tabcolsep}{0.3pt} \small
\begin{tabular}{l|c}
\toprule
Model                 & Prediction    \\ \hline
\determin         & dbp:northeast \\
\ddsl+MLP              & dbp:northeast \\
\resnetoned              & dbo:isPartOf  \\
\nuftmlp[\geometricfreqmtd]+MLP & dbo:isPartOf  \\ \hline
Ground Truth          & dbo:isPartOf \\
\bottomrule
\end{tabular}
}
\end{minipage}
\caption{
\reviseone{One example in \dbtopocomplex~\rvtwo{testing} set to show how the sliver polygons can affect the the prediction results of different polygon encoders. 
Red geometry (subject): \textit{dbr:Oxnard,\_California};
Blue geometry (object): \textit{dbr:Ventura\_County,\_California}.
The table shows the predictions of different polygon encoders on this example.
}
}
\label{fig:sliver_pgon_eval_39293}
\end{figure}

\begin{figure}
  \begin{minipage}[c]{0.5\linewidth}
    \centering
    \includegraphics[width=1.0\textwidth]{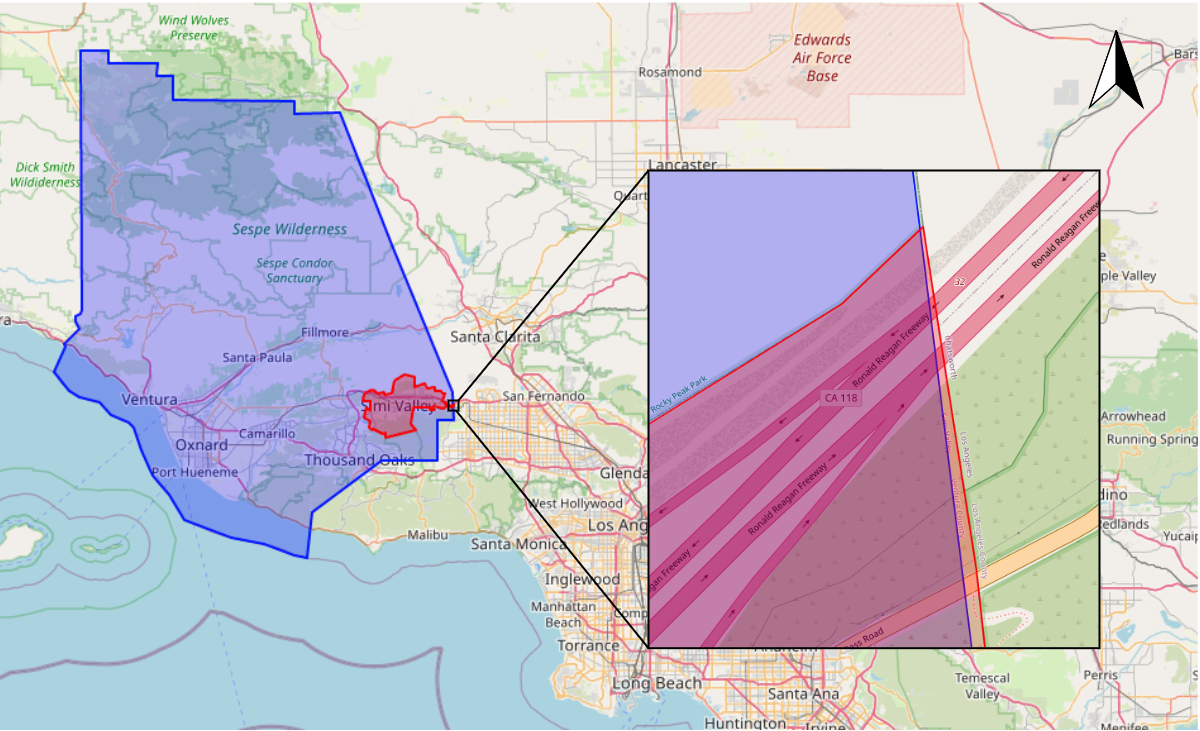}
  \end{minipage}\hspace{0.2cm}
  \begin{minipage}[c]{0.30\linewidth}
    \centering
{\setlength{\tabcolsep}{0.3pt} \small
\begin{tabular}{l|c}
\toprule
Model                 & Prediction    \\ \hline
\determin         & dbp:northwest \\
\ddsl+MLP              & dbp:northeast \\
\resnetoned           & dbo:isPartOf  \\
\nuftmlp[\geometricfreqmtd]+MLP & dbo:isPartOf  \\ \hline
Ground Truth          & dbo:isPartOf  \\
\bottomrule
\end{tabular}
}
\end{minipage}
\caption{
\reviseone{Another example in \dbtopocomplex~\rvtwo{testing} set to demonstrate the affect of the sliver polygon problem similar to Figure \ref{fig:sliver_pgon_eval_39293}. 
Red geometry (subject): \textit{dbr:Simi\_Valley,\_California};
Blue geometry (object): \textit{dbr:Ventura\_County,\_California}.
The table shows the predictions of different polygon encoders on this example.
}
}
\label{fig:sliver_pgon_eval_39356}
\end{figure}

\reviseone{To qualitatively show how our proposed polygon encoders mitigate the sliver polygon problem, we show two examples in Figure \ref{fig:sliver_pgon_eval_39293} and \ref{fig:sliver_pgon_eval_39356}. 
In both examples, the \rvtwo{subjects} indicated by the red polygonal \rvtwo{geometries} should be \textit{part of} the object\rvtwo{s}, denoted by the blue polygonal \rvtwo{geometries}. However, because of the sliver polygons which are illustrated in the zoom-in windows, both \determin~and \ddsl+MLP fail to make the correct prediction\rvtwo{s} while both \resnetoned~and \nuftmlp[\geometricfreqmtd]+MLP can predict the correct relations.}

\vspace{-0.3cm}

\subsection{\reviseone{Analysis of \rvtwo{the} Scale Problem}} \label{sec:scale_analysis}

\reviseone{
To evaluate how well different polygon encoders can handle the scale problem shown in Figure \ref{fig:scale}, we evaluate the performance of each models under different area ratio groups. 
For each triple in the \rvtwo{testing} set of \dbtopocomplex, an area ratio $\arearatio$ is computed as the ratio between the areas of the subject geometry and object geometry. $\arearatio$ shows the scale different between the subjects and objects. 
Based on $\arearatio$, \dbtopocomplex~\rvtwo{testing} set are divided into seven different mutually exclusive area ratio groups. Table \ref{tab:scale_eval} shows the performances of different polygon encoders in each area ratio groups. We can see that in all area ratio groups except $\arearatio \in [1, 1.1)$, \nuftmlp[\geometricfreqmtd]+MLP can outperform all other polygon encoders. This indicates that \nuftmlp~is flexible to handle spatial relation prediction task under different area ratio settings.
}

\begin{table}[h!]
\caption{
\reviseone{Evaluation results of different polygon encoders on different area ratio groups of \dbtopocomplex~\rvtwo{testing} set. Given a triple, the area ratio $\arearatio$ is the ratio between the areas of the subject geometry and object geometry. 
}
	}
	\label{tab:scale_eval}
	\centering
	\reviseone{
	\setlength{\tabcolsep}{1.0pt}
\begin{tabular}{l|c|c|c|c|c|c|c}
\toprule
\multirow{2}{*}{Model} & \multicolumn{7}{c}{Area Ratio $\arearatio$}                                                                                          \\ \cline{2-8}
                       & $[0, 0.1)$     & $[0.1, 0.2)$   & $[0.2, 0.3)$   & $[0.3, 1)$     & $[1, 1.1)$     & $[1.1, 1.2)$   & $[1.2, \infty)$ \\ \hline
\# Triples                  & 1027           & 156            & 167            & 3152           & 675            & 411            & 2148              \\ \hline
\determin          & 66.60          & 50.00          & 62.87          & 74.62          & 81.19          & 74.70          & 76.49             \\
\veercnn                & 86.95          & 51.92          & 50.30          & 75.44          & 82.67          & 76.89          & 76.96             \\
\ddsllenet            & 87.93          & 58.33          & 61.68          & 76.84          & 84.44          & 81.75          & 77.37             \\
\ddsl+MLP               & 87.24          & 54.49          & 63.47          & 79.09          & 84.44          & 80.78          & 78.68             \\ \hline
\resnetoned               & 87.05          & 45.51          & 52.69          & 76.78          & 84.59          & 79.81          & 78.26             \\
\nuftmlp[\linearfreqmtd]+MLP  & 87.15          & 57.05          & 61.08          & 77.35          & \textbf{85.63} & 81.75          & 78.35             \\
\nuftmlp[\geometricfreqmtd]+MLP  & \textbf{87.93} & \textbf{59.62} & \textbf{64.07} & \textbf{79.22} & 85.33          & \textbf{83.94} & \textbf{79.24}      \\
\bottomrule
\end{tabular}
}
\end{table}

\subsection{\reviseone{\resnetoned~v.s. \nuftmlp}} \label{sec:spa_rel_rs_vs_nuft}

\begin{figure}
  \begin{minipage}[c]{0.5\linewidth}
    \centering
    \includegraphics[width=1.0\textwidth]{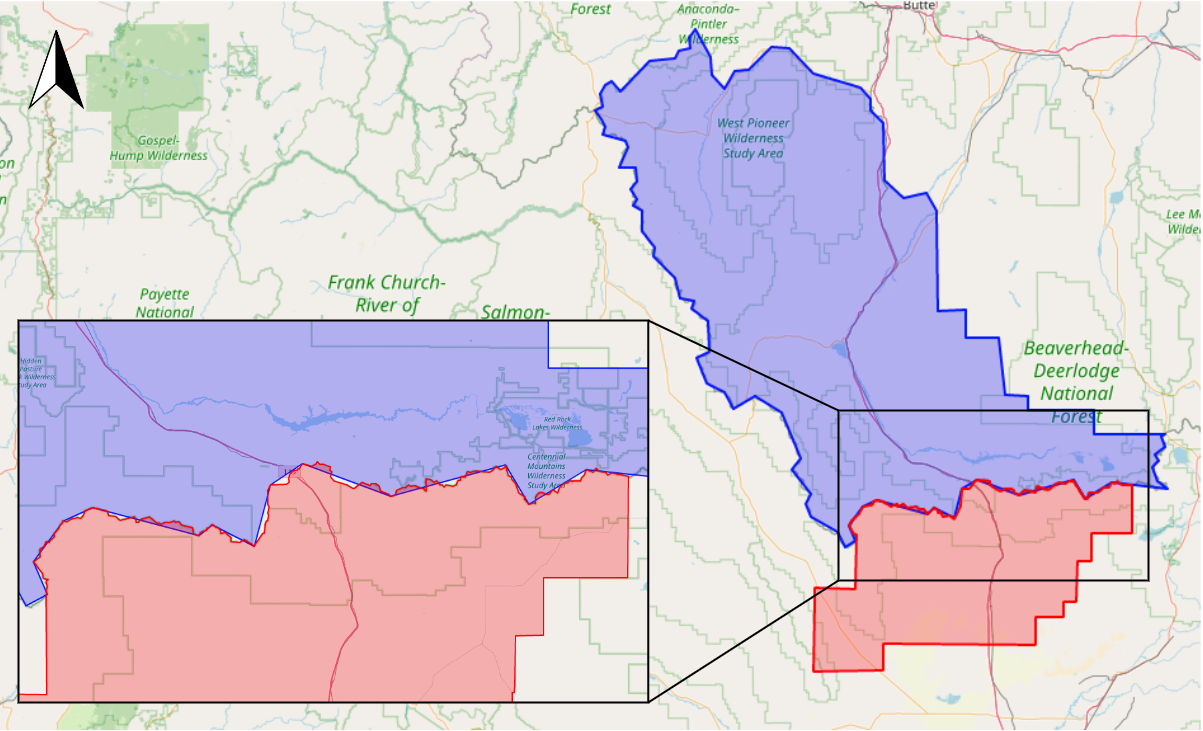}
  \end{minipage}\hspace{0.1in}
  \begin{minipage}[c]{0.30\linewidth}
    \centering
{\setlength{\tabcolsep}{0.3pt} \small
\begin{tabular}{l|c}
\toprule
Model                 & Prediction   \\ \hline
\determin         & dbp:north    \\
\veercnn               & dbo:isPartOf \\
\ddsllenet           & dbp:north    \\
\ddsl+MLP              & dbp:north    \\
\resnetoned              & dbo:isPartOf \\
\nuftmlp[\linearfreqmtd]+MLP & dbp:north    \\ \hline
Ground Truth          & dbp:north    \\
\bottomrule
\end{tabular}
}
\end{minipage}
\caption{
\reviseone{An example in \dbtopo~relation prediction \rvtwo{testing} set to show the drawback of \resnetoned. 
Red geometry (subject): \textit{dbr:Clark\_County,\_Idaho};
Blue geometry (object): \textit{dbr:Beaverhead\_County,\_Montana}.
The table shows the predictions of different polygon encoders for this example.
}
}
\label{fig:sliver_pgon_eval_40486}
\end{figure}

\begin{figure}
  \begin{minipage}[c]{0.5\linewidth}
    \centering
    \includegraphics[width=1.0\textwidth]{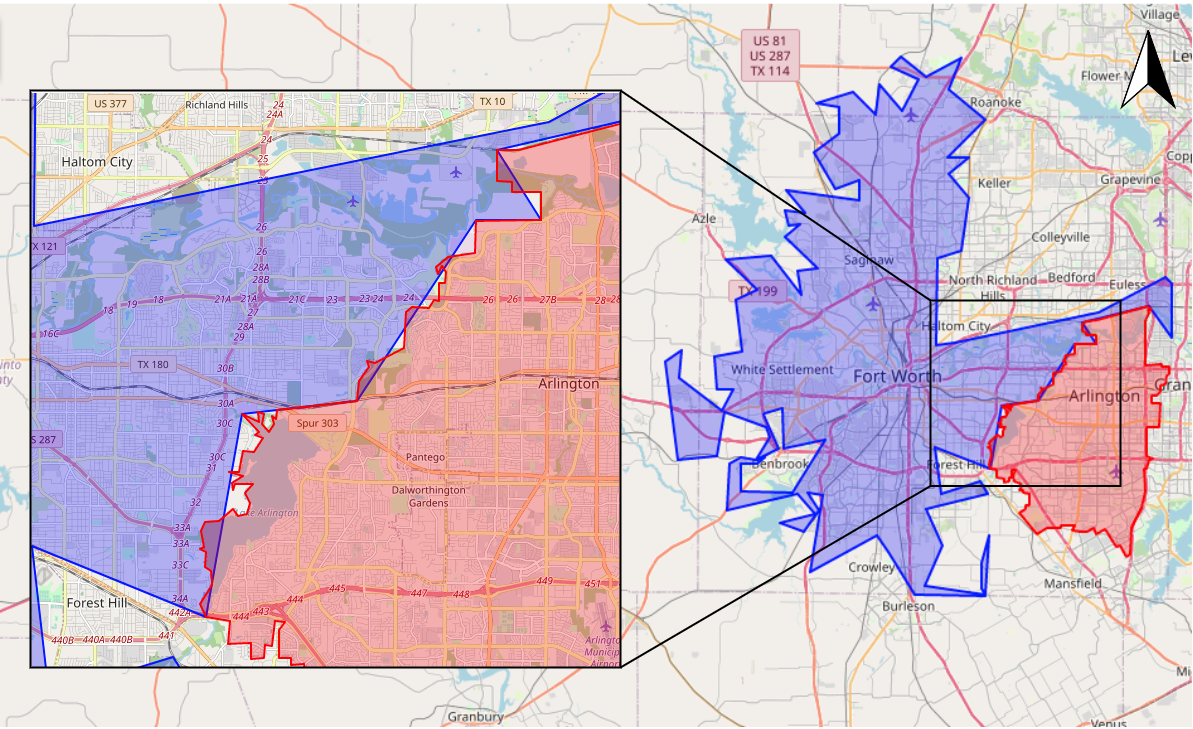}
  \end{minipage}\hspace{0.1in}
  \begin{minipage}[c]{0.30\linewidth}
    \centering
{\setlength{\tabcolsep}{0.3pt} \small
\begin{tabular}{l|c}
\toprule
Model                 & Prediction    \\ \hline
\determin         & dbp:west      \\
\veercnn               & dbo:isPartOf  \\
\ddsllenet           & dbp:northwest \\
\ddsl+MLP              & dbp:northwest \\
\resnetoned              & dbo:isPartOf  \\
\nuftmlp[\linearfreqmtd]+MLP & dbp:west      \\ \hline
Ground Truth          & dbp:west \\
\bottomrule
\end{tabular}
}
\end{minipage}
\caption{
\reviseone{An example in \dbtopo~relation prediction \rvtwo{testing} set to show the drawback of \resnetoned.
Red geometry (subject): \textit{dbr:Arlington,\_Texas};
Blue geometry (object): \textit{dbr:Fort\_Worth,\_Texas}.
The table shows the predictions of different polygon encoders for this example.
}
}
\label{fig:sliver_pgon_eval_42586}
\end{figure}

\reviseone{
Comparing the evaluation results from both tasks, we notice that \resnetoned~can outperform \nuftmlp~on shape classification task but lose on spatial relation prediction. To understand the reason why \resnetoned~fail\rvtwo{s} on the spatial relation prediction task, we explore triples in \dbtopo~\rvtwo{testing} set on which \nuftmlp[\linearfreqmtd]+MLP can make the correct predictions while \resnetoned~fails. Two representative examples are shown in Figure \ref{fig:sliver_pgon_eval_40486} and \ref{fig:sliver_pgon_eval_42586}. In both cases, \nuftmlp[\linearfreqmtd]+MLP and \determin~can correctly predict the spatial relation. However, two spatial domain polygon encoders -- \veercnn~and \resnetoned~-- predict \rvtwo{\textit{dbo:isPartOf}}, although the subject geometry is clearly not part of the object geometry. We guess the reason is that both \veercnn~and \resnetoned~only encode vertex information but are not aware of the topology of the polygonal geometries. So it is hard for them to understand the concept of ``interior'' and ``exterior'' of polygons. Even there are sliver polygons between subject and object polygons, \veercnn~and \resnetoned~still give \rvtwo{\textit{dbo:isPartOf}} as predictions.
}

 \section{Conclusion}   \label{sec:conclusion}

In this work, we formally discuss the problem of polygon encoding \reviseone{--} a general-purpose representation learning model for polygonal geometries that can be used in various polygon-based tasks including shape classification, spatial relation prediction, building pattern classification, geographic question answering, and so on.
However, polygon encoding is not an easy task given the fact that polygonal geometries can have rather irregular structure, containing holes or multiple sub-polygons which cannot be easily handled by existing neural network architectures. 
To \reviseone{highlight} the uniqueness of this problem, we point out four important polygon encoding properties including loop \reviseone{origin} invariance, trivial vertex invariance, part permutation invariance, and topology awareness. 

To design a polygon encoder that can handle complex polygonal geometries (including polygons with holes and multipolygons) and at the same time satisfy those four properties, we propose the \nuftmlp~polygon encoder which utilizes Non-uniform Fourier transformation (NUFT) to transform a polygonal geometry into the spectral space and then learn the polygon embedding from these spectral features. 
We also propose another 1D CNN-based polygon encoder called \resnetoned~as a representative model \reviseone{of spatial domain polygon encoders. }
\resnetoned~ utilizes circular padding to achieve loop \reviseone{origin} invariance on simple polygons \reviseone{but fail to satisfy other three properties.}

To investigate the effectiveness and robustness of these two polygon encoders, we evaluate them and multiple existing baselines on two representative polygon-based tasks -- shape classification and spatial relation prediction. 

For the shape classification task, we construct a polygon-based shape classification dataset, \mnistcomplex, based on the well-known MNIST dataset. Experiment results show that both \resnetoned~and \nuftmlp~can outperform all baselines with statistical significant margins. Moreover, \nuftmlp~is robust to many shape-invariant geometry modifications including \reviseone{loop origin randomization}, vertex upsampling, and part permutation and is more sensitive to topological changes due to the invariance \reviseone{inherented from} the NUFT representation. In contrast, models that directly utilize the polygon vertex features such as
\resnetoned~suffer performance degradations under these geometry modifications. 

For the spatial relation prediction task, we construct two real-world datasets - \dbtopo~ and \dbtopocomplex~ based on DBpedia Knowledge Graph and OpenStreetMap. Evaluation results show that \nuftmlp~\reviseone{outperforms} all baselines on both datasets and is very robust when we vary the number of sampled frequencies in NUFT.

In addition, compared to other NUFT-based methods such as \ddsl,
\reviseone{\nuftmlp~does not use the Inverse Fast Fourier Trasnform (IFFT) so it is more flexible for the choice of the NUFT frequency maps. }
Both experiments show that by using non-integer frequency maps such as geometric grid $\fftfreqmat^{(\geometricfreqmtd)}$, \nuftmlp~can outperform \ddsl~on both tasks.

Despite these \reviseone{success, }several issues still remain to be solved.
First, the training dataset of \dbtopo~ and \dbtopocomplex~ is very
\reviseone{unbalanced since there are more \textit{dbo:isPartOf} triples than triples with other relations. This imbalance causes an overfitting issue for all current models. }
How to design spatial relation prediction model which is more robust for dataset imbalance is a very interesting research direction. Second, how to effectively utilize NUFT features in the spectral domain is also an interesting future research direction. 
\rvtwo{Moreover, instead of using predefined Fourier frequency maps such as $\fftfreqmat^{(\geometricfreqmtd)}$ and $\fftfreqmat^{(\linearfreqmtd)}$, can we let the neural network learn the optimal $\fftfreqmat$ based on network backpropagation?}
\reviseone{Finally, another interesting future direction is to combine \nuftmlp~and \resnetoned~for their different strengths.}

\rvtwo{
We believe that a general-purpose representation learning model for polygonal geometries will be a critical component of so-called \textit{spatially explicit artificial intelligence} \citep{mai2019relaxing,yan2017itdl,yan2019spatially,janowicz2020geoai,mai2020se,mai2020multiscale,li2021tobler}. It can also serve as an important building block to develop a foundation model for geospatial artificial intelligence \citep{mai2022foundation} in general.
}

\section*{Statements and Declarations}

\rvtwo{
\bmhead{Data availability}
DBpedia uses GNU General Public License and OpenStreetMap uses Open Database License. Both of them are open dataset for academic usage.
They do not have personally identifiable information for the privacy protection purpose. Experimental data and the methods developed will be openly shared for reproducibility and replicability \url{https://github.com/gengchenmai/polygon_encoder}.
}

\rvtwo{
\bmhead{Conficts of interest}
This paper has been approved by all co-authors. The authors have no competing interests to declare that are relevant to the content of this article. 
}

\rvtwo{
\bmhead{Acknowledgments}
This work is mainly funded by the National Science Foundation under Grant No. 2033521 A1 -- KnowWhereGraph: Enriching and Linking Cross-Domain Knowledge Graphs using Spatially-Explicit AI Technologies and the Office of the Director of National Intelligence (ODNI), Intelligence Advanced Research Projects Activity (IARPA), via 2021-2011000004. \rvtwo{Stefano Ermon acknowledges support from NSF (\#1651565), AFOSR (FA95501910024), ARO (W911NF-21-1-0125), Sloan Fellowship, and CZ Biohub.}
Any opinions, findings, and conclusions or recommendations expressed in this material are those of the authors and do not necessarily reflect the views of the National Science Foundation.
}


\begin{thebibliography}{82}
	\ifx \bisbn   \undefined \def \bisbn  #1{ISBN #1}\fi
	\ifx \binits  \undefined \def \binits#1{#1}\fi
	\ifx \bauthor  \undefined \def \bauthor#1{#1}\fi
	\ifx \batitle  \undefined \def \batitle#1{#1}\fi
	\ifx \bjtitle  \undefined \def \bjtitle#1{#1}\fi
	\ifx \bvolume  \undefined \def \bvolume#1{\textbf{#1}}\fi
	\ifx \byear  \undefined \def \byear#1{#1}\fi
	\ifx \bissue  \undefined \def \bissue#1{#1}\fi
	\ifx \bfpage  \undefined \def \bfpage#1{#1}\fi
	\ifx \blpage  \undefined \def \blpage #1{#1}\fi
	\ifx \burl  \undefined \def \burl#1{\textsf{#1}}\fi
	\ifx \doiurl  \undefined \def \doiurl#1{\url{https://doi.org/#1}}\fi
	\ifx \betal  \undefined \def \betal{\textit{et al.}}\fi
	\ifx \binstitute  \undefined \def \binstitute#1{#1}\fi
	\ifx \binstitutionaled  \undefined \def \binstitutionaled#1{#1}\fi
	\ifx \bctitle  \undefined \def \bctitle#1{#1}\fi
	\ifx \beditor  \undefined \def \beditor#1{#1}\fi
	\ifx \bpublisher  \undefined \def \bpublisher#1{#1}\fi
	\ifx \bbtitle  \undefined \def \bbtitle#1{#1}\fi
	\ifx \bedition  \undefined \def \bedition#1{#1}\fi
	\ifx \bseriesno  \undefined \def \bseriesno#1{#1}\fi
	\ifx \blocation  \undefined \def \blocation#1{#1}\fi
	\ifx \bsertitle  \undefined \def \bsertitle#1{#1}\fi
	\ifx \bsnm \undefined \def \bsnm#1{#1}\fi
	\ifx \bsuffix \undefined \def \bsuffix#1{#1}\fi
	\ifx \bparticle \undefined \def \bparticle#1{#1}\fi
	\ifx \barticle \undefined \def \barticle#1{#1}\fi
	\bibcommenthead
	\ifx \bconfdate \undefined \def \bconfdate #1{#1}\fi
	\ifx \botherref \undefined \def \botherref #1{#1}\fi
	\ifx \url \undefined \def \url#1{\textsf{#1}}\fi
	\ifx \bchapter \undefined \def \bchapter#1{#1}\fi
	\ifx \bbook \undefined \def \bbook#1{#1}\fi
	\ifx \bcomment \undefined \def \bcomment#1{#1}\fi
	\ifx \oauthor \undefined \def \oauthor#1{#1}\fi
	\ifx \citeauthoryear \undefined \def \citeauthoryear#1{#1}\fi
	\ifx \endbibitem  \undefined \def \endbibitem {}\fi
	\ifx \bconflocation  \undefined \def \bconflocation#1{#1}\fi
	\ifx \arxivurl  \undefined \def \arxivurl#1{\textsf{#1}}\fi
	\csname PreBibitemsHook\endcsname
	
	\bibitem{bronstein2017geometric}
	\begin{barticle}
		\bauthor{\bsnm{Bronstein}, \binits{M.M.}},
		\bauthor{\bsnm{Bruna}, \binits{J.}},
		\bauthor{\bsnm{LeCun}, \binits{Y.}},
		\bauthor{\bsnm{Szlam}, \binits{A.}},
		\bauthor{\bsnm{Vandergheynst}, \binits{P.}}:
		\batitle{Geometric deep learning: going beyond euclidean data}.
		\bjtitle{IEEE Signal Processing Magazine}
		\bvolume{34}(\bissue{4}),
		\bfpage{18}--\blpage{42}
		(\byear{2017})
	\end{barticle}
	\endbibitem
	
	\bibitem{mai2021review}
	\begin{botherref}
		\oauthor{\bsnm{Mai}, \binits{G.}},
		\oauthor{\bsnm{Janowicz}, \binits{K.}},
		\oauthor{\bsnm{Hu}, \binits{Y.}},
		\oauthor{\bsnm{Gao}, \binits{S.}},
		\oauthor{\bsnm{Yan}, \binits{B.}},
		\oauthor{\bsnm{Zhu}, \binits{R.}},
		\oauthor{\bsnm{Cai}, \binits{L.}},
		\oauthor{\bsnm{Lao}, \binits{N.}}:
		{A Review of Location Encoding for GeoAI: Methods and Applications}.
		International Journal of Geographical Information Science
		(2021)
	\end{botherref}
	\endbibitem
	
	\bibitem{monti2017geometric}
	\begin{bchapter}
		\bauthor{\bsnm{Monti}, \binits{F.}},
		\bauthor{\bsnm{Boscaini}, \binits{D.}},
		\bauthor{\bsnm{Masci}, \binits{J.}},
		\bauthor{\bsnm{Rodola}, \binits{E.}},
		\bauthor{\bsnm{Svoboda}, \binits{J.}},
		\bauthor{\bsnm{Bronstein}, \binits{M.M.}}:
		\bctitle{Geometric deep learning on graphs and manifolds using mixture model
			cnns}.
		In: \bbtitle{Proceedings of the IEEE Conference on Computer Vision and Pattern
			Recognition},
		pp. \bfpage{5115}--\blpage{5124}
		(\byear{2017})
	\end{bchapter}
	\endbibitem
	
	\bibitem{defferrard2016convolutional}
	\begin{bchapter}
		\bauthor{\bsnm{Defferrard}, \binits{M.}},
		\bauthor{\bsnm{Bresson}, \binits{X.}},
		\bauthor{\bsnm{Vandergheynst}, \binits{P.}}:
		\bctitle{Convolutional neural networks on graphs with fast localized spectral
			filtering}.
		In: \bbtitle{NIPS}
		(\byear{2016})
	\end{bchapter}
	\endbibitem
	
	\bibitem{kipf2016semi}
	\begin{botherref}
		\oauthor{\bsnm{Kipf}, \binits{T.N.}},
		\oauthor{\bsnm{Welling}, \binits{M.}}:
		Semi-supervised classification with graph convolutional networks.
		arXiv preprint arXiv:1609.02907
		(2016)
	\end{botherref}
	\endbibitem
	
	\bibitem{hamilton2017inductive}
	\begin{bchapter}
		\bauthor{\bsnm{Hamilton}, \binits{W.L.}},
		\bauthor{\bsnm{Ying}, \binits{R.}},
		\bauthor{\bsnm{Leskovec}, \binits{J.}}:
		\bctitle{Inductive representation learning on large graphs}.
		In: \bbtitle{Proceedings of the 31st International Conference on Neural
			Information Processing Systems},
		pp. \bfpage{1025}--\blpage{1035}
		(\byear{2017})
	\end{bchapter}
	\endbibitem
	
	\bibitem{schlichtkrull2018modeling}
	\begin{bchapter}
		\bauthor{\bsnm{Schlichtkrull}, \binits{M.}},
		\bauthor{\bsnm{Kipf}, \binits{T.N.}},
		\bauthor{\bsnm{Bloem}, \binits{P.}},
		\bauthor{\bsnm{Van Den~Berg}, \binits{R.}},
		\bauthor{\bsnm{Titov}, \binits{I.}},
		\bauthor{\bsnm{Welling}, \binits{M.}}:
		\bctitle{Modeling relational data with graph convolutional networks}.
		In: \bbtitle{European Semantic Web Conference},
		pp. \bfpage{593}--\blpage{607}
		(\byear{2018}).
		\bcomment{Springer}
	\end{bchapter}
	\endbibitem
	
	\bibitem{cai2019transgcn}
	\begin{bchapter}
		\bauthor{\bsnm{Cai}, \binits{L.}},
		\bauthor{\bsnm{Yan}, \binits{B.}},
		\bauthor{\bsnm{Mai}, \binits{G.}},
		\bauthor{\bsnm{Janowicz}, \binits{K.}},
		\bauthor{\bsnm{Zhu}, \binits{R.}}:
		\bctitle{Transgcn: Coupling transformation assumptions with graph convolutional
			networks for link prediction}.
		In: \bbtitle{Proceedings of the 10th International Conference on Knowledge
			Capture},
		pp. \bfpage{131}--\blpage{138}
		(\byear{2019})
	\end{bchapter}
	\endbibitem
	
	\bibitem{mai2020se}
	\begin{barticle}
		\bauthor{\bsnm{Mai}, \binits{G.}},
		\bauthor{\bsnm{Janowicz}, \binits{K.}},
		\bauthor{\bsnm{Cai}, \binits{L.}},
		\bauthor{\bsnm{Zhu}, \binits{R.}},
		\bauthor{\bsnm{Regalia}, \binits{B.}},
		\bauthor{\bsnm{Yan}, \binits{B.}},
		\bauthor{\bsnm{Shi}, \binits{M.}},
		\bauthor{\bsnm{Lao}, \binits{N.}}:
		\batitle{{SE}-{KGE}: A location-aware knowledge graph embedding model for
			geographic question answering and spatial semantic lifting}.
		\bjtitle{Transactions in GIS}
		(\byear{2020}).
		\doiurl{10.1111/tgis.12629}
	\end{barticle}
	\endbibitem
	
	\bibitem{qi2017pointnet}
	\begin{bchapter}
		\bauthor{\bsnm{Qi}, \binits{C.R.}},
		\bauthor{\bsnm{Su}, \binits{H.}},
		\bauthor{\bsnm{Mo}, \binits{K.}},
		\bauthor{\bsnm{Guibas}, \binits{L.J.}}:
		\bctitle{Pointnet: Deep learning on point sets for 3d classification and
			segmentation}.
		In: \bbtitle{Proceedings of the IEEE Conference on Computer Vision and Pattern
			Recognition},
		pp. \bfpage{652}--\blpage{660}
		(\byear{2017})
	\end{bchapter}
	\endbibitem
	
	\bibitem{li2018pointcnn}
	\begin{barticle}
		\bauthor{\bsnm{Li}, \binits{Y.}},
		\bauthor{\bsnm{Bu}, \binits{R.}},
		\bauthor{\bsnm{Sun}, \binits{M.}},
		\bauthor{\bsnm{Wu}, \binits{W.}},
		\bauthor{\bsnm{Di}, \binits{X.}},
		\bauthor{\bsnm{Chen}, \binits{B.}}:
		\batitle{Pointcnn: Convolution on x-transformed points}.
		\bjtitle{Advances in neural information processing systems}
		\bvolume{31},
		\bfpage{820}--\blpage{830}
		(\byear{2018})
	\end{barticle}
	\endbibitem
	
	\bibitem{mac2019presence}
	\begin{bchapter}
		\bauthor{\bsnm{Mac~Aodha}, \binits{O.}},
		\bauthor{\bsnm{Cole}, \binits{E.}},
		\bauthor{\bsnm{Perona}, \binits{P.}}:
		\bctitle{Presence-only geographical priors for fine-grained image
			classification}.
		In: \bbtitle{Proceedings of the IEEE International Conference on Computer
			Vision},
		pp. \bfpage{9596}--\blpage{9606}
		(\byear{2019})
	\end{bchapter}
	\endbibitem
	
	\bibitem{mai2020multiscale}
	\begin{bchapter}
		\bauthor{\bsnm{Mai}, \binits{G.}},
		\bauthor{\bsnm{Janowicz}, \binits{K.}},
		\bauthor{\bsnm{Yan}, \binits{B.}},
		\bauthor{\bsnm{Zhu}, \binits{R.}},
		\bauthor{\bsnm{Cai}, \binits{L.}},
		\bauthor{\bsnm{Lao}, \binits{N.}}:
		\bctitle{Multi-scale representation learning for spatial feature distributions
			using grid cells}.
		In: \bbtitle{The Eighth International Conference on Learning Representations}
		(\byear{2020}).
		\bcomment{openreview}
	\end{bchapter}
	\endbibitem
	
	\bibitem{masci2015geodesic}
	\begin{bchapter}
		\bauthor{\bsnm{Masci}, \binits{J.}},
		\bauthor{\bsnm{Boscaini}, \binits{D.}},
		\bauthor{\bsnm{Bronstein}, \binits{M.}},
		\bauthor{\bsnm{Vandergheynst}, \binits{P.}}:
		\bctitle{Geodesic convolutional neural networks on riemannian manifolds}.
		In: \bbtitle{Proceedings of the IEEE International Conference on Computer
			Vision Workshops},
		pp. \bfpage{37}--\blpage{45}
		(\byear{2015})
	\end{bchapter}
	\endbibitem
	
	\bibitem{lazer2009life}
	\begin{barticle}
		\bauthor{\bsnm{Lazer}, \binits{D.}},
		\bauthor{\bsnm{Pentland}, \binits{A.S.}},
		\bauthor{\bsnm{Adamic}, \binits{L.}},
		\bauthor{\bsnm{Aral}, \binits{S.}},
		\bauthor{\bsnm{Barabasi}, \binits{A.L.}},
		\bauthor{\bsnm{Brewer}, \binits{D.}},
		\bauthor{\bsnm{Christakis}, \binits{N.}},
		\bauthor{\bsnm{Contractor}, \binits{N.}},
		\bauthor{\bsnm{Fowler}, \binits{J.}},
		\bauthor{\bsnm{Gutmann}, \binits{M.}}, \betal:
		\batitle{Life in the network: the coming age of computational social science}.
		\bjtitle{Science (New York, NY)}
		\bvolume{323}(\bissue{5915}),
		\bfpage{721}
		(\byear{2009})
	\end{barticle}
	\endbibitem
	
	\bibitem{fan2019graph}
	\begin{bchapter}
		\bauthor{\bsnm{Fan}, \binits{W.}},
		\bauthor{\bsnm{Ma}, \binits{Y.}},
		\bauthor{\bsnm{Li}, \binits{Q.}},
		\bauthor{\bsnm{He}, \binits{Y.}},
		\bauthor{\bsnm{Zhao}, \binits{E.}},
		\bauthor{\bsnm{Tang}, \binits{J.}},
		\bauthor{\bsnm{Yin}, \binits{D.}}:
		\bctitle{Graph neural networks for social recommendation}.
		In: \bbtitle{The World Wide Web Conference},
		pp. \bfpage{417}--\blpage{426}
		(\byear{2019})
	\end{bchapter}
	\endbibitem
	
	\bibitem{gilmer2017neural}
	\begin{bchapter}
		\bauthor{\bsnm{Gilmer}, \binits{J.}},
		\bauthor{\bsnm{Schoenholz}, \binits{S.S.}},
		\bauthor{\bsnm{Riley}, \binits{P.F.}},
		\bauthor{\bsnm{Vinyals}, \binits{O.}},
		\bauthor{\bsnm{Dahl}, \binits{G.E.}}:
		\bctitle{Neural message passing for quantum chemistry}.
		In: \bbtitle{ICML}
		(\byear{2017})
	\end{bchapter}
	\endbibitem
	
	\bibitem{davidson2002genomic}
	\begin{barticle}
		\bauthor{\bsnm{Davidson}, \binits{E.H.}},
		\bauthor{\bsnm{Rast}, \binits{J.P.}},
		\bauthor{\bsnm{Oliveri}, \binits{P.}},
		\bauthor{\bsnm{Ransick}, \binits{A.}},
		\bauthor{\bsnm{Calestani}, \binits{C.}},
		\bauthor{\bsnm{Yuh}, \binits{C.-H.}},
		\bauthor{\bsnm{Minokawa}, \binits{T.}},
		\bauthor{\bsnm{Amore}, \binits{G.}},
		\bauthor{\bsnm{Hinman}, \binits{V.}},
		\bauthor{\bsnm{Arenas-Mena}, \binits{C.}}, \betal:
		\batitle{A genomic regulatory network for development}.
		\bjtitle{science}
		\bvolume{295}(\bissue{5560}),
		\bfpage{1669}--\blpage{1678}
		(\byear{2002})
	\end{barticle}
	\endbibitem
	
	\bibitem{li2019diffusion}
	\begin{bchapter}
		\bauthor{\bsnm{Li}, \binits{Y.}},
		\bauthor{\bsnm{Yu}, \binits{R.}},
		\bauthor{\bsnm{Shahabi}, \binits{C.}},
		\bauthor{\bsnm{Liu}, \binits{Y.}}:
		\bctitle{Diffusion convolutional recurrent neural network: Data-driven traffic
			forecasting}.
		In: \bbtitle{International Conference on Learning Representations}
		(\byear{2019})
	\end{bchapter}
	\endbibitem
	
	\bibitem{cai2020traffic}
	\begin{barticle}
		\bauthor{\bsnm{Cai}, \binits{L.}},
		\bauthor{\bsnm{Janowicz}, \binits{K.}},
		\bauthor{\bsnm{Mai}, \binits{G.}},
		\bauthor{\bsnm{Yan}, \binits{B.}},
		\bauthor{\bsnm{Zhu}, \binits{R.}}:
		\batitle{Traffic transformer: Capturing the continuity and periodicity of time
			series for traffic forecasting}.
		\bjtitle{Transactions in GIS}
		\bvolume{24}(\bissue{3}),
		\bfpage{736}--\blpage{755}
		(\byear{2020})
	\end{barticle}
	\endbibitem
	
	\bibitem{lin2018exploiting}
	\begin{bchapter}
		\bauthor{\bsnm{Lin}, \binits{Y.}},
		\bauthor{\bsnm{Mago}, \binits{N.}},
		\bauthor{\bsnm{Gao}, \binits{Y.}},
		\bauthor{\bsnm{Li}, \binits{Y.}},
		\bauthor{\bsnm{Chiang}, \binits{Y.-Y.}},
		\bauthor{\bsnm{Shahabi}, \binits{C.}},
		\bauthor{\bsnm{Ambite}, \binits{J.L.}}:
		\bctitle{Exploiting spatiotemporal patterns for accurate air quality
			forecasting using deep learning}.
		In: \bbtitle{Proceedings of the 26th ACM SIGSPATIAL International Conference on
			Advances in Geographic Information Systems},
		pp. \bfpage{359}--\blpage{368}
		(\byear{2018})
	\end{bchapter}
	\endbibitem
	
	\bibitem{apple2020kriging}
	\begin{bchapter}
		\bauthor{\bsnm{Appleby}, \binits{G.}},
		\bauthor{\bsnm{Liu}, \binits{L.}},
		\bauthor{\bsnm{Liu}, \binits{L.-P.}}:
		\bctitle{Kriging convolutional networks}.
		In: \bbtitle{Proceedinngs of AAAI 2020}
		(\byear{2020})
	\end{bchapter}
	\endbibitem
	
	\bibitem{wu2021inductive}
	\begin{bchapter}
		\bauthor{\bsnm{Wu}, \binits{Y.}},
		\bauthor{\bsnm{Zhuang}, \binits{D.}},
		\bauthor{\bsnm{Labbe}, \binits{A.}},
		\bauthor{\bsnm{Sun}, \binits{L.}}:
		\bctitle{Inductive graph neural networks for spatiotemporal kriging}.
		In: \bbtitle{Proceedings of the AAAI Conference on Artificial Intelligence},
		vol. \bseriesno{35},
		pp. \bfpage{4478}--\blpage{4485}
		(\byear{2021})
	\end{bchapter}
	\endbibitem
	
	\bibitem{xu2018encoding}
	\begin{bchapter}
		\bauthor{\bsnm{Xu}, \binits{Y.}},
		\bauthor{\bsnm{Piao}, \binits{Z.}},
		\bauthor{\bsnm{Gao}, \binits{S.}}:
		\bctitle{Encoding crowd interaction with deep neural network for pedestrian
			trajectory prediction}.
		In: \bbtitle{CVPR 2018},
		pp. \bfpage{5275}--\blpage{5284}
		(\byear{2018})
	\end{bchapter}
	\endbibitem
	
	\bibitem{zhang2019sr}
	\begin{bchapter}
		\bauthor{\bsnm{Zhang}, \binits{P.}},
		\bauthor{\bsnm{Ouyang}, \binits{W.}},
		\bauthor{\bsnm{Zhang}, \binits{P.}},
		\bauthor{\bsnm{Xue}, \binits{J.}},
		\bauthor{\bsnm{Zheng}, \binits{N.}}:
		\bctitle{Sr-lstm: State refinement for lstm towards pedestrian trajectory
			prediction}.
		In: \bbtitle{Proceedings of the IEEE/CVF Conference on Computer Vision and
			Pattern Recognition},
		pp. \bfpage{12085}--\blpage{12094}
		(\byear{2019})
	\end{bchapter}
	\endbibitem
	
	\bibitem{rao2020lstm}
	\begin{bchapter}
		\bauthor{\bsnm{Rao}, \binits{J.}},
		\bauthor{\bsnm{Gao}, \binits{S.}},
		\bauthor{\bsnm{Kang}, \binits{Y.}},
		\bauthor{\bsnm{Huang}, \binits{Q.}}:
		\bctitle{{LSTM-TrajGAN}: A deep learning approach to trajectory privacy
			protection}.
		In: \bbtitle{GIScience 2020},
		pp. \bfpage{12}--\blpage{11217}
		(\byear{2020})
	\end{bchapter}
	\endbibitem
	
	\bibitem{li2017diffusion}
	\begin{bchapter}
		\bauthor{\bsnm{Li}, \binits{Y.}},
		\bauthor{\bsnm{Yu}, \binits{R.}},
		\bauthor{\bsnm{Shahabi}, \binits{C.}},
		\bauthor{\bsnm{Liu}, \binits{Y.}}:
		\bctitle{Diffusion convolutional recurrent neural network: Data-driven traffic
			forecasting}.
		In: \bbtitle{ICLR 2018}
		(\byear{2018})
	\end{bchapter}
	\endbibitem
	
	\bibitem{veer2018deep}
	\begin{botherref}
		\oauthor{\bsnm{Veer}, \binits{R.v.}},
		\oauthor{\bsnm{Bloem}, \binits{P.}},
		\oauthor{\bsnm{Folmer}, \binits{E.}}:
		Deep learning for classification tasks on geospatial vector polygons.
		arXiv preprint arXiv:1806.03857
		(2018)
	\end{botherref}
	\endbibitem
	
	\bibitem{yan2021graph}
	\begin{barticle}
		\bauthor{\bsnm{Yan}, \binits{X.}},
		\bauthor{\bsnm{Ai}, \binits{T.}},
		\bauthor{\bsnm{Yang}, \binits{M.}},
		\bauthor{\bsnm{Tong}, \binits{X.}}:
		\batitle{Graph convolutional autoencoder model for the shape coding and
			cognition of buildings in maps}.
		\bjtitle{International Journal of Geographical Information Science}
		\bvolume{35}(\bissue{3}),
		\bfpage{490}--\blpage{512}
		(\byear{2021})
	\end{barticle}
	\endbibitem
	
	\bibitem{he2018recognition}
	\begin{barticle}
		\bauthor{\bsnm{He}, \binits{X.}},
		\bauthor{\bsnm{Zhang}, \binits{X.}},
		\bauthor{\bsnm{Xin}, \binits{Q.}}:
		\batitle{Recognition of building group patterns in topographic maps based on
			graph partitioning and random forest}.
		\bjtitle{ISPRS Journal of Photogrammetry and Remote Sensing}
		\bvolume{136},
		\bfpage{26}--\blpage{40}
		(\byear{2018})
	\end{barticle}
	\endbibitem
	
	\bibitem{yan2019graph}
	\begin{barticle}
		\bauthor{\bsnm{Yan}, \binits{X.}},
		\bauthor{\bsnm{Ai}, \binits{T.}},
		\bauthor{\bsnm{Yang}, \binits{M.}},
		\bauthor{\bsnm{Yin}, \binits{H.}}:
		\batitle{A graph convolutional neural network for classification of building
			patterns using spatial vector data}.
		\bjtitle{ISPRS journal of photogrammetry and remote sensing}
		\bvolume{150},
		\bfpage{259}--\blpage{273}
		(\byear{2019})
	\end{barticle}
	\endbibitem
	
	\bibitem{bei2019spatial}
	\begin{barticle}
		\bauthor{\bsnm{Bei}, \binits{W.}},
		\bauthor{\bsnm{Guo}, \binits{M.}},
		\bauthor{\bsnm{Huang}, \binits{Y.}}:
		\batitle{A spatial adaptive algorithm framework for building pattern
			recognition using graph convolutional networks}.
		\bjtitle{Sensors}
		\bvolume{19}(\bissue{24}),
		\bfpage{5518}
		(\byear{2019})
	\end{barticle}
	\endbibitem
	
	\bibitem{yan2020graph}
	\begin{botherref}
		\oauthor{\bsnm{Yan}, \binits{X.}},
		\oauthor{\bsnm{Ai}, \binits{T.}},
		\oauthor{\bsnm{Yang}, \binits{M.}},
		\oauthor{\bsnm{Tong}, \binits{X.}},
		\oauthor{\bsnm{Liu}, \binits{Q.}}:
		A graph deep learning approach for urban building grouping.
		Geocarto International,
		1--24
		(2020)
	\end{botherref}
	\endbibitem
	
	\bibitem{feng2019learning}
	\begin{barticle}
		\bauthor{\bsnm{Feng}, \binits{Y.}},
		\bauthor{\bsnm{Thiemann}, \binits{F.}},
		\bauthor{\bsnm{Sester}, \binits{M.}}:
		\batitle{Learning cartographic building generalization with deep convolutional
			neural networks}.
		\bjtitle{ISPRS International Journal of Geo-Information}
		\bvolume{8}(\bissue{6}),
		\bfpage{258}
		(\byear{2019})
	\end{barticle}
	\endbibitem
	
	\bibitem{zelle1996learning}
	\begin{bchapter}
		\bauthor{\bsnm{Zelle}, \binits{J.M.}},
		\bauthor{\bsnm{Mooney}, \binits{R.J.}}:
		\bctitle{Learning to parse database queries using inductive logic programming}.
		In: \bbtitle{Proceedings of the National Conference on Artificial
			Intelligence},
		pp. \bfpage{1050}--\blpage{1055}
		(\byear{1996})
	\end{bchapter}
	\endbibitem
	
	\bibitem{punjani2018template}
	\begin{bchapter}
		\bauthor{\bsnm{Punjani}, \binits{D.}},
		\bauthor{\bsnm{Singh}, \binits{K.}},
		\bauthor{\bsnm{Both}, \binits{A.}},
		\bauthor{\bsnm{Koubarakis}, \binits{M.}},
		\bauthor{\bsnm{Angelidis}, \binits{I.}},
		\bauthor{\bsnm{Bereta}, \binits{K.}},
		\bauthor{\bsnm{Beris}, \binits{T.}},
		\bauthor{\bsnm{Bilidas}, \binits{D.}},
		\bauthor{\bsnm{Ioannidis}, \binits{T.}},
		\bauthor{\bsnm{Karalis}, \binits{N.}}, \betal:
		\bctitle{Template-based question answering over linked geospatial data}.
		In: \bbtitle{Proceedings of the 12th Workshop on Geographic Information
			Retrieval},
		pp. \bfpage{1}--\blpage{10}
		(\byear{2018})
	\end{bchapter}
	\endbibitem
	
	\bibitem{scheider2021geo}
	\begin{barticle}
		\bauthor{\bsnm{Scheider}, \binits{S.}},
		\bauthor{\bsnm{Nyamsuren}, \binits{E.}},
		\bauthor{\bsnm{Kruiger}, \binits{H.}},
		\bauthor{\bsnm{Xu}, \binits{H.}}:
		\batitle{Geo-analytical question-answering with gis}.
		\bjtitle{International Journal of Digital Earth}
		\bvolume{14}(\bissue{1}),
		\bfpage{1}--\blpage{14}
		(\byear{2021})
	\end{barticle}
	\endbibitem
	
	\bibitem{mai2019relaxing}
	\begin{bchapter}
		\bauthor{\bsnm{Mai}, \binits{G.}},
		\bauthor{\bsnm{Yan}, \binits{B.}},
		\bauthor{\bsnm{Janowicz}, \binits{K.}},
		\bauthor{\bsnm{Zhu}, \binits{R.}}:
		\bctitle{Relaxing unanswerable geographic questions using a spatially explicit
			knowledge graph embedding model}.
		In: \bbtitle{AGILE},
		pp. \bfpage{21}--\blpage{39}
		(\byear{2019}).
		\bcomment{Springer}
	\end{bchapter}
	\endbibitem
	
	\bibitem{mai2021geographic}
	\begin{barticle}
		\bauthor{\bsnm{Mai}, \binits{G.}},
		\bauthor{\bsnm{Janowicz}, \binits{K.}},
		\bauthor{\bsnm{Zhu}, \binits{R.}},
		\bauthor{\bsnm{Cai}, \binits{L.}},
		\bauthor{\bsnm{Lao}, \binits{N.}}:
		\batitle{Geographic question answering: Challenges, uniqueness, classification,
			and future directions}.
		\bjtitle{AGILE: GIScience Series}
		\bvolume{2},
		\bfpage{1}--\blpage{21}
		(\byear{2021})
	\end{barticle}
	\endbibitem
	
	\bibitem{sun2014free}
	\begin{bchapter}
		\bauthor{\bsnm{Sun}, \binits{X.}},
		\bauthor{\bsnm{Christoudias}, \binits{C.M.}},
		\bauthor{\bsnm{Fua}, \binits{P.}}:
		\bctitle{Free-shape polygonal object localization}.
		In: \bbtitle{European Conference on Computer Vision},
		pp. \bfpage{317}--\blpage{332}
		(\byear{2014}).
		\bcomment{Springer}
	\end{bchapter}
	\endbibitem
	
	\bibitem{castrejon2017annotating}
	\begin{bchapter}
		\bauthor{\bsnm{Castrejon}, \binits{L.}},
		\bauthor{\bsnm{Kundu}, \binits{K.}},
		\bauthor{\bsnm{Urtasun}, \binits{R.}},
		\bauthor{\bsnm{Fidler}, \binits{S.}}:
		\bctitle{Annotating object instances with a {Polygon-RNN}}.
		In: \bbtitle{Proceedings of the IEEE Conference on Computer Vision and Pattern
			Recognition},
		pp. \bfpage{5230}--\blpage{5238}
		(\byear{2017})
	\end{bchapter}
	\endbibitem
	
	\bibitem{acuna2018efficient}
	\begin{bchapter}
		\bauthor{\bsnm{Acuna}, \binits{D.}},
		\bauthor{\bsnm{Ling}, \binits{H.}},
		\bauthor{\bsnm{Kar}, \binits{A.}},
		\bauthor{\bsnm{Fidler}, \binits{S.}}:
		\bctitle{Efficient interactive annotation of segmentation datasets with
			{Polygon-RNN++}}.
		In: \bbtitle{Proceedings of the IEEE Conference on Computer Vision and Pattern
			Recognition},
		pp. \bfpage{859}--\blpage{868}
		(\byear{2018})
	\end{bchapter}
	\endbibitem
	
	\bibitem{bai2009integrating}
	\begin{bchapter}
		\bauthor{\bsnm{Bai}, \binits{X.}},
		\bauthor{\bsnm{Liu}, \binits{W.}},
		\bauthor{\bsnm{Tu}, \binits{Z.}}:
		\bctitle{Integrating contour and skeleton for shape classification}.
		In: \bbtitle{2009 IEEE 12th International Conference on Computer Vision
			Workshops, ICCV Workshops},
		pp. \bfpage{360}--\blpage{367}
		(\byear{2009}).
		\bcomment{IEEE}
	\end{bchapter}
	\endbibitem
	
	\bibitem{wang2014bag}
	\begin{barticle}
		\bauthor{\bsnm{Wang}, \binits{X.}},
		\bauthor{\bsnm{Feng}, \binits{B.}},
		\bauthor{\bsnm{Bai}, \binits{X.}},
		\bauthor{\bsnm{Liu}, \binits{W.}},
		\bauthor{\bsnm{Latecki}, \binits{L.J.}}:
		\batitle{Bag of contour fragments for robust shape classification}.
		\bjtitle{Pattern Recognition}
		\bvolume{47}(\bissue{6}),
		\bfpage{2116}--\blpage{2125}
		(\byear{2014})
	\end{barticle}
	\endbibitem
	
	\bibitem{regalia2019computing}
	\begin{barticle}
		\bauthor{\bsnm{Regalia}, \binits{B.}},
		\bauthor{\bsnm{Janowicz}, \binits{K.}},
		\bauthor{\bsnm{McKenzie}, \binits{G.}}:
		\batitle{Computing and querying strict, approximate, and metrically refined
			topological relations in linked geographic data}.
		\bjtitle{Transactions in GIS}
		\bvolume{23}(\bissue{3}),
		\bfpage{601}--\blpage{619}
		(\byear{2019})
	\end{barticle}
	\endbibitem
	
	\bibitem{jiang2019ddsl}
	\begin{bchapter}
		\bauthor{\bsnm{Jiang}, \binits{C.}},
		\bauthor{\bsnm{Lansigan}, \binits{D.}},
		\bauthor{\bsnm{Marcus}, \binits{P.}},
		\bauthor{\bsnm{Nie{\ss}ner}, \binits{M.}}, \betal:
		\bctitle{{DDSL}: Deep differentiable simplex layer for learning geometric
			signals}.
		In: \bbtitle{Proceedings of the IEEE/CVF International Conference on Computer
			Vision},
		pp. \bfpage{8769}--\blpage{8778}
		(\byear{2019})
	\end{bchapter}
	\endbibitem
	
	\bibitem{jiang2019convolutional}
	\begin{bchapter}
		\bauthor{\bsnm{Jiang}, \binits{C.M.}},
		\bauthor{\bsnm{Wang}, \binits{D.}},
		\bauthor{\bsnm{Huang}, \binits{J.}},
		\bauthor{\bsnm{Marcus}, \binits{P.}},
		\bauthor{\bsnm{Niessner}, \binits{M.}}:
		\bctitle{Convolutional neural networks on non-uniform geometrical signals using
			euclidean spectral transformation}.
		In: \bbtitle{International Conference on Learning Representations}
		(\byear{2019})
	\end{bchapter}
	\endbibitem
	
	\bibitem{kurnianggoro2018survey}
	\begin{barticle}
		\bauthor{\bsnm{Kurnianggoro}, \binits{L.}},
		\bauthor{\bsnm{Jo}, \binits{K.-H.}}, \betal:
		\batitle{A survey of 2d shape representation: Methods, evaluations, and future
			research directions}.
		\bjtitle{Neurocomputing}
		\bvolume{300},
		\bfpage{1}--\blpage{16}
		(\byear{2018})
	\end{barticle}
	\endbibitem
	
	\bibitem{lecun1998gradient}
	\begin{barticle}
		\bauthor{\bsnm{LeCun}, \binits{Y.}},
		\bauthor{\bsnm{Bottou}, \binits{L.}},
		\bauthor{\bsnm{Bengio}, \binits{Y.}},
		\bauthor{\bsnm{Haffner}, \binits{P.}}:
		\batitle{Gradient-based learning applied to document recognition}.
		\bjtitle{Proceedings of the IEEE}
		\bvolume{86}(\bissue{11}),
		\bfpage{2278}--\blpage{2324}
		(\byear{1998})
	\end{barticle}
	\endbibitem
	
	\bibitem{randell1992spatial}
	\begin{bchapter}
		\bauthor{\bsnm{Randell}, \binits{D.A.}},
		\bauthor{\bsnm{Cui}, \binits{Z.}},
		\bauthor{\bsnm{Cohn}, \binits{A.G.}}:
		\bctitle{A spatial logic based on regions and connection}.
		In: \bbtitle{3rd International Conference on Knowledge Representation and
			Reasoning},
		pp. \bfpage{165}--\blpage{176}
		(\byear{1992})
	\end{bchapter}
	\endbibitem
	
	\bibitem{egenhofer1991point}
	\begin{barticle}
		\bauthor{\bsnm{Egenhofer}, \binits{M.J.}},
		\bauthor{\bsnm{Franzosa}, \binits{R.D.}}:
		\batitle{Point-set topological spatial relations}.
		\bjtitle{International Journal of Geographical Information System}
		\bvolume{5}(\bissue{2}),
		\bfpage{161}--\blpage{174}
		(\byear{1991})
	\end{barticle}
	\endbibitem
	
	\bibitem{zhang2012superedge}
	\begin{bchapter}
		\bauthor{\bsnm{Zhang}, \binits{Z.}},
		\bauthor{\bsnm{Fidler}, \binits{S.}},
		\bauthor{\bsnm{Waggoner}, \binits{J.}},
		\bauthor{\bsnm{Cao}, \binits{Y.}},
		\bauthor{\bsnm{Dickinson}, \binits{S.}},
		\bauthor{\bsnm{Siskind}, \binits{J.M.}},
		\bauthor{\bsnm{Wang}, \binits{S.}}:
		\bctitle{Superedge grouping for object localization by combining appearance and
			shape information}.
		In: \bbtitle{2012 IEEE Conference on Computer Vision and Pattern Recognition},
		pp. \bfpage{3266}--\blpage{3273}
		(\byear{2012}).
		\bcomment{IEEE}
	\end{bchapter}
	\endbibitem
	
	\bibitem{simonyan2014very}
	\begin{botherref}
		\oauthor{\bsnm{Simonyan}, \binits{K.}},
		\oauthor{\bsnm{Zisserman}, \binits{A.}}:
		Very deep convolutional networks for large-scale image recognition.
		arXiv preprint arXiv:1409.1556
		(2014)
	\end{botherref}
	\endbibitem
	
	\bibitem{li2016gated}
	\begin{bchapter}
		\bauthor{\bsnm{Li}, \binits{Y.}},
		\bauthor{\bsnm{Tarlow}, \binits{D.}},
		\bauthor{\bsnm{Brockschmidt}, \binits{M.}},
		\bauthor{\bsnm{Zemel}, \binits{R.}}:
		\bctitle{Gated graph sequence neural networks}.
		In: \bbtitle{ICLR 2016}
		(\byear{2016})
	\end{bchapter}
	\endbibitem
	
	\bibitem{liang2020polytransform}
	\begin{bchapter}
		\bauthor{\bsnm{Liang}, \binits{J.}},
		\bauthor{\bsnm{Homayounfar}, \binits{N.}},
		\bauthor{\bsnm{Ma}, \binits{W.-C.}},
		\bauthor{\bsnm{Xiong}, \binits{Y.}},
		\bauthor{\bsnm{Hu}, \binits{R.}},
		\bauthor{\bsnm{Urtasun}, \binits{R.}}:
		\bctitle{Polytransform: Deep polygon transformer for instance segmentation}.
		In: \bbtitle{Proceedings of the IEEE/CVF Conference on Computer Vision and
			Pattern Recognition},
		pp. \bfpage{9131}--\blpage{9140}
		(\byear{2020})
	\end{bchapter}
	\endbibitem
	
	\bibitem{atabay2016binary}
	\begin{barticle}
		\bauthor{\bsnm{Atabay}, \binits{H.A.}}:
		\batitle{Binary shape classification using convolutional neural networks}.
		\bjtitle{IIOAB J}
		\bvolume{7}(\bissue{5}),
		\bfpage{332}--\blpage{336}
		(\byear{2016})
	\end{barticle}
	\endbibitem
	
	\bibitem{atabay2016convolutional}
	\begin{barticle}
		\bauthor{\bsnm{Atabay}, \binits{H.A.}}:
		\batitle{A convolutional neural network with a new architecture applied on leaf
			classification}.
		\bjtitle{IIOAB J}
		\bvolume{7}(\bissue{5}),
		\bfpage{226}--\blpage{331}
		(\byear{2016})
	\end{barticle}
	\endbibitem
	
	\bibitem{hofer2017deep}
	\begin{bchapter}
		\bauthor{\bsnm{Hofer}, \binits{C.}},
		\bauthor{\bsnm{Kwitt}, \binits{R.}},
		\bauthor{\bsnm{Niethammer}, \binits{M.}},
		\bauthor{\bsnm{Uhl}, \binits{A.}}:
		\bctitle{Deep learning with topological signatures}.
		In: \bbtitle{NIPS}
		(\byear{2017})
	\end{bchapter}
	\endbibitem
	
	\bibitem{baker2018deep}
	\begin{barticle}
		\bauthor{\bsnm{Baker}, \binits{N.}},
		\bauthor{\bsnm{Lu}, \binits{H.}},
		\bauthor{\bsnm{Erlikhman}, \binits{G.}},
		\bauthor{\bsnm{Kellman}, \binits{P.J.}}:
		\batitle{Deep convolutional networks do not classify based on global object
			shape}.
		\bjtitle{PLoS computational biology}
		\bvolume{14}(\bissue{12}),
		\bfpage{1006613}
		(\byear{2018})
	\end{barticle}
	\endbibitem
	
	\bibitem{latecki2000shape}
	\begin{bchapter}
		\bauthor{\bsnm{Latecki}, \binits{L.J.}},
		\bauthor{\bsnm{Lakamper}, \binits{R.}},
		\bauthor{\bsnm{Eckhardt}, \binits{T.}}:
		\bctitle{Shape descriptors for non-rigid shapes with a single closed contour}.
		In: \bbtitle{Proceedings IEEE Conference on Computer Vision and Pattern
			Recognition. CVPR 2000 (Cat. No. PR00662)},
		vol. \bseriesno{1},
		pp. \bfpage{424}--\blpage{429}
		(\byear{2000}).
		\bcomment{IEEE}
	\end{bchapter}
	\endbibitem
	
	\bibitem{soderkvist2001computer}
	\begin{botherref}
		\oauthor{\bsnm{S{\"o}derkvist}, \binits{O.}}:
		Computer vision classification of leaves from swedish trees.
		PhD thesis
		(2001)
	\end{botherref}
	\endbibitem
	
	\bibitem{leibe2003analyzing}
	\begin{bchapter}
		\bauthor{\bsnm{Leibe}, \binits{B.}},
		\bauthor{\bsnm{Schiele}, \binits{B.}}:
		\bctitle{Analyzing appearance and contour based methods for object
			categorization}.
		In: \bbtitle{2003 IEEE Computer Society Conference on Computer Vision and
			Pattern Recognition, 2003. Proceedings.},
		vol. \bseriesno{2},
		p. \bfpage{409}
		(\byear{2003}).
		\bcomment{IEEE}
	\end{bchapter}
	\endbibitem
	
	\bibitem{mallah2013plant}
	\begin{botherref}
		\oauthor{\bsnm{Mallah}, \binits{C.}},
		\oauthor{\bsnm{Cope}, \binits{J.}},
		\oauthor{\bsnm{Orwell}, \binits{J.}}, et al.:
		Plant leaf classification using probabilistic integration of shape, texture and
		margin features.
		Signal Processing, Pattern Recognition and Applications
		\textbf{5}(1)
		(2013)
	\end{botherref}
	\endbibitem
	
	\bibitem{sebastian2005curves}
	\begin{barticle}
		\bauthor{\bsnm{Sebastian}, \binits{T.B.}},
		\bauthor{\bsnm{Kimia}, \binits{B.B.}}:
		\batitle{Curves vs. skeletons in object recognition}.
		\bjtitle{Signal processing}
		\bvolume{85}(\bissue{2}),
		\bfpage{247}--\blpage{263}
		(\byear{2005})
	\end{barticle}
	\endbibitem
	
	\bibitem{ronneberger2015u}
	\begin{bchapter}
		\bauthor{\bsnm{Ronneberger}, \binits{O.}},
		\bauthor{\bsnm{Fischer}, \binits{P.}},
		\bauthor{\bsnm{Brox}, \binits{T.}}:
		\bctitle{U-net: Convolutional networks for biomedical image segmentation}.
		In: \bbtitle{International Conference on Medical Image Computing and
			Computer-assisted Intervention},
		pp. \bfpage{234}--\blpage{241}
		(\byear{2015}).
		\bcomment{Springer}
	\end{bchapter}
	\endbibitem
	
	\bibitem{he2016deep}
	\begin{bchapter}
		\bauthor{\bsnm{He}, \binits{K.}},
		\bauthor{\bsnm{Zhang}, \binits{X.}},
		\bauthor{\bsnm{Ren}, \binits{S.}},
		\bauthor{\bsnm{Sun}, \binits{J.}}:
		\bctitle{Deep residual learning for image recognition}.
		In: \bbtitle{Proceedings of the IEEE Conference on Computer Vision and Pattern
			Recognition},
		pp. \bfpage{770}--\blpage{778}
		(\byear{2016})
	\end{bchapter}
	\endbibitem
	
	\bibitem{yu2018deep}
	\begin{bchapter}
		\bauthor{\bsnm{Yu}, \binits{F.}},
		\bauthor{\bsnm{Wang}, \binits{D.}},
		\bauthor{\bsnm{Shelhamer}, \binits{E.}},
		\bauthor{\bsnm{Darrell}, \binits{T.}}:
		\bctitle{Deep layer aggregation}.
		In: \bbtitle{Proceedings of the IEEE Conference on Computer Vision and Pattern
			Recognition},
		pp. \bfpage{2403}--\blpage{2412}
		(\byear{2018})
	\end{bchapter}
	\endbibitem
	
	\bibitem{rippel2015spectral}
	\begin{bchapter}
		\bauthor{\bsnm{Rippel}, \binits{O.}},
		\bauthor{\bsnm{Snoek}, \binits{J.}},
		\bauthor{\bsnm{Adams}, \binits{R.P.}}:
		\bctitle{Spectral representations for convolutional neural networks}.
		In: \bbtitle{Proceedings of the 28th International Conference on Neural
			Information Processing Systems-Volume 2},
		pp. \bfpage{2449}--\blpage{2457}
		(\byear{2015})
	\end{bchapter}
	\endbibitem
	
	\bibitem{mildenhall2020nerf}
	\begin{bchapter}
		\bauthor{\bsnm{Mildenhall}, \binits{B.}},
		\bauthor{\bsnm{Srinivasan}, \binits{P.P.}},
		\bauthor{\bsnm{Tancik}, \binits{M.}},
		\bauthor{\bsnm{Barron}, \binits{J.T.}},
		\bauthor{\bsnm{Ramamoorthi}, \binits{R.}},
		\bauthor{\bsnm{Ng}, \binits{R.}}:
		\bctitle{Nerf: Representing scenes as neural radiance fields for view
			synthesis}.
		In: \bbtitle{European Conference on Computer Vision},
		pp. \bfpage{405}--\blpage{421}
		(\byear{2020}).
		\bcomment{Springer}
	\end{bchapter}
	\endbibitem
	
	\bibitem{tancik2020fourier}
	\begin{botherref}
		\oauthor{\bsnm{Tancik}, \binits{M.}},
		\oauthor{\bsnm{Srinivasan}, \binits{P.P.}},
		\oauthor{\bsnm{Mildenhall}, \binits{B.}},
		\oauthor{\bsnm{Fridovich-Keil}, \binits{S.}},
		\oauthor{\bsnm{Raghavan}, \binits{N.}},
		\oauthor{\bsnm{Singhal}, \binits{U.}},
		\oauthor{\bsnm{Ramamoorthi}, \binits{R.}},
		\oauthor{\bsnm{Barron}, \binits{J.T.}},
		\oauthor{\bsnm{Ng}, \binits{R.}}:
		Fourier features let networks learn high frequency functions in low dimensional
		domains.
		arXiv preprint arXiv:2006.10739
		(2020)
	\end{botherref}
	\endbibitem
	
	\bibitem{vaswani2017attention}
	\begin{bchapter}
		\bauthor{\bsnm{Vaswani}, \binits{A.}},
		\bauthor{\bsnm{Shazeer}, \binits{N.}},
		\bauthor{\bsnm{Parmar}, \binits{N.}},
		\bauthor{\bsnm{Uszkoreit}, \binits{J.}},
		\bauthor{\bsnm{Jones}, \binits{L.}},
		\bauthor{\bsnm{Gomez}, \binits{A.N.}},
		\bauthor{\bsnm{Kaiser}, \binits{{\L}.}},
		\bauthor{\bsnm{Polosukhin}, \binits{I.}}:
		\bctitle{Attention is all you need}.
		In: \bbtitle{Advances in Neural Information Processing Systems},
		pp. \bfpage{5998}--\blpage{6008}
		(\byear{2017})
	\end{bchapter}
	\endbibitem
	
	\bibitem{ba2016layer}
	\begin{botherref}
		\oauthor{\bsnm{Ba}, \binits{J.L.}},
		\oauthor{\bsnm{Kiros}, \binits{J.R.}},
		\oauthor{\bsnm{Hinton}, \binits{G.E.}}:
		Layer normalization.
		arXiv preprint arXiv:1607.06450
		(2016)
	\end{botherref}
	\endbibitem
	
	\bibitem{ha2018sketchrnn}
	\begin{bchapter}
		\bauthor{\bsnm{Ha}, \binits{D.}},
		\bauthor{\bsnm{Eck}, \binits{D.}}:
		\bctitle{A neural representation of sketch drawings}.
		In: \bbtitle{International Conference on Learning Representations}
		(\byear{2018})
	\end{bchapter}
	\endbibitem
	
	\bibitem{deng2021vector}
	\begin{bchapter}
		\bauthor{\bsnm{Deng}, \binits{C.}},
		\bauthor{\bsnm{Litany}, \binits{O.}},
		\bauthor{\bsnm{Duan}, \binits{Y.}},
		\bauthor{\bsnm{Poulenard}, \binits{A.}},
		\bauthor{\bsnm{Tagliasacchi}, \binits{A.}},
		\bauthor{\bsnm{Guibas}, \binits{L.J.}}:
		\bctitle{Vector neurons: A general framework for so (3)-equivariant networks}.
		In: \bbtitle{Proceedings of the IEEE/CVF International Conference on Computer
			Vision},
		pp. \bfpage{12200}--\blpage{12209}
		(\byear{2021})
	\end{bchapter}
	\endbibitem
	
	\bibitem{esteves2018learning}
	\begin{bchapter}
		\bauthor{\bsnm{Esteves}, \binits{C.}},
		\bauthor{\bsnm{Allen-Blanchette}, \binits{C.}},
		\bauthor{\bsnm{Makadia}, \binits{A.}},
		\bauthor{\bsnm{Daniilidis}, \binits{K.}}:
		\bctitle{Learning so (3) equivariant representations with spherical cnns}.
		In: \bbtitle{Proceedings of the European Conference on Computer Vision (ECCV)},
		pp. \bfpage{52}--\blpage{68}
		(\byear{2018})
	\end{bchapter}
	\endbibitem
	
	\bibitem{bordes2013translating}
	\begin{bchapter}
		\bauthor{\bsnm{Bordes}, \binits{A.}},
		\bauthor{\bsnm{Usunier}, \binits{N.}},
		\bauthor{\bsnm{Garcia-Duran}, \binits{A.}},
		\bauthor{\bsnm{Weston}, \binits{J.}},
		\bauthor{\bsnm{Yakhnenko}, \binits{O.}}:
		\bctitle{Translating embeddings for modeling multi-relational data}.
		In: \bbtitle{Neural Information Processing Systems (NIPS)},
		pp. \bfpage{1}--\blpage{9}
		(\byear{2013})
	\end{bchapter}
	\endbibitem
	
	\bibitem{chen2014parameterized}
	\begin{bchapter}
		\bauthor{\bsnm{Chen}, \binits{W.}}:
		\bctitle{Parameterized spatial sql translation for geographic question
			answering}.
		In: \bbtitle{2014 IEEE International Conference on Semantic Computing},
		pp. \bfpage{23}--\blpage{27}
		(\byear{2014}).
		\bcomment{IEEE}
	\end{bchapter}
	\endbibitem
	
	\bibitem{yan2017itdl}
	\begin{bchapter}
		\bauthor{\bsnm{Yan}, \binits{B.}},
		\bauthor{\bsnm{Janowicz}, \binits{K.}},
		\bauthor{\bsnm{Mai}, \binits{G.}},
		\bauthor{\bsnm{Gao}, \binits{S.}}:
		\bctitle{From itdl to place2vec: Reasoning about place type similarity and
			relatedness by learning embeddings from augmented spatial contexts}.
		In: \bbtitle{Proceedings of the 25th ACM SIGSPATIAL International Conference on
			Advances in Geographic Information Systems},
		pp. \bfpage{1}--\blpage{10}
		(\byear{2017})
	\end{bchapter}
	\endbibitem
	
	\bibitem{yan2019spatially}
	\begin{barticle}
		\bauthor{\bsnm{Yan}, \binits{B.}},
		\bauthor{\bsnm{Janowicz}, \binits{K.}},
		\bauthor{\bsnm{Mai}, \binits{G.}},
		\bauthor{\bsnm{Zhu}, \binits{R.}}:
		\batitle{A spatially explicit reinforcement learning model for geographic
			knowledge graph summarization}.
		\bjtitle{Transactions in GIS}
		\bvolume{23}(\bissue{3}),
		\bfpage{620}--\blpage{640}
		(\byear{2019})
	\end{barticle}
	\endbibitem
	
	\bibitem{janowicz2020geoai}
	\begin{botherref}
		\oauthor{\bsnm{Janowicz}, \binits{K.}},
		\oauthor{\bsnm{Gao}, \binits{S.}},
		\oauthor{\bsnm{McKenzie}, \binits{G.}},
		\oauthor{\bsnm{Hu}, \binits{Y.}},
		\oauthor{\bsnm{Bhaduri}, \binits{B.}}:
		GeoAI: spatially explicit artificial intelligence techniques for geographic
		knowledge discovery and beyond.
		Taylor \& Francis
		(2020)
	\end{botherref}
	\endbibitem
	
	\bibitem{li2021tobler}
	\begin{barticle}
		\bauthor{\bsnm{Li}, \binits{W.}},
		\bauthor{\bsnm{Hsu}, \binits{C.-Y.}},
		\bauthor{\bsnm{Hu}, \binits{M.}}:
		\batitle{Tobler’s first law in geoai: A spatially explicit deep learning
			model for terrain feature detection under weak supervision}.
		\bjtitle{Annals of the American Association of Geographers}
		\bvolume{111}(\bissue{7}),
		\bfpage{1887}--\blpage{1905}
		(\byear{2021})
	\end{barticle}
	\endbibitem
	
	\bibitem{mai2022foundation}
	\begin{bchapter}
		\bauthor{\bsnm{Mai}, \binits{G.M.}},
		\bauthor{\bsnm{Cundy}, \binits{C.}},
		\bauthor{\bsnm{Choi}, \binits{K.}},
		\bauthor{\bsnm{Hu}, \binits{Y.}},
		\bauthor{\bsnm{Lao}, \binits{N.}},
		\bauthor{\bsnm{Ermon}, \binits{S.}}:
		\bctitle{Towards a foundation model for geospatial artificial intelligence}.
		In: \bbtitle{Proceedings of the 30th SIGSPATIAL International Conference on
			Advances in Geographic Information Systems}
		(\byear{2022}).
		\doiurl{10.1145/3557915.3561043}
	\end{bchapter}
	\endbibitem
	
\end{thebibliography}

\newpage
\begin{appendices}

\section{Appendix}

\subsection{The Type Statistic of Polygonal Geometries in \dbtopocomplex} \label{sec:place_type_map}

\vspace{-0.3cm}
\begin{table}[ht!]
\caption{The place type statistic of geographic entities in \dbtopo~and \dbtopocomplex~dataset.
	}
	\label{tab:dbsparel_type}
	\centering
\begin{tabular}{l|c}
\toprule
Place Type           & Entity Count \\ \hline
City                 & 7887         \\
Town                 & 6668         \\
Settlement           & 3688         \\
Village              & 2502         \\
AdministrativeRegion & 1420         \\
CityDistrict         & 980          \\
Unknown              & 57           \\
ProtectedArea        & 20           \\
BodyOfWater          & 12           \\
ManMadeFeatures      & 12           \\
Park                 & 9            \\
HistoricPlace        & 3            \\
Island               & 2            \\
Location             & 2            \\
MountainPass         & 1            \\
Mountain             & 1            \\
\bottomrule
\end{tabular}
\end{table}

\subsection{Model Hyperparameter Tuning} \label{sec:para_tune}
We use grid search for hyperparameter tuning. For all polygon encoders on both tasks, we tune the learning rate $\lr$ over $\{0.02, 0.01, 0.005, 0.002, 0.001\}$,  the polygon embedding dimension over $ \pgonembdim \in \{256, 512, 1024\}$. As for all \ddsl~and \nuftmlp-based models, we tune the frequency number $\fftfreqnumX = \{16, 20, 24, 28, 32, 36, 40, 44\}$ for the shape classification task while $\fftfreqnumX = \{16, 32, 64\}$ for the spatial relation prediction task. As for \nuftmlp[\geometricfreqmtd]-based models, we tune the $\minscale=\{0.2, 0.4, 0.5, 0.8, 1.0\}$ and we tune $\maxscale$ around $\fftfreqnumX/2$. For all PCA models, we vary $\pcadim$ such that the top $\pcadim$ PCA components can account for different data variance $\pcavar = \{80\%, 85\%, 90\%, 95\%\}$.
As for \resnetoned, we tune the \kdeltaenc~point encoder's neighbor size $ 2\kdelta \in \{0, 2, 4, 6, 8, 10, 12, 14, 16, 18, 20\}$ and tune the number of $\resnetonedlayer$ - $\resnetonedlayernum \in \{ 1, 2, 3\}$. For \ddsllenet, we tune the hidden dimension of LeNet5 over $\{128, 256, 512, 1024\}$. As to \nuftmlp-based models, \ddsl+MLP, and \ddsl+PCA+MLP, we tune the number of hidden layers $\nuftnumhiddenlayer$ and the number of hidden dimension $\nuftnumhiddendim$ in $\nuftspecmlpfunc(\cdot)$ over $\nuftnumhiddenlayer = \{1, 2, 3\}$, $\nuftnumhiddendim = \{512, 1024\}$. We also try different NUFT spectral feature normalization method $\nuftspecnorm(\cdot)$ such as no normalization, L2 normalization, and batch normalization. We find out no normalization usually leads to the best performance on all three datasets.

The best hyperparameter combinations for all models on \mnistcomplex~are shown in Table \ref{tab:mnist_hyperpara}. As for \dbtopo~and \dbtopocomplex, each model's best hyperparameter combinations are shown in Table \ref{tab:dbtopo_hyperpara}.

\begin{table}[]
\caption{The best hyperparameter combinations for each model on  \mnistcomplex~dataset.
	}
	\label{tab:mnist_hyperpara}
	\centering
	\setlength{\tabcolsep}{3pt}
\begin{tabular}{l|r|l|l|l|l|l|l|l|l}
\toprule
Model                     & $\lr$     & $\pgonembdim$ & $\fftfreqnumX$ & $\minscale$ & $\maxscale$ & $\pcavar$ & $\pcadim$ &  $\resnetonedlayernum$ & $2\kdelta$ \\ \hline
\veercnn                   & 0.01   & 1024             & -   & -                 & -                 & -                       & -              & -                          & -        \\
\resnetoned                  & 0.01   & 512              & -   & -                 & -                 & -                       & -              & 3                          & 12       \\
\ddsl+MLP                  & 0.001  & 512              & 24  & -                 & -                 & -                       & -              & -                          & -        \\
\ddsl+PCA+MLP              & 0.0005 & 512              & 24  & -                 & -                 & 90\%                    & 39             & -                          & -        \\
\nuftmlp[\linearfreqmtd]+MLP     & 0.0005 & 512              & 24  & -                 & -                 & -                       & -              & -                          & -        \\
\nuftmlp[\linearfreqmtd]+PCA+MLP & 0.0005 & 512              & 24  & -                 & -                 & 80\%                    & 42             & -                          & -        \\
\nuftmlp[\geometricfreqmtd]+MLP     & 0.0005 & 512              & 24  & 0.5               & 12                & -                       & -              & -                          & -        \\
\nuftmlp[\geometricfreqmtd]+PCA+MLP & 0.0005 & 512              & 24  & 0.5               & 12                & 95\%                    & 46             & -                          & -       \\
                              \bottomrule
\end{tabular}
\end{table}

\begin{table}[]
\caption{The best hyperparameter combinations for each model on  \dbtopo~and \dbtopocomplex~dataset.
	}
	\label{tab:dbtopo_hyperpara}
	\centering
	\setlength{\tabcolsep}{4pt}
\begin{tabular}{l|l|r|c|c|c|c|c|c}
\toprule
Dataset                        & Model                 & $\lr$    & $\pgonembdim$ & $\fftfreqnumX$ & $\minscale$ & $\maxscale$ & $\resnetonedlayernum$ & $2\kdelta$ \\  \hline
\multirow{6}{*}{\dbtopo}        & \veercnn               & 0.01  & 1024             & -            & -                 & -                 & -                          & -        \\
                               & \ddsllenet           & 0.01  & \reviseone{512}                & 32           & -                 & -                 & -                          & -        \\
                               & \ddsl+MLP              & 0.001 & 512              & 32           & -                 & -                 & -                          & -        \\
                               & \resnetoned              & 0.01  & 512              & -            & -                 & -                 & 1                          & 4        \\
                               & \nuftmlp[\linearfreqmtd]+MLP & 0.01  & 512              & 16           & -                 & -                 & -                          & -        \\
                               & \nuftmlp[\geometricfreqmtd]+MLP & 0.002 & 512              & 32           & 0.8               & 16                & -                          & -        \\ \hline
\multirow{6}{*}{\dbtopocomplex} & \veercnn               & 0.02  & 512              & -            & -                 & -                 & -                          & -        \\
                               & \ddsllenet           & 0.01  & \reviseone{512}                & 32           & -                 & -                 & -                          & -        \\
                               & \ddsl+MLP              & 0.001 & 512              & 32           & -                 & -                 & -                          & -        \\
                               & \resnetoned              & 0.02  & 512              & -            & -                 & -                 & 1                          & 4        \\
                               & \nuftmlp[\linearfreqmtd]+MLP & 0.01  & 512              & 32           & -                 & -                 & -                          & -        \\
                               & \nuftmlp[\geometricfreqmtd]+MLP & 0.002 & 512              & 32           & 0.5               & 20                & -                          & -       \\
                               \bottomrule
\end{tabular}
\end{table}

 \end{appendices}

\end{document}